\documentclass[runningheads]{llncs}

\usepackage{eccv}

\usepackage{eccvabbrv}

\usepackage{graphicx}
\usepackage{multirow}
\usepackage{booktabs}
\usepackage{pifont}
\usepackage{bm}
\usepackage{makecell}
\usepackage{algorithm,algorithmic}

\usepackage[accsupp]{axessibility}  

\usepackage[pagebackref,breaklinks,colorlinks,citecolor=eccvblue]{hyperref}

\usepackage{orcidlink}

\begin{document}

\title{OvSW: Overcoming Silent Weights for Accurate Binary Neural Networks} 


\author{
Jingyang Xiang\inst{1}\orcidlink{0000-0001-5350-1528} \and
Zuohui Chen\inst{2}\orcidlink{0000-0003-1806-6676} \and
Siqi Li\inst{1}\orcidlink{0009-0000-4632-9010} \and
Qing Wu\inst{3} \and
Yong Liu\inst{1,4}\orcidlink{0000-0003-4822-8939}\thanks{Corresponding author: yongliu@iipc.zju.edu.cn.}
}

\authorrunning{Xiang et al.}

\institute{APRIL Lab, Zhejiang University \and
Zhejiang University of Technology \and
College of Computer Science, Hangzhou Dianzi University \and 
Huzhou Institute, Zhejiang University}

\maketitle

\begin{abstract}
Binary Neural Networks~(BNNs) have been proven to be highly effective for deploying deep neural networks on mobile and embedded platforms.
Most existing works focus on minimizing quantization errors, improving representation ability, or designing gradient approximations to alleviate gradient mismatch in BNNs,
while leaving the weight sign flipping, a critical factor for achieving powerful BNNs, untouched.
In this paper, we investigate the efficiency of weight sign updates in BNNs.
We observe that, for vanilla BNNs, over 50\% of the weights remain their signs unchanged during training,
and these weights are not only distributed at the tails of the weight distribution but also universally present in the vicinity of zero.
We refer to these weights as ``silent weights'', which slow down convergence and lead to a significant accuracy degradation.
Theoretically, we reveal this is due to the independence of the BNNs gradient from the latent weight distribution.
To address the issue, we propose Overcome Silent Weights~(OvSW).
OvSW first employs Adaptive Gradient Scaling~(AGS) to establish a relationship between the gradient and the latent weight distribution, thereby improving the overall efficiency of weight sign updates.
Additionally, we design Silence Awareness Decaying~(SAD) to automatically identify ``silent weights'' by tracking weight flipping state,
and apply an additional penalty to ``silent weights'' to facilitate their flipping.
By efficiently updating weight signs, our method achieves faster convergence and state-of-the-art performance on CIFAR10 and ImageNet1K dataset with various architectures.
For example, OvSW obtains 61.6\% and 65.5\% top-1 accuracy on the ImageNet1K using binarized ResNet18 and ResNet34 architecture respectively.
Codes are available at \url{https://github.com/JingyangXiang/OvSW}.
\end{abstract}

\section{Introduction}
\label{sec:introduction}

Deep neural networks~(DNNs) have shown tremendous success in various computer vision tasks,
including image classfication~\cite{krizhevsky2012imagenet, he2016deep}, object detection~\cite{girshick2014rich, he2015spatial, ren2016faster}, and semantic segmentation~\cite{long2015fully, he2017mask}.
However, the remarkable performance is always attributed to deeper and wider architectures~\cite{simonyan2014very, he2016deep},
which comes with expensive memory and computational overhead and makes it challenging to deploy DNNs on resource-constrained edge platforms.
%

%
The community has been delving into model compression,
which aims to reduce inference overhead for DNNs while preserving their performance.
Typical techniques include, but are not limited to, efficient architecture design~\cite{zhang2018shufflenet, ding2021repvgg}, neural architecture search~\cite{su2021prioritized, duan2021transnas},
network pruning~\cite{luo2020autopruner, lin2020hrank}, knowledge distillation~\cite{hinton2015distilling, li2022learning}, and network quantization~\cite{gong2019differentiable, zhu2020towards, chen2020learning}.
Among them, network quantization is widely suggested as a promising solution.
It effectively enhances memory efficiency and execution speed on embedding platforms by reducing the weight and activation bits and replacing expensive floating-point arithmetic with relatively inexpensive fixed-point arithmetic.
Extensive studies~\cite{jacob2018quantization, zhang2018lq, gong2019differentiable, yang2019quantization} have demonstrated its effectiveness and yielded light and efficient DNNs.

In this paper, we focus on studying binary neural networks~(BNNs), which are an aggressive form of quantized neural networks~(QNNs).
BNNs binarize both weights and activations to discrete values~($\{+1, -1\}$), which can be represented by 1-bit and reduce memory usage by 32$\times$. 
On the other hand, by employing efficient bitwise operations instead of floating-point ones,
BNNs also significantly reduce computation complexity, providing 58$\times$ speedup as reported by XNOR-Net~\cite{rastegari2016xnor}. 
These characteristics make them well-suited for deployment on low-power embedding platforms, including FPGA, ASICs, and IoT devices~\cite{horowitz20141}.

\begin{figure*}[t]
{
    \centering
    \begin{subtable}[h]{0.32\textwidth}
        \includegraphics[width=1.\linewidth]{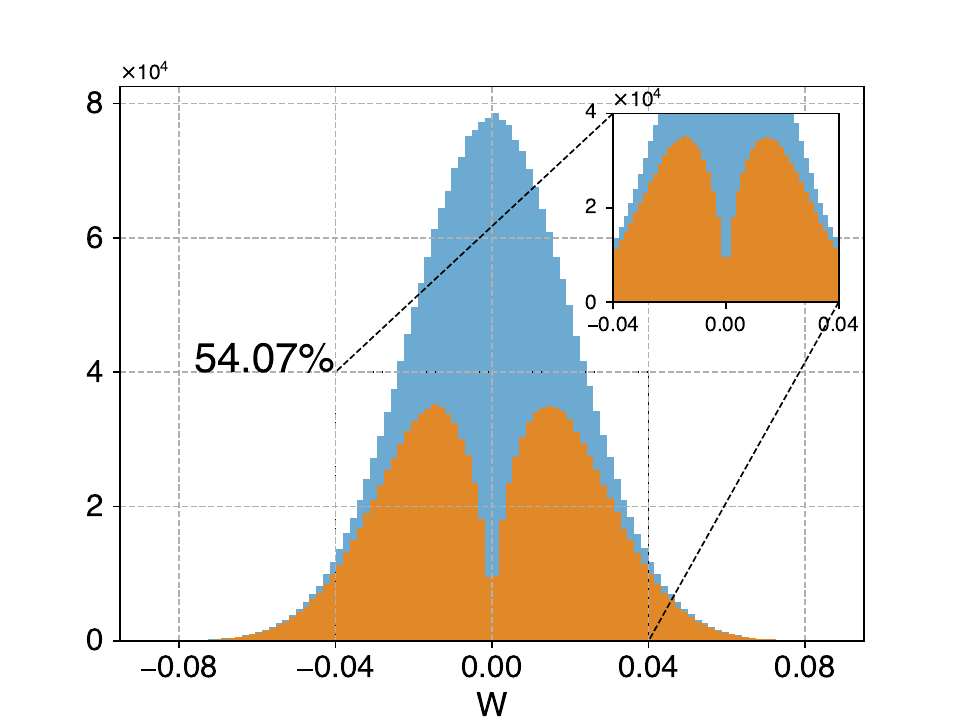}
        \caption{Vanilla}
        \label{fig:flip_xnor}
    \end{subtable}
    \begin{subtable}[h]{0.32\textwidth}
        \includegraphics[width=1\linewidth]{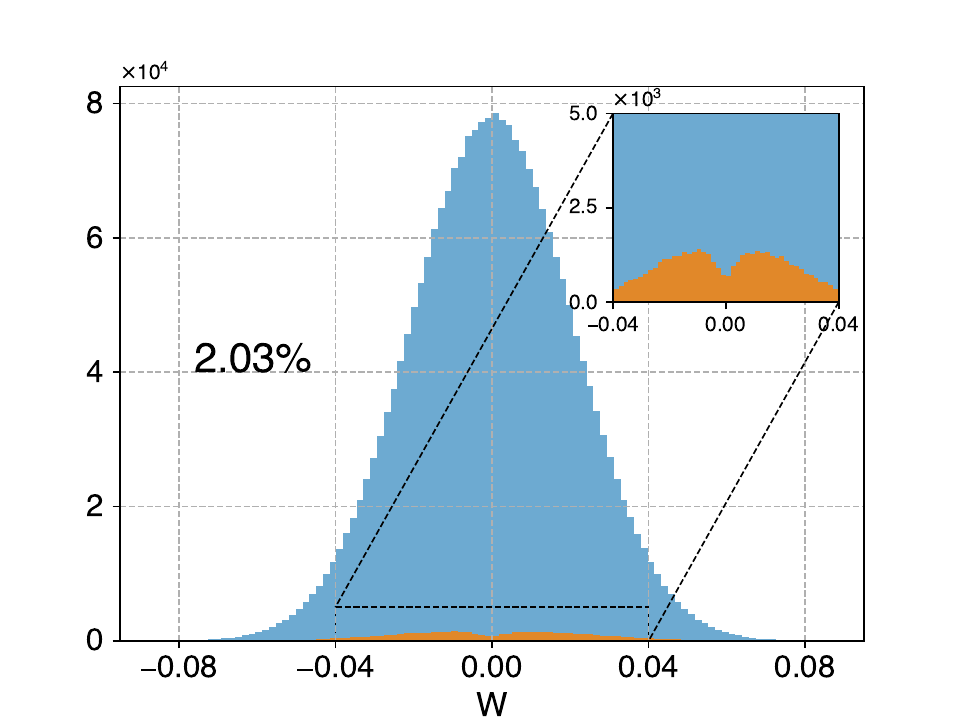}
        \caption{OvSW}
        \label{fig:flip_ovsw}
    \end{subtable}
    \begin{subtable}[h]{0.32\textwidth}
        \includegraphics[width=1\linewidth]{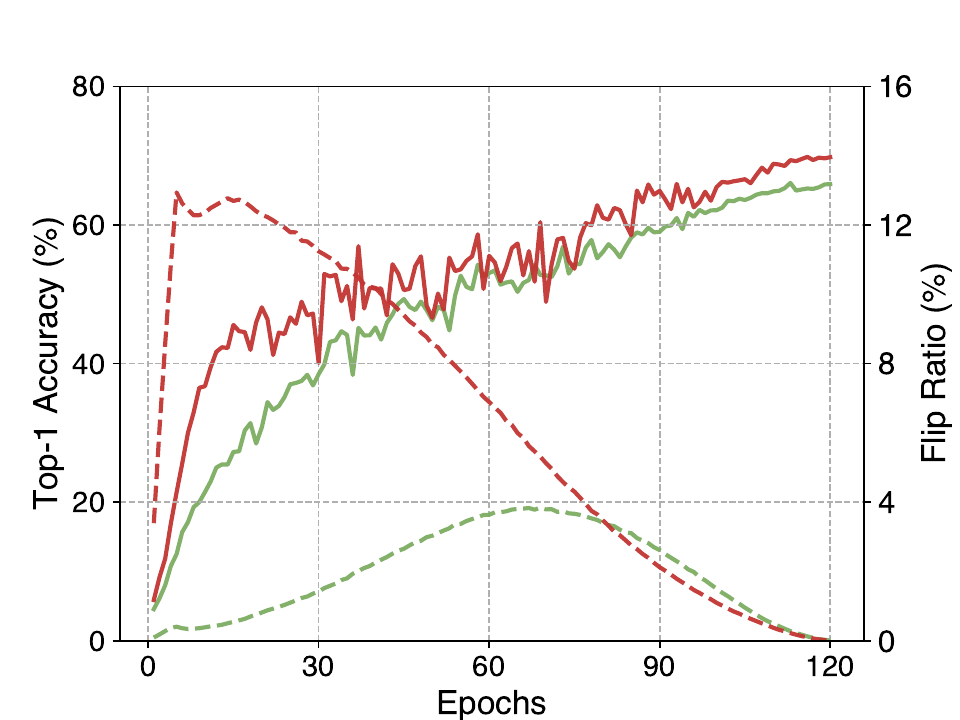}
        \caption{Top-1 Acc \& Flip Ratio}
        \label{fig:convergence_and_flip}
    \end{subtable}

    \caption{
    (a) and (b): Histogram of the initialized weight distribution~(blue) and the weights
    that never update signs throughout training~(orange).
    54.07\% and 2.03\% represent the ratio of the corresponding orange area to the blue.
    (c) Top-1 Accuracy~(solid lines) and Flip Ratio~(dashed lines) are for Vanilla~(green) and OvSW~(red).
    %
    %
    (a), (b) and Flip Ratio in (c) is for layer4.conv2.weight.
    }
    \label{fig:main_1}
    \vspace{-0.2in}
}
\end{figure*}

However, the application of BNNs is still limited due to constraining both weights and activations to 1-bit significantly reducing accuracy compared to full-precision models. 
To address this problem, many methods have been developed to reduce quantization error and enhance the representation capability, thereby closing the performance gap between BNNs and their real-valued counterparts.

Although these methods have effectively narrowed the performance gap,
they still overlook the essence of BNN optimization is to update the signs of latent weights~\cite{helwegen2019latent}, instead of updating their values.
In other words, if the signs of the latent weights don't change, the BNNs are hardly ever updated.
To demonstrate the weight sign flipping of vanilla BNNs, we train binarized ResNet18 on CIFAR100 with 120 epochs and plot the flipping information for layer4.conv2.weight.
As shown in \cref{fig:flip_xnor}, more than 50\% of the weights never change their signs throughout training process, which slows down convergence
and leads to a significant accuracy degradation for BNNs.
We denote the part of weights that hardly change their signs during training as ``silent weights''.

%

%
It's worth noting that, as a study mostly related to ours, ReCU~\cite{xu2021recu} proposes a rectified clamp unit to revive the ``dead weights'' for updating,
which refers to a group of weights that are distributed at the tails of the weight distribution and barely update their signs during training.
They claim that the magnitudes of latent weights do not contribute to the forward propagation.
If two weights have the same sign, they have the same effect on the forward propagation.
Intuitively, the signs of the weights around the zero are easily changed, while weights distributed at the tails of the weight distribution tend to have difficulties in changing their signs.
Although ReCU facilitates the updating for ``dead weights'', it ignores the fact, the gradient distribution across layers and weights have great distinction.
Weights with larger magnitudes are intuitively more difficult to flip their signs, but small weights may also suffer from silence once their gradients are relatively small compared to their magnitudes.
As shown in \cref{fig:flip_xnor}, for a vanilla BNN, these weights are not only distributed at the tails of the weight distribution but also universally present in the vicinity of zero.
As a result, it is still sub-optimal and remains to be improved.

In this paper, we reveal that the root cause of the large number of ``silent weights'' is attributed to
the independence of the BNN gradients from the latent weight distribution
via a systematical and theoretical analysis.
To this end, two simple yet effective techniques, including Adaptive Gradient Scaling~(AGS) and Silence Awareness Decaying~(SAD), are introduced to Overcome Silent Weights~(OvSW).
Specifically, AGS adaptively scales the gradients by establishing a relationship between the gradients and the latent weight distribution, thereby improving the overall efficiency of weight sign updates.
Meanwhile, SAD automatically identifies ``silent weights'' by tracking weight sign flipping state and applies an additional penalty to them,
further promoting the updating efficiency of their signs.
Benefiting from AGS and SAD, OvSW enables efficient flipping of weight signs, significantly accelerates the convergence and promotes the performance for BNNs as shown in \cref{fig:flip_ovsw} and \cref{fig:convergence_and_flip}.
In comparison to the expensive computational cost of forward and backward propagation,
OvSW introduces negligible extra overhead to the training process, maintaining its efficiency.
Furthermore, since OvSW aims to facilitate the weight sign flipping for BNNs, which is orthogonal to the previous work,
it has good compatibility and can be used as plug-and-play modules to enhance other BNNs' performance.

The contributions of our work are highlighted as follows:
\begin{itemize}
\item We are the first to find that a large number of weights in BNNs fail to update their signs throughout the training.
These ``silent weights'' are not only distributed at the tails of the weight distribution but also universally present in the vicinity of zero.
Theoretically, we reveal this is attributed to the independence of the gradients from the distribution of the latent weights.

\item We propose to overcome ``silent weights'' with two novel techniques:
(1) an adaptive gradient scaling method to scale the gradients according to the distribution of the latent weights,
which improves overall efficiency in updating signs for weights;
(2) a silence awareness decaying strategy to identify ``silent weights'' by tracking weight sign flipping state and introduce an additional penalty to them,
further facilitating the flipping of their signs.

\item Extensive experiments for various BNNs including ResNet18/20~\cite{he2016deep} and VGGsmall~\cite{simonyan2014very} on CIFAR10~\cite{krizhevsky2009learning} and ResNet18/34~\cite{he2016deep} on ImageNet1K~\cite{russakovsky2015imagenet} demonstrate the effectiveness of our method.
For example, OvSW achieves 61.6\% and 65.5\% top-1 accuracy on the ImageNet for binarized ResNet18 and ResNet34, improving ReCU~\cite{xu2021recu} by 0.6\% and 0.4\% respectively.
%
%
Besides, for ResNet18,
%
OvSW achieves 2.00\% and 2.83\% improvement on CIFAR100~\cite{krizhevsky2009learning}
when combined with AdaBin~(enhancing representation ability) and RBNN~(training-aware gradient approximation) respectively, demonstrating its good compatibility with the other methods.

\end{itemize}
\section{Related Work}
\label{sec:related_work}

Great efforts have been put into reducing the performance gap between BNNs and their real-valued counterparts.
The pioneering BNN work dates back to Courbariaux~\etal~\cite{courbariaux2016binarized}, which binarizes weights and activations to +1 or -1 by $\mathrm{sign}$ function.
However, this aggressive approach limits the representation ability of BNNs to the binary space, resulting in a significant accuracy degradation.
To reduce the quantization error and improve their accuracy, XNOR-Net~\cite{rastegari2016xnor} introduces a scaling factor obtained through the $\ell_1$-norm of weights or activations.
Furthermore, XNOR-Net++~\cite{bulat2019xnor} merges two scale factors from the weights and activations into a trainable parameter and optimizes them via backpropagation.
ABC-Net~\cite{lin2017towards} employs a linear combination of multiple binary weight sets to approximate the real-valued weights and alleviate the information loss.
Bi-Real Net~\cite{liu2018bi} connects the real-valued activations to the consecutive block via an identity shortcut,
which significantly enhances network representation ability while incurring negligible computational overhead.
AdaBin~\cite{tu2022adabin} introduces adaptive binary sets to fit different distributions, enhancing binarized representation ability.
UaBNN~\cite{zhao2021uncertainty} introduces an uncertainty-aware BNN to reduce these weights’ uncertainties.
INSTA-BNN~\cite{lee2023insta} controls the quantization threshold in an input instance-aware manner, taking higher-order statistics into account.

Apart from enhancing the representation ability of BNNs, gradient estimation is also one of the critical research directions,
since gradients in the sign function are almost zero everywhere.
Straight through estimator~(STE)~\cite{bengio2013estimating} is the most widely used function to enable the gradient to backpropagate.
However, the gradient error is huge for STE and will accumulate during backpropagation, leading to instability in training and severe accuracy degradation.
To alleviate this, Bi-Real Net~\cite{liu2018bi} utilizes a piecewise polynomial function as the approximation function.
IR-Net~\cite{qin2020forward} proposes an error decay estimator and RBNN~\cite{lin2020rotated} employs a training-aware approximation function to replace the sign function.
EWGS~\cite{lee2021network} takes discretization error between input and output into account and introduces element-wise gradient scaling to adaptively scale up or down each gradient element.
ReSTE~\cite{wu2023estimator} revises the original STE and balance the estimating error and the gradient stability well.
All of these methods effectively reduce the gradient error and achieve consistent improvement in both training stability and accuracy compared to the vanilla STE.
\section{Background}
\label{sec:background}

We briefly review the optimization process of BNNs in this section.  
Given a DNN, we denote $\mathcal{W}_{j}\in \mathbb{R}^{C^{j}_{\text{out}}\times C_{\text{in}}^{j}\times K^{j}_{\text{h}} \times K^{j}_{\text{w}}}$ as the real-valued weight in the $j$-th layer,
$C^{j}_{\text{out}}$, $C^{j}_{\text{in}}$, $K^{j}_{\text{h}}$ and $K^{j}_{\text{w}}$ are the number of output channels, input channels, kernel height, and kernel width, respectively. 
Let the real-valued activation be $\mathcal{A}_{j}$, then the convolution process can be formulated as
\begin{equation}
    \mathcal{A}_{j+1}=\mathcal{W}_{j}*\mathcal{A}_{j},
    \label{eq:conv}
\end{equation}
where $*$ represents the standard convolution operation.

BNNs aim to binarize weights $\mathcal{W}_{j}$ and activations $\mathcal{A}_{j}$ to discrete values~($\{+1,-1\}$) through $\mathrm{sign}$ function:
\begin{equation}
    \label{eq:sign}
        \hat{x}=\mathrm{sign}\left( x \right) =\left\{ 
            \begin{matrix}
            	+1,& \mathrm{if}\ x\ge 0,\\
            	-1,& \mathrm{otherwise}.\\
            \end{matrix} \right.
\end{equation}

%
To reduce the quantization error in BNNs,
XNOR-Net~\cite{rastegari2016xnor} introduces two scale factors to approximate the binarized weights $\hat{\mathcal{W}}^{b}_{j}$ and activation $\hat{\mathcal{A}}^{b}_{j}$.
Furthermore, XNOR-Net++~\cite{bulat2019xnor} proposes fusing the activation and weight scaling factors into one and optimizing it via backpropagation.
This approach significantly outperforms XNOR-Net within the same computational budget and has been widely adopted by recent works~\cite{liu2018bi, qin2020forward, liu2020reactnet, xu2021recu}.
Following them, we denote the scaling factor as $\alpha_j$.
Then the binary convolution operation can be formulated as
\begin{equation}
    \mathcal{A}_{j+1}=\left(\hat{\mathcal{A}}_{j}\circledast \hat{\mathcal{W}}_{j} \right) \odot \alpha_j,
    \label{eq:binary_conv}
\end{equation}
%
where $\circledast$ is the efficient XNOR and Bitcount operation
and $\odot$ is the hadamard product.
In the implementation of BNNs, $\mathcal{A}_{j+1}$ will be processed through several layers, \eg, Batch Normalization~(BN) layer, non-linear activation layer, and max-pooling layer.
In this section, we omit these operations for simplicity.

To train a BNN, the forward propagation includes \cref{eq:sign} and \cref{eq:binary_conv},
where their real-value counterparts $\mathcal{W}_j$ and $\mathcal{A}_{j}$ are used for calculating gradients and updating during the backpropagation.
However, the $\mathrm{sign}$ is not differentiable, thereby gradient estimation is important in BNNs.
Following the previous studies, we use straight-through estimator~(STE)~\cite{bengio2013estimating} to approximate the gradient of the loss \wrt the weight $w \in \mathcal{W}$:
\begin{equation}
\frac{\partial  \mathcal{L}}{\partial w}=\frac{\partial  \mathcal{L}}{\partial \hat{w}}\cdot \frac{\partial \hat{w}}{\partial {w}}
\approx \frac{\partial  \mathcal{L}}{\partial \hat{w}},
\label{eq:ste}
\end{equation}
where $ \mathcal{L}$ denotes the loss function.
For the gradient \wrt the activation $a \in \mathcal{A}$, we adopt the piece-wise polynomial gradient estimation function~\cite{liu2018bi} as follows:
\begin{equation}
\frac{\partial  \mathcal{L}}{\partial a}=\frac{\partial  \mathcal{L}}{\partial \hat{a}}\cdot \frac{\partial \hat{a} }{\partial a}\approx \frac{\partial  \mathcal{L}}{\partial \hat{a}}\cdot \frac{\partial F\left( a \right)}{\partial a},
\label{eq:gradient_to_act}
\end{equation}
where
\begin{equation}
\frac{\partial F\left( a \right)}{\partial a}=\left\{ \begin{array}{l}
	2+2a,\\
	2-2a,\\
	0,\\
\end{array}	\begin{array}{l}
	\mathrm{if} -1\le {a}<0,\\
	\mathrm{if}\quad \, 0\le a<1,\\
	\mathrm{otherwise}.\\
\end{array} \right.
\label{eq:gradient_to_act_f}
\end{equation}
\section{Methodology}
\label{sec:method}

In this section, we show that the distribution of gradients and weights for BNNs is independent by systematical and theoretical analysis.
Then, we propose Adaptive Gradient Scaling~(AGS) to scale the gradient and introduce Silence Awareness Decaying~(SAD) to detect ``silent weights'', moving them towards zero.
Both of them can enhance the efficiency of weight sign flipping.

%
\subsection{The Independence of the Gradient and Weight Distribution}
\label{sec:gradient_and_weight}

\begin{figure}[t]
    \centering
    \includegraphics[width=\linewidth]{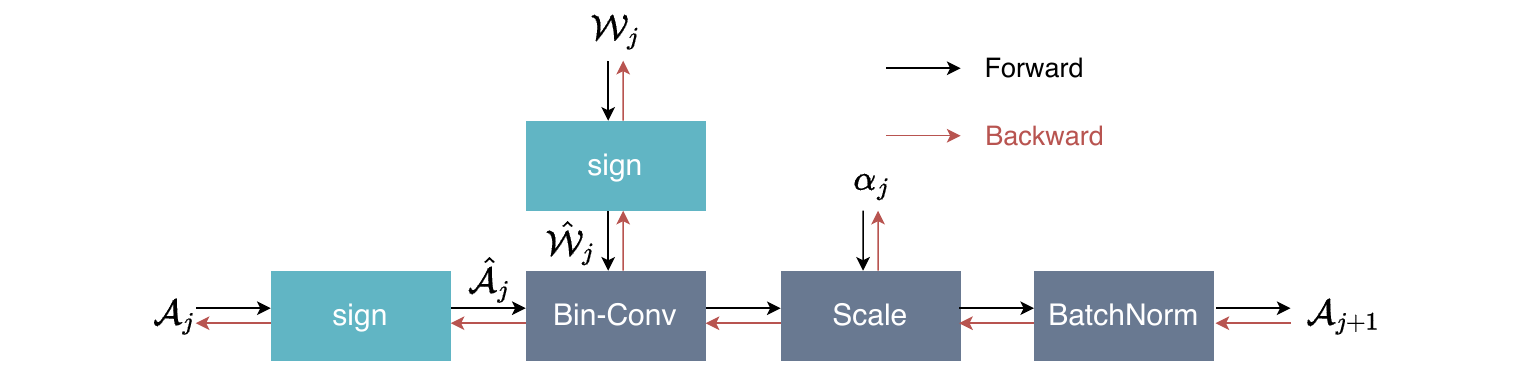}
    \caption{Forward and backward computation graph for binary convolutional operation with quantization aware training.}
    \label{fig:illustration}
    \vspace{-0.2in}
\end{figure}

\Cref{fig:illustration} demonstrates forward and backward computation graph for binary convolutional~(Bin-Conv) operation.
As seen, Bin-Conv is often followed by a BN layer.
We can conclude that once two BNNs $\mathcal{N}$ and $\mathcal{N}^{'}$ satisfy:
\begin{equation}
    \hat{\mathcal{W}}_{j}=\hat{\mathcal{W}}^{'}_{j}, \alpha_{j}=\alpha_j^{'}, \mathrm{BN}_j=\mathrm{BN}^{'}_j, \forall j,
    \label{eq:condition}
\end{equation}
%
for the same input $\hat{\mathcal{A}}_{j}$, the output $\mathcal{A}_{j+1}$ and $\mathcal{A}^{'}_{j+1}$ can be formulated as:
\begin{equation}
    \begin{aligned}
        \mathcal{A}_{j+1} 
        & = \mathrm{BN}\left(\left(\hat{\mathcal{A}}_{j}\circledast \hat{\mathcal{W}}_{j} \right) \odot \alpha_j \right)  
        = \mathrm{BN^{'}}\left(\left(\hat{\mathcal{A}}_{j}\circledast \hat{\mathcal{W}}_{j}^{'} \right) \odot \alpha_j^{'} \right)  = \mathcal{A}_{j+1}^{{'}}. 
    \end{aligned}
\end{equation}
Through mathematical induction, we can know that the $\mathcal{N}$ and $\mathcal{N}^{'}$ will have the same output and loss, \ie $\mathcal{L}=\mathcal{L}^{'}$.
Therefore, the backpropagation for them is:
\begin{equation}
    \begin{aligned}
        \frac{\partial  \mathcal{L}}{\partial \mathcal{W}_{j}} 
        & = \frac{\partial  \mathcal{L}}{\partial \mathcal{A}_{j+1}}
        \frac{\partial  \mathcal{A}_{j+1}}{\partial \hat{\mathcal{W}}_{j}} \frac{\partial \hat{\mathcal{W}}_{j}}{\partial \mathcal{W}_{j}} 
        = \frac{\partial  \mathcal{L}^{'}}{\partial \mathcal{A}_{j+1}^{{'}}}
         \frac{\partial  \mathcal{A}_{j+1}^{{'}}}{\partial \hat{\mathcal{W}}_{j}^{{'}}} \frac{\hat{\mathcal{W}}_{j}^{{'}}}{\partial \mathcal{W}_{j}^{{'}}}
        = \frac{\partial  \mathcal{L}^{'}}{\partial \mathcal{W}_{j}^{{'}}}. 
    \end{aligned}
\end{equation}

Without loss of generality, we can obtain that once two networks satisfy the conditions in \cref{eq:condition}, the gradients obtained from the backpropagation are the same~(a more generalized scenario are extrapolated in \cref{appendix:sec:generalize}).
The gradient has nothing to do with the magnitude of weight and weight with larger magnitude does not necessarily corresponding to larger gradient.
Therefore, although intuitively, weight with large magnitude tends to remain its sign unchanged,
it is also potential for weight with small magnitude to suffer from silence if its gradient is relatively small compared to its magnitude.
Furthermore, if we assume the weights further satisfy $\mathcal{W}_{j}^{{'}}=\gamma \mathcal{W}_{j}$, where $\gamma>1$ and networks are optimized via vanilla SGD,
the optimization of $t$-step can be formulated as:
%
%
\begin{equation}
    \begin{split}
       & \mathcal{W}_{j}^{{'}}\left(t+1\right)=\mathcal{W}_{j}^{{'}}\left(t\right)-\beta(t)\frac{\partial  \mathcal{L^{'}}(t)}{\partial \mathcal{W}_{j}^{{'}}(t)}
    =\gamma\mathcal{W}_{j}\left(t\right)-\beta(t)\frac{\partial  \mathcal{L}(t)}{\partial \mathcal{W}_{j}(t)}
    \\ &=\left(\gamma-1\right)\mathcal{W}_{j}\left(t\right)+\mathcal{W}_{j}\left(t\right)-\beta(t)\frac{\partial  \mathcal{L}(t)}{\partial \mathcal{W}_{j}(t)}
    =\left(\gamma-1\right)\mathcal{W}_{j}\left(t\right)+\mathcal{W}_{j}\left(t+1\right),
    \end{split}
    \label{eq:optimize}
\end{equation}
where $\beta\left(t\right)$ is the learning rate at $t$-step. \Cref{eq:optimize} can be viewed as a variant of exponential moving averages
and the effect of $\mathcal{W}_{j}\left(t\right)$ on $\mathcal{W}_{j}^{'}\left(t+1\right)$ will gradually increases as $\gamma$ increases.
Once $k$ is large enough:
\begin{equation}
\lim_{\gamma \to \infty} \mathcal{W}_{j}^{{'}}\left(t+1\right)
=\lim_{\gamma \to \infty}\left[\left(\gamma-1\right)\mathcal{W}_{j}\left(t\right)+\mathcal{W}_{j}\left(t+1\right)\right]
=\left(\gamma-1\right)\mathcal{W}_{j}\left(t\right).
\end{equation}
The signs of $\mathcal{W}_{j}^{'}\left(t+1\right)$ will be the same as $\mathcal{W}_{j}\left(t\right)$.

However, BNNs binarize $\mathcal{W}_{j}^{{'}}$ to $\{+1,-1\}$ at each training step.
Once the signs of $\mathcal{W}_{j}^{{'}}$ do not change, the BNNs hardly ever update.
It not only slows down convergence, but also leads to significant accuracy degradation.
Please see \cref{appdendix:sec:ablation_analysis_weight_init} for a detailed ablation analysis for $\gamma$.

\subsection{Adaptive Gradient Scaling}
\label{subsec:ags}

To improve the efficiency of the update, an intuitive approach is to scale $\beta\left(t\right)$ to $\gamma\beta\left(t\right)$.
In this case, $\mathcal{W}_{j}^{{'}}\left(t+1\right)$ can be formulated as:
\begin{equation}
    \begin{split}
        \mathcal{W}_{j}^{{'}}\left(t+1\right)&=\mathcal{W}_{j}^{{'}}\left(t\right)-\gamma\beta(t)\frac{\partial  \mathcal{L}(t)}{\partial \mathcal{W}_{j}^{{'}}(t)}
    \\ &=\gamma\mathcal{W}_{j}\left(t\right)-\gamma\beta(t)\frac{\partial  \mathcal{L}(t)}{\partial \mathcal{W}_{j}(t)}=\gamma\mathcal{W}_{j}\left(t+1\right).
\end{split}
\end{equation}
However, selecting the appropriate scaling factor is tricky.
Meanwhile, an inappropriate scaling factor may make weights with small magnitudes correspond to large gradients,
leading them to oscillate frequently in $\{+1,-1\}$ and introducing instability to BNNs training.

To overcome this issue, we introduce ``Adaptive Gradient Scaling''~(AGS).
Let $\mathcal{G}_{j} \in \mathbb{R}^{C^{j}_{\text{out}}\times C_{\text{in}}^{j}\times K^{j}_{\text{h}} \times K^{j}_{\text{w}}}$ denote the gradient with respect to $\mathcal{W}_{j}$,
\ie, $\frac{\partial  \mathcal{L}}{\partial \mathcal{W}_{j}}$,
and $\|\cdot\|_{F}$ denote the Forbenius norm, \ie,
the $k$-th filter norm in $\mathcal{W}_{j}$ can be formulated as:
%
%
\begin{equation}
    \|\mathcal{W}_{j}^{k}\|_F=
    \sqrt{{\textstyle \sum_{l=1}^{C_{\text{in}}^{j}} \sum_{m=1}^{K^{j}_{\text{h}}} \sum_{n=1}^{K^{j}_{\text{w}}}}
    \mathcal{W}_{j}^{k,l,m,n}}.
\end{equation}

The AGS algorithm is motivated by the observation that the ratio of the norm of $\mathcal{G}_{j}^{k}$ to $\mathcal{W}_{j}^{k}$~($\frac{\|\mathcal{G}_{j}^{k}\|_F}{\|\mathcal{W}_{j}^{k}\|_F}$)
provides a simple measure of how much a single gradient descent step will change the original weight $\mathcal{W}_{j}$.
For instance, if we train BNNs via vanilla SGD without momentum, then $\frac{\|\triangle \mathcal{W}_{j}^{k}\|_F}{\|\mathcal{W}_{j}^{k}\|_F}
=\beta\frac{\|\mathcal{G}_{j}^{k}\|_F}{\|\mathcal{W}_{j}^{k}\|_F}$,
where the parameter update for the $\mathcal{W}_{j}^{k}$ is given by $\triangle \mathcal{W}_{j}^{k}=-\beta\mathcal{G}_{j}^{k}$.
We can conclude that the small value of $\frac{\|\mathcal{G}_{j}^{k}\|_F}{\|\mathcal{W}_{j}^{k}\|_F}$ during training is the root cause why a large number of weights fail to flip signs and adaptively scaling $\frac{\|\mathcal{G}_{j}^{k}\|_F}{\|\mathcal{W}_{j}^{k}\|_F}$ plays a crucial role in promoting the update efficiency for BNNs.
Specifically, in AGS algorithm, ${\mathcal{G}}_{j}^{k,l,m,n}$ is scaled as:
\begin{equation}
\overline{\mathcal{G}}_{j}^{k,l,m,n}=
\begin{cases}
    \lambda\frac{\|\mathcal{W}_j^{k}\|_F}{\|{\mathcal{G}}_{j}^{k}\|_F}{\mathcal{G}}_{j}^{k,l,m,n} & \text{if $\frac{\|{\mathcal{G}}_{j}^{k}\|_F}{\|\mathcal{W}_j^{k}\|_F}<\lambda$}, \\
    {\mathcal{G}}_{j}^{k,l,m,n} & \text{otherwise.}
\end{cases}
\label{eq:ags}
\end{equation}
where $\lambda$ is a scalar scaling threshold to limits the lower bound of $\frac{\|{\mathcal{G}}_{j}^{k}\|_F}{\|\mathcal{W}_j^{k}\|_F}$.
Ablation analysis for $\lambda$ can be found in \cref{subsec:ablation}.

AGS and ``Adaptive Gradient Clipping''~(AGC)~\cite{brock2021high} are closely related but fundamentally different,
as the former restricts the lower bound of $\frac{\|{\mathcal{G}}_{j}^{k}\|_F}{\|\mathcal{W}_j^{k}\|_F}$ to enhance the flipping efficiency of BNN weight signs,
while the latter restricts upper bound and is designed to improve training stability.
AGS also can be viewd as a adaptive varient to Layer-wise Adaptive Rate Scaling~(LARS)~\cite{you2017large}, which sets the norm of update parameter to a fixed ratio of the parameter norm,
and completely ignores the gradient magnitude to real-valued networks.
Although LARS is also able to improve the accuracy for BNNs, we find that LARS and AGS is much different in the update of gradient momentum
and doing so degrades performance compared to AGS.
More details for comparing AGS with LARS and ablation analysis for LARS can be found in \cref{appendix:sec:ags_vs_lars}.

\begin{algorithm}[t]
\small
\caption{Overview of the OvSW method.\label{algorithm:overview}}
{\bf Input:} 
A minibatch of inputs and their labels, real-valued weights $\mathcal{W}(t)$, scaling factor $\alpha(t)$,
$\lambda$ for AGS, $\tau$ for SAD, ($\mathcal{S}(t)$, $m$, $\sigma$) for flipping state detection.

{\bf Output:}
Updated $\mathcal{W}(t+1)$, $\alpha(t+1)$ and $\mathcal{S}(t+1)$.
\begin{algorithmic}[1]
	\WHILE{Forward propagation}
        \STATE $\hat{\mathcal{A}}_{j} \gets$ ${\rm sign}({\mathcal{A}}_{j})$.
	\STATE $\hat{\mathcal{W}}_{j} \gets$ ${\rm sign}({\mathcal{W}}_{j})$.
	\STATE Computing features via \cref{eq:sign} and \cref{eq:binary_conv}.
        \STATE Computing loss $\mathcal{L}$.
	\ENDWHILE
	\WHILE{Backward propagation}
	\STATE Computing $\frac{\partial \mathcal{L}}{\partial \mathcal{W}_j}$, \ie $\mathcal{G}_{j}$, and $\frac{\partial \mathcal{L}}{\partial \mathcal{A}_j}$ via \cref{eq:ste} and \cref{eq:gradient_to_act}\&\cref{eq:gradient_to_act_f}
        \STATE Scaling ${\mathcal{G}}_{j}$ to $\overline{\mathcal{G}}_{j}$ via \cref{eq:ags}.
        \STATE Adding penalties to $\overline{\mathcal{G}}_{j}$ via \cref{eq:sad}.
        \STATE Update $\mathcal{W}_{j}(t+1)$ and $\alpha_j(t+1)$ via SGD optimizer.
        \STATE Update $\mathcal{S}_{j}(t+1)$ via \cref{eq:ema}.
	\ENDWHILE
\end{algorithmic}
\end{algorithm}

\subsection{Silence Awareness Decaying}
\label{subsec:sad}

We propose another approach~(Silence Awareness Decaying, SAD) orthogonal to AGS to detect and prevent ``silent weights''.
Specifically, we track the flipping state of $\mathcal{W}_j$ over time using an exponential moving average~(EMA) strategy,
which is formulated as:
\begin{equation}
\mathcal{S}_j(t)=m\cdot\mathcal{S}_j(t-1)+(1-m)\cdot\frac{|\mathrm{sign}\left(\mathcal{W}_j\left(t\right )\right)
-\mathrm{sign}\left(\mathcal{W}_j\left(t-1\right )\right)|_{abs}}{2},
\label{eq:ema}
\end{equation}
where $m$ is the momentum and $\mathcal{S}_j$ is the auxiliary variable.
Nagel~\etal~\cite{nagel2022overcoming} employ this technique to identify and dampen oscillations prematurely
while we employ it to identify ``silent weights'' and dynamically introduce additional weight penalties to move them towards zero.
In our algorithm, we think if $\mathcal{S}_{j}^{k,l,m,n}$ is less than a pre-defined threshold $\sigma$,
its corresponding weight $\mathcal{W}_{j}^{k,l,m,n}$ is considered as a ``silent weight''
and will be applied with an additional penalty.
The silence awareness decaying process is formulated as:
\begin{equation}
    {\overline{\overline{\mathcal{G}}}_{j}^{k,l,m,n}(t)}=\begin{cases}
    \overline{\mathcal{G}}_{j}^{k,l,m,n}(t)+\gamma \mathcal{W}_{j}^{k,l,m,n}(t),  & \text{if}~\mathcal{S}_{j}^{k,l,m,n}(t)<\sigma, \\
    \overline{\mathcal{G}}_{j}^{k,l,m,n}(t),  & \text{otherwise}, \\
    \end{cases}
    \label{eq:sad}
\end{equation}
where $\gamma$ is the proportion of penalty term.
%

%
It is worth noting that while both AGS and SAD improve the efficiency of weight sign flipping, they solve the problem in fundamentally different ways.
AGS facilitates sign flipping for the whole weights by adaptively scaling the gradient,
while SAD detects ``silent weights'' by tracking their flipping state and applies additional penalties.
In \cref{subsec:ablation}, we show that both methods achieve significant performance improvement and they are complementary to each other.
As an algorithm guideline, the pseudo-code of OvSW is provided in \cref{algorithm:overview}.
\section{Experiment}
\label{sec:experiment}

\begin{table}[t]
\caption{Performance comparison with SOTAs on CIFAR10.
We report the Top-1 Accuracy performance on ResNet18, ResNet20, and VGGsmall. W/A denotes the bit-width of weights/activations.}
\vspace{-0.2in}
\setlength{\tabcolsep}{0.5em}
	\begin{center}
		\begin{tabular}{clccc}
		      \toprule
			Networks                      & Methods                             & Bit-width (W/A)   & Top-1 Acc.(\%)       \\
   		  \midrule
			\multirow{7}{*}{ResNet18}     & Full-precision                      & 32/32             & 94.8           \\ 
                                              & IR-Net~\cite{qin2020forward}        & 1/1               & 91.5           \\ 
			                               & RBNN~\cite{lin2020rotated}          & 1/1               & 92.2           \\ 
                                	       & CMIM~\cite{shang2022network}        & 1/1               & 92.2           \\ 
                                		     & SiMaN~\cite{lin2022siman}           & 1/1               & 92.5           \\ 
                                	       & ReCU~\cite{xu2021recu}              & 1/1               & 92.8           \\ 
			                                 & \textbf{OvSW~(Ours)}                & 1/1               & \textbf{93.2}  \\
                \midrule
                \multirow{7}{*}{ResNet20}     & Full-precision                      & 32/32             & 92.1           \\ 
                                              & SLB~\cite{yang2020searching}        & 1/1               & 85.5           \\ 
                                              & FDA-BNN~\cite{xu2021learning}       & 1/1               & 86.2           \\
                                              & IR-Net~\cite{qin2020forward}        & 1/1               & 86.5           \\ 
                                	       & CMIM~\cite{shang2022network}        & 1/1               & 87.3           \\ 
                                		     & SiMaN~\cite{lin2022siman}           & 1/1               & 87.4           \\ 
                                	       & ReCU~\cite{xu2021recu}              & 1/1               & 87.4           \\ 
			                                 & \textbf{OvSW~(Ours)}                & 1/1              & \textbf{87.7}  \\
                \midrule
			\multirow{11}{*}{VGGsmall}    & Full-precision                      & 32/32             & 94.1           \\ 
                                              & DoReFa~\cite{zhou2016dorefa}        & 1/1               & 90.2           \\
                                		   & RAD~\cite{ding2019regularizing}     & 1/1               & 90.5           \\ 
                                		   & RBNN~\cite{lin2020rotated}          & 1/1               & 91.3           \\ 
                                	       & DSQ~\cite{gong2019differentiable}   & 1/1               & 91.7           \\ 
                                              & Proxy-BNN~\cite{he2020proxybnn}     & 1/1               & 91.8           \\ 
                                		   & SLB~\cite{yang2020searching}        & 1/1               & 92.0           \\ 
                                		   & ReCU~\cite{xu2021recu}              & 1/1               & 92.2           \\ 
                                              & FDA-BNN~\cite{xu2021learning}       & 1/1               & 92.5           \\
                                	       & SiMaN~\cite{lin2022siman}           & 1/1               & 92.5           \\
                                		   & \textbf{OvSW~(Ours)}                & 1/1               & \textbf{92.8}  \\
                \bottomrule
            \end{tabular}
            \label{tab:cifar10}
	\end{center}
\vspace{-0.4in}
\end{table}

In this section, we conduct extensive image classification experiments for OvSW and compare it to state-of-the-art (SOTA) methods on CIFAR10 and ImageNet1K
with various architectures.
Then, we discuss the hyperparameter settings for OvSW, including $\lambda$ for AGS and $\sigma$ for SAD and convergence speed on CIFAR100.
We also conduct ablation study to demonstrate the compatibility and visualize the loss landscape for OvSW.
Finally, we deploy OvSW to a real-world mobile device to exhibit its efficiency.
To train OvSW, we use one NVIDIA RTX 3090 on the CIFAR10 and CIFAR100 and four on the ImageNet1K.
All experiments are implemented on PyTorch~\cite{Paszke_PyTorch_An_Imperative_2019}.

\begin{table*}[t]
\caption{Performance comparison with SOTAs on ImageNet1K.
We report the Top-1 and Top-5 Accuracy performance on ResNet18 and ResNet34. W/A denotes the bit-width of weights/activations. $^*$ means using the two-step training setting as ReActNet.}
        \vspace{-0.2in}
        \setlength{\tabcolsep}{0.5em}
	\begin{center}
		\begin{tabular}{clccccc}
			\toprule
    		Model                           & Method                              & {Bit-width (W/A)} & Top-1 Acc.(\%)   & Top-5 Acc.(\%) \\
			\midrule
			\multirow{12}{*}{ResNet18}      & Full-precision                      & 32/32             & 69.6             & 89.2 \\  \cmidrule(l){2-5}
                                                & XNOR~\cite{rastegari2016xnor}       & 1/1               & 51.2             & 73.2 \\   
                                			& BiReal~\cite{liu2018bi}             & 1/1               & 56.4             & 79.5 \\
                                			& IR-Net~\cite{qin2020forward}        & 1/1               & 58.1             & 80.0 \\
                                			& RBNN~\cite{lin2020rotated}          & 1/1               & 59.9             & 81.9 \\
                                			& SiMaN~\cite{lin2022siman}           & 1/1               & 60.1             & 82.3 \\
                                			& FDA-BNN~\cite{xu2021learning}       & 1/1               & 60.2             & 82.3 \\
                                			& ReCU~\cite{xu2021recu}              & 1/1               & 61.0             & 82.6 \\
                                			& \textbf{OvSW~(Ours)}                & 1/1               & \textbf{61.6}    & \textbf{83.1} \\  \cmidrule(l){2-5}
                                                & ReActNet~\cite{liu2020reactnet}     & 1/1               & 65.9             & 86.1 \\
                                                & ReCU~\cite{xu2021recu}              & 1/1               & 66.4             & 86.5 \\
                                                & \textbf{OvSW$^*$~(Ours)}            & 1/1               & \textbf{66.6}    & \textbf{86.7} \\
                \midrule
			\multirow{9}{*}{ResNet34}       & Full-precision                      & 32/32             & 73.3             & 91.3 \\ \cmidrule(l){2-5}
                                			& XNOR++~\cite{bulat2019xnor}         & 1/1               & 57.1             & 79.9 \\ 
                                			& BiReal~\cite{liu2018bi}             & 1/1               & 62.2             & 83.9 \\
                                			& IR-Net~\cite{qin2020forward}        & 1/1               & 62.9             & 84.1 \\
                                			& RBNN~\cite{lin2020rotated}          & 1/1               & 63.1             & 84.4 \\
                                			& SiMaN~\cite{lin2022siman}           & 1/1               & 63.9             & 84.8 \\
                                			& CMIM~\cite{shang2022network}        & 1/1               & 65.0             & 85.7 \\
                                			& ReCU~\cite{xu2021recu}              & 1/1               & 65.1             & 85.8 \\
                                			& \textbf{OvSW~(Ours)}                & 1/1               & \textbf{65.5}    & \textbf{86.1} \\
                \bottomrule
		\end{tabular}
        \label{tab:imagenet}
	\end{center}
 \vspace{-0.4in}
\end{table*}

\subsection{Results on CIFAR10}
We trained OvSW for CIFAR10 with 600 epochs, where the batch size is set to 256 and the initial learning rate to 0.1, decaying with CosineAnealing.
We adopt SGD optimizer with a momentum of 0.9 and weight decay of 5e-4 and employ the same data augmentation in ReCU~\cite{xu2021recu}.
$\lambda$ and $\sigma$ are set to 0.04 and 9e-4 respectively.
We compare OvSW with IR-Net~\cite{qin2020forward}, RBNN~\cite{lin2020rotated}, CMIM~\cite{shang2022network}, SiMaN~\cite{lin2022siman}, ReCU~\cite{xu2021recu},
SLB~\cite{yang2020searching}, FDA-BNN~\cite{xu2021learning}, DoReFa~\cite{zhou2016dorefa},
RAD~\cite{ding2019regularizing}, DSQ~\cite{gong2019differentiable}, and Proxy-BNN~\cite{he2020proxybnn}.
As shown in \cref{tab:cifar10}, OvSW achieves the best performance among all methods.
For ResNet18, OvSW obtains 93.2\% top-1 accuracy, which outperforms SiMaN and ReCU by 0.7\% and 0.4\% respectively, reducing the accuracy gap between BNNs and full-precision model to 1.6\%.
In addition, it yields 87.7\% and 92.8\% top-1 accuracy on ResNet20 and VGGsmall, which succeeds ReCU and SiMaN by 0.3\% and 0.3\% respectively.


\subsection{Results on ImageNet1K}
On ImageNet1K, OvSW is trained from scratch.
We train OvSW with 200 epochs, where the batch size is set to 512 and the initial learning rate is set to 0.1, decaying with CosineAnnealing.
We adopt SGD optimizer with a momentum of 0.9 and weight decay of 1e-4 and the data augmentation is the same as ReCU~\cite{xu2021recu}.
%
$\lambda$ and $\sigma$ are set to 0.02 and 2e-5 respectively.
We demonstrate the ImageNet1K performance of ResNet18/34 and compare OvSW with SOTA methods,
including one-stage training methods XNOR~\cite{rastegari2016xnor}, BiReal~\cite{liu2018bi}, IR-Net~\cite{qin2020forward}, RBNN~\cite{lin2020rotated}, SiMaN~\cite{lin2022siman}, FDA-BNN~\cite{xu2021learning}, ReCU~\cite{xu2021recu},
CMIM~\cite{shang2022network}, and two-stage training method~\cite{martinez2020training} adopted by ReActNet~\cite{liu2020reactnet}.
As shown in \cref{tab:imagenet}, OvSW also achieves the best performance.
For ResNet18, OvSW achieves 61.6\% top-1 and 83.1\% top-5 accuracy compared to ReCU's 61.0\% top-1 and 82.6\% top-5 accuracy,
which demonstrates the efficiency of overcoming the ``silent weights''.
For ResNet34, OvSW achieves 65.5\% top-1 accuracy, which is also better than ReCU.
We further compare OvSW with the two-stage training method ReActNet.
OvSW obtains 66.6\% top-1 accuracy, succeeding ReActNet and ReCU by 0.7\% and 0.2\% respectively.

\subsection{Ablation Analysis}
\label{subsec:ablation}

We investigate the effectiveness of hyper-parameters, including $\lambda$ and $\sigma$, convergence speed, different components, and compatibility through ablation analysis.
All the following results are based on binarized ResNet18 for CIFAR100.
%

\begin{figure*}[t]
    \centering
    \begin{subtable}[h]{0.32\textwidth}
        \includegraphics[width=1.\linewidth]{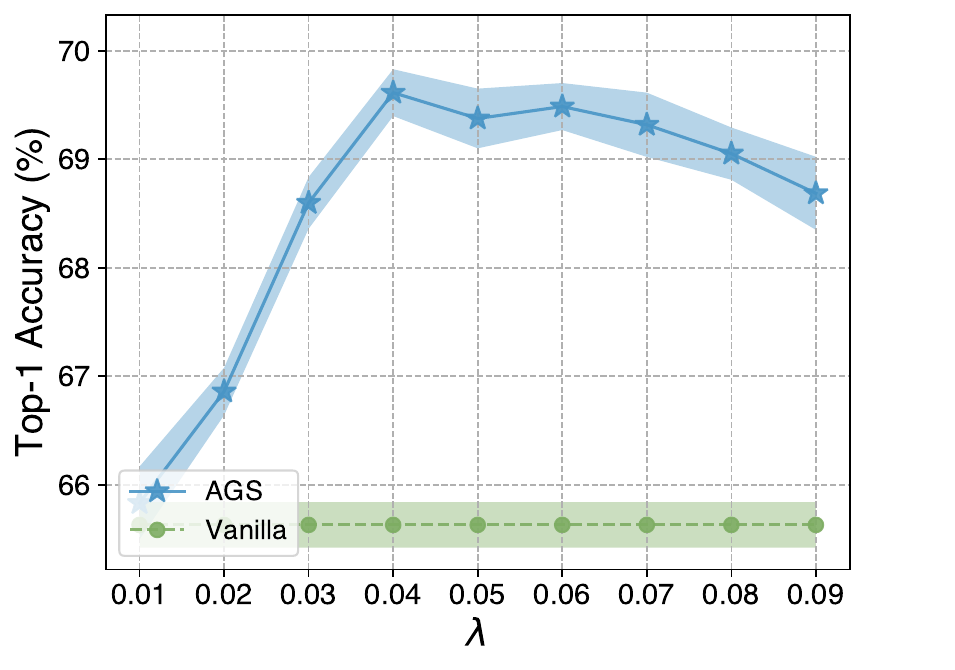}
        \caption{AGS vs Vanilla}
        \label{fig:ags_vs_vanilla}	
    \end{subtable}
    \begin{subtable}[h]{0.32\textwidth}
        \includegraphics[width=1\linewidth]{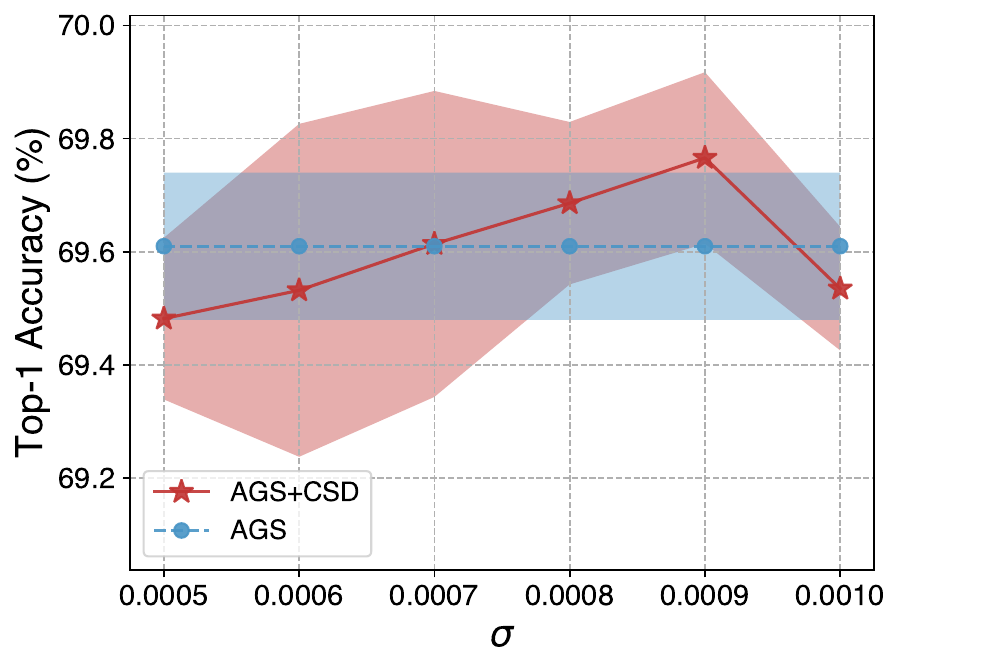}
        \caption{AGS+SAD\,(OvSW) vs AGS}
        \label{fig:ags_csd_vs_ags}	
    \end{subtable}
    \begin{subtable}[h]{0.32\textwidth}
        \includegraphics[width=1\linewidth]{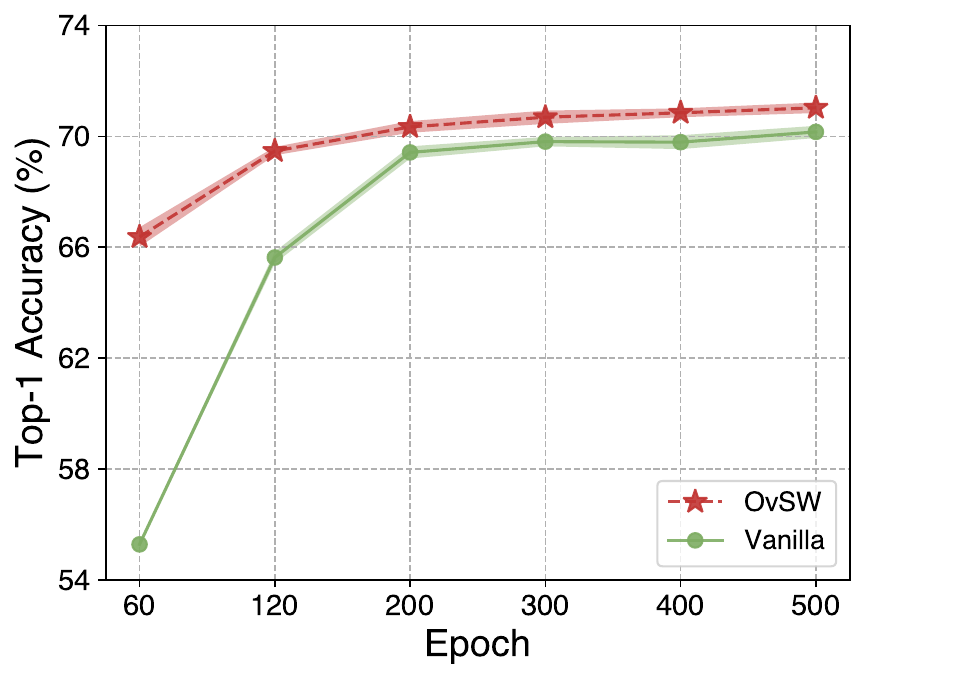}
        \caption{Convergence}
        \label{fig:ablation_convergence}
    \end{subtable}
    \vspace{-0.1in}
    \caption{Top-1 accuracy~(mean$\pm$std) of binarized ResNet18 \wrt different values of $\lambda$~(a), $\sigma$~(b) and epoch~(c) on CIFAR100.}
\vspace{-0.2in}
\end{figure*}

%
\noindent \textbf{$\bm{\lambda}$ for AGS and $\bm{\sigma}$ for SAD.}
We first compare AGS with vanilla BNNs and analyze the $\lambda$ for AGS.
In \cref{fig:ags_vs_vanilla}, we present the results for nine different settings of $\lambda$,
which vary from 0.01 to 0.09.
As seen, the appropriate $\lambda$ can significantly improve the performance of BNNs, and $\lambda=0.04$ achieves the best performance.
%
%
If $\lambda$ is too small, the sign of the weights can flip inefficiently;
while $\lambda$ is too large, the sign of the weights can flip dramatically, introducing instability to the training.
Based on this result, we further introduce SAD to AGS~($\lambda=0.04$) to analyze $\sigma$.
\Cref{fig:ags_csd_vs_ags} demonstrates that the performance of the model can be further improved by identifying ``silent weights'' and applying additional penalties to them via SAD~($\sigma=0.0009$).

\noindent \textbf{Convergence for OvSW.}
To verify that OvSW facilitates the flipping of weight signs and thus improves the efficiency of convergence,
we fix the $\lambda$ and $\sigma$ to 0.04 and 0.0009 respectively, and record the top-1 accuracy over the training epoch from 60 to 500.
As shown in \cref{fig:ablation_convergence}, OvSW effectively accelerates training convergence and achieves better performance.
For example, OvSW achieves 66.37$\pm$0.35\% top-1 accuracy with only 60 traning epochs, while vanilla BNNs only reaches 55.28$\pm$0.07\% and 65.23$\pm$0.21\% top-1 accuracy with 60 and 120 training epochs respectively.
It indicates that OvSW can achieve significant performance gains in scenarios with limited training resources, such as training the network on edge devices.
Meanwhile, although increasing epochs improve the final performance for both OvSW and vanilla BNNs, OvSW consistently outperforms the vanilla BNNs.

\begin{table}[h]
\caption{Left: Ablation study of different components in OvSW. Right: Applying OvSW as a plug-and-play module to other methods.}
\centering
\begingroup
\setlength{\tabcolsep}{3mm}
\begin{tabular}{ccc}
        \toprule
        AGS                         & SAD                   &  mean$\pm$std~(\%)\\ \midrule
        \ding{55}                   & \ding{55}             &  65.23$\pm$0.21   \\
        \ding{51}                   & \ding{55}             &  69.61$\pm$0.22   \\
        \ding{55}                   & \ding{51}             &  69.45$\pm$0.18   \\ \midrule
        \ding{51}                   & \ding{51}             &  {69.77$\pm$0.15}   \\ 
        \bottomrule
\end{tabular}
\endgroup
\hspace{0.10in}
\begingroup
\setlength{\tabcolsep}{3mm}
\begin{tabular}{lc}
\toprule
Method      & mean$\pm$std~(\%)      \\ \midrule
AdaBin      & 70.56$\pm$0.19         \\
AdaBin+OvSW & 72.56$\pm$0.11         \\ 
\midrule
RBNN        & 67.15$\pm$0.23         \\
RBNN+OvSW   & 69.98$\pm$0.13         \\
\bottomrule
\end{tabular}
\label{tab:components}
\endgroup
\vspace{-0.2in}
\end{table}

\noindent \textbf{Components.}
AGS and SAD promote sign flipping for overall weights and ``silent weights'' respectively.
To prove that they are orthogonal to each other, we conduct components ablation study for different modules and show the results in \cref{tab:components}~(left).
As seen, AGS and SAD achieve 69.61$\pm$0.22\% and 69.45$\pm$0.18\% top-1 accuracy respectively and both of them succeed vanilla BNNs.
It shows that they are effective in improving the efficiency of weight signs flipping.
By combining them, OvSW~(AGS+SAD) further achieves the best top-1 accuracy with 69.77$\pm$0.15\%.
%

%
\noindent \textbf{Compatibility.}
We show the good compatibility of OvSW by inserting it as a plug-and-play module into the current state-of-the-art methods,
including AdaBin~\cite{tu2022adabin} and RBNN~\cite{lin2020rotated}.
The former introduces an adaptive binary set to enhance the feature representation of BNNs,
while the latter proposes a training-aware approximation function to reduce the gradient estimation error during training.
\Cref{tab:components}~(right) demonstrates OvSW achieves 72.56$\pm$0.11\% top-1 accuracy on AdaBin and 69.98$\pm$0.13\% on RBNN,
succeeding their original performance.
This result indicates that OvSW has good compatibility and can effectively enhance the performance of existing methods.

\begin{figure*}[h]
    \centering
    \begin{subtable}[t]{0.32\textwidth}
        \includegraphics[width=1\linewidth]{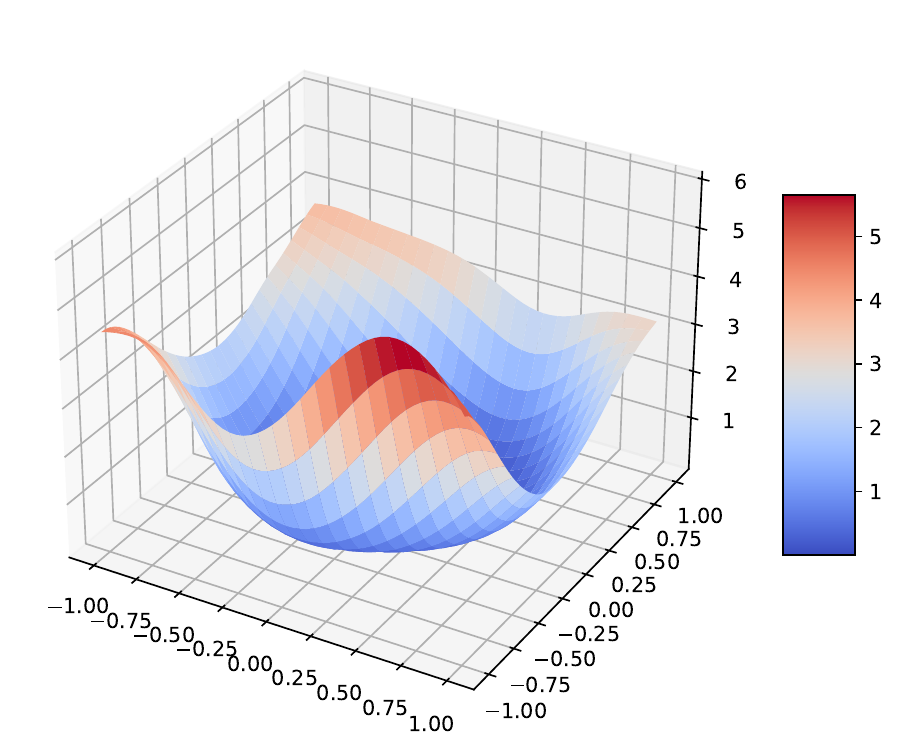}
        \caption{Full-precision}
        \label{fig:full_precision}	
    \end{subtable}
    \begin{subtable}[t]{0.32\textwidth}
        \includegraphics[width=1\linewidth]{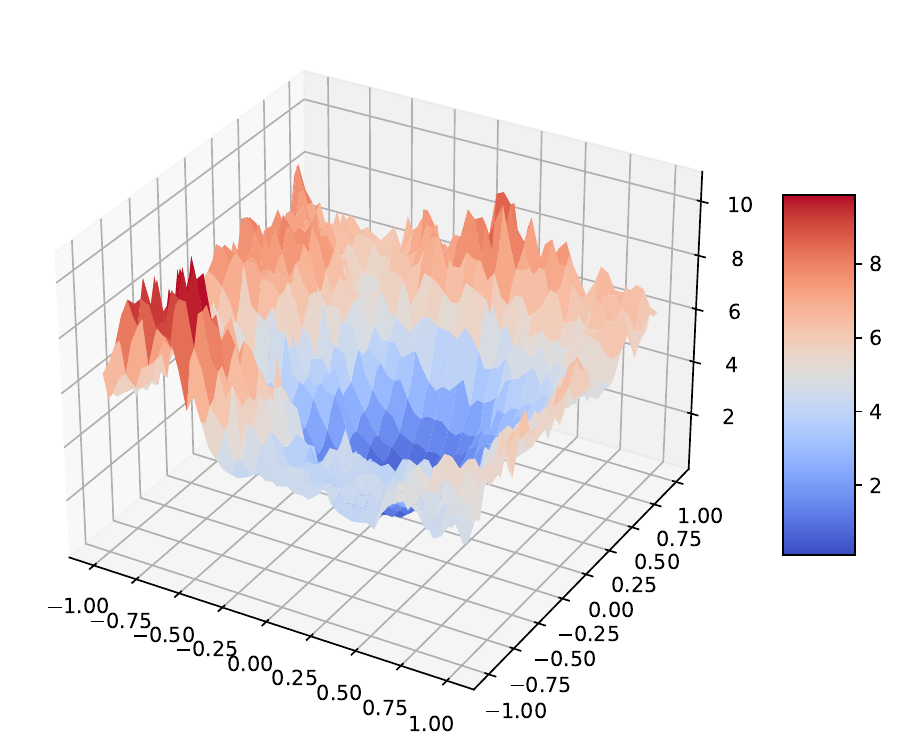}
        \caption{XNOR-Net++}
        \label{fig:xnor18}	
    \end{subtable}
    \begin{subtable}[t]{0.32\textwidth}
        \includegraphics[width=1\linewidth]{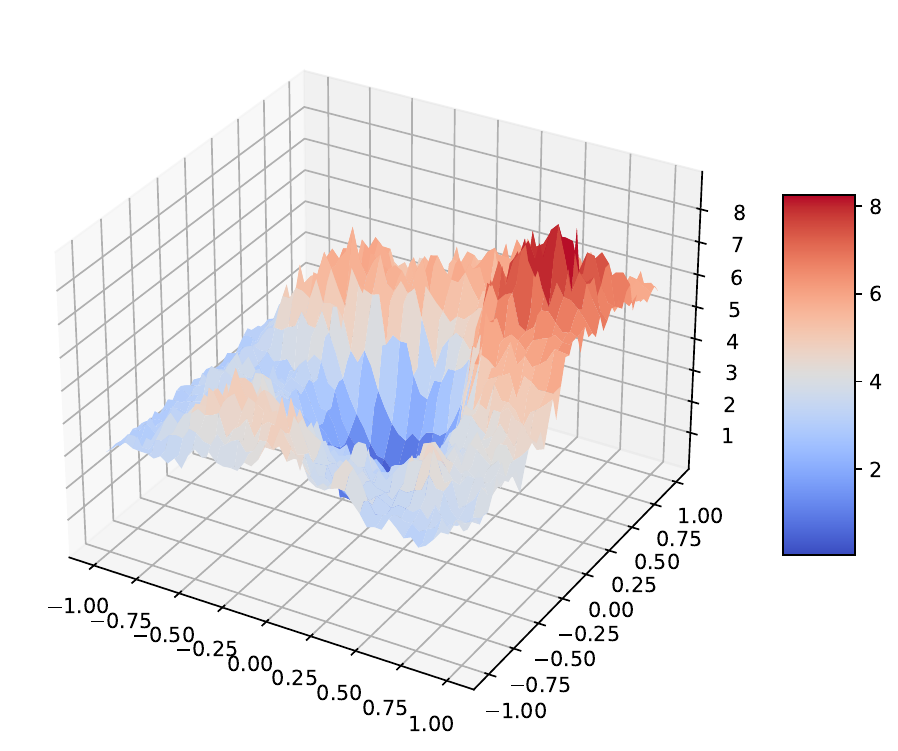}
        \caption{OvSW}
        \label{fig:ovsw}	
    \end{subtable}
    \vspace{-0.1in}
    \caption{3D visualization of the loss surfaces of ResNet18 on CIFAR100, which is used to enable comparisons of sharpness/flatness of different methods.\label{fig:loss_landscape}}
    \vspace{-0.4in}
\end{figure*}

\subsection{Loss Landscape Visualization}
BNNs restrict the weights and activations to discrete values, which naturally limits the representational capacity of the model and further result in disparate optimization landscapes compared to real-valued ones \cite{liu2021adam}.
As illustrated in \cref{fig:loss_landscape}, we follow the method in \cite{visualloss} to plot the actual optimization landscape about our OvSW and compare it with the same architecture to real-valued and XNOR-Net++. 
As seen, our OvSW has a significantly smoother loss-landscape and minor loss elevation compared to XNOR-Net++, which confirms the effectiveness of OvSW in the BNN optimization.

\subsection{Deployment Efficiency}
We implement 1-bit models on the M1 Pro, which features 8 high-performance Firestorm cores and 2 efficient Icestorm cores in a hybrid design.
The Firestorm cores can be clocked up to 3.2 GHz, while the Icestorm cores can reach 2.1 GHz.
Our OvSW shows significant efficiency gains when deployed on real-world mobile devices, as evidenced by practical speed evaluations.
To make our inference framework BOLT~\cite{feng2021bolt} compatible with OvSW, we leverage the ARM NEON SIMD instruction SSHL. 
We compare OvSW with 32-bit and 16-bit backbones.
As shown in \cref{tab:deploy}, OvSW inference speed is substantially faster with the highly efficient BOLT framework in a single thread. 
For instance, OvSW achieves an acceleration rate of about 3.47$\times$ on ResNet18 compared to its 32-bit counterpart. 
For the ResNet34 backbone,
OvSW achieves a 3.42$\times$ acceleration rate with the BOLT framework on hardware, which is significant for computer vision applications on real-world edge devices.
At the same time, OvSW can save memory by a factor of 16.64$\times$ and 21.16$\times$, which demonstrates its potential for applications with limited memory resources.

\begin{table}[t]
\caption{Comparing OvSW~(1-bit) with 32-bit and 16-bit backbones on M1 Pro. }
\centering
\setlength{\tabcolsep}{2.2mm}
\begin{tabular}{@{}lccccc@{}}
\toprule
Network                   & W/A     & Size~(MB)   & Memory  Saving & Latency~(ms)   & Acceleration \\ 
\midrule
\multirow{3}{*}{ResNet18} & 32/32   & 46.76       & -              & 27.67          & -            \\
                          & 16/16   & 23.38       & 2$\times$      & 16.65          & 1.66$\times$ \\
                          & 1/1     & 2.81        & 16.64$\times$  & 7.97           & 3.47$\times$ \\ 
\midrule
\multirow{3}{*}{ResNet34} & 32/32   & 87.19       & -              & 48.82          & -            \\
                          & 16/16   & 43.60       & 2$\times$      & 29.01          & 1.68$\times$ \\
                          & 1/1     & 4.12        & 21.16$\times$  & 14.28          & 3.42$\times$ \\ 
\bottomrule
\label{tab:deploy}
\end{tabular}
\vspace{-0.4in}
\end{table}
\section{Conclusion}
\label{sec:conclusion}
BNN is a crucial method to compress deep learning models and reduce inference overhead.
In this paper, systematically and theoretically, we prove the distribution of gradients is independent of latent weights,
which is the root cause of inefficient updating and performance degradation of BNNs.
To this end, Adaptive Gradient Scaling~(AGS) and Silence Awareness Decaying~(SAD), are proposed to Overcome Silent Weights~(OvSW) and achieve SOTA performance on BNNs.
Specifically, AGS adaptively scales the gradient based on the distribution of weights, improving the efficiency of sign flipping for the overall weights.
SAD can effectively measure the states of weight flipping, detect the ``silent weights'' and introduce additional penalty to them to facilitate their flipping.
Extensive experiments demonstrate that OvSW can achieve notable performance gains over the SOTA methods on various datasets and networks.
In addition, OvSW has better convergence efficiency and excellent compatibility, which can be combined with existing methods to further enhance the performance of BNNs.



%
%
\bibliographystyle{splncs04}

\begin{thebibliography}{}
\providecommand{\url}[1]{\texttt{#1}}
\providecommand{\urlprefix}{URL }
\providecommand{\doi}[1]{https://doi.org/#1}

\end{thebibliography}


\begin{thebibliography}{10}
\providecommand{\url}[1]{\texttt{#1}}
\providecommand{\urlprefix}{URL }
\providecommand{\doi}[1]{https://doi.org/#1}

\bibitem{bengio2013estimating}
Bengio, Y., L{\'e}onard, N., Courville, A.: Estimating or propagating gradients
  through stochastic neurons for conditional computation. arXiv preprint
  arXiv:1308.3432  (2013)

\bibitem{brock2021high}
Brock, A., De, S., Smith, S.L., Simonyan, K.: High-performance large-scale
  image recognition without normalization. In: Int. Conf. Mach. Learn. pp.
  1059--1071. PMLR (2021)

\bibitem{bulat2019xnor}
Bulat, A., Tzimiropoulos, G.: Xnor-net++: Improved binary neural networks.
  arXiv preprint arXiv:1909.13863  (2019)

\bibitem{chen2020learning}
Chen, J., Liu, L., Liu, Y., Zeng, X.: A learning framework for n-bit quantized
  neural networks toward fpgas. IEEE Transactions on Neural Networks and
  Learning Systems  \textbf{32}(3),  1067--1081 (2020)

\bibitem{courbariaux2016binarized}
Courbariaux, M., Hubara, I., Soudry, D., El-Yaniv, R., Bengio, Y.: Binarized
  neural networks: Training deep neural networks with weights and activations
  constrained to+ 1 or-1. arXiv preprint arXiv:1602.02830  (2016)

\bibitem{ding2019regularizing}
Ding, R., Chin, T.W., Liu, Z., Marculescu, D.: Regularizing activation
  distribution for training binarized deep networks. In: IEEE Conf. Comput.
  Vis. Pattern Recog. pp. 11408--11417 (2019)

\bibitem{ding2021repvgg}
Ding, X., Zhang, X., Ma, N., Han, J., Ding, G., Sun, J.: Repvgg: Making
  vgg-style convnets great again. In: IEEE Conf. Comput. Vis. Pattern Recog.
  pp. 13733--13742 (2021)

\bibitem{duan2021transnas}
Duan, Y., Chen, X., Xu, H., Chen, Z., Liang, X., Zhang, T., Li, Z.:
  Transnas-bench-101: Improving transferability and generalizability of
  cross-task neural architecture search. In: IEEE Conf. Comput. Vis. Pattern
  Recog. pp. 5251--5260 (2021)

\bibitem{feng2021bolt}
Feng, J.: Bolt. \url{https://github.com/huawei-noah/bolt} (2021)

\bibitem{girshick2014rich}
Girshick, R., Donahue, J., Darrell, T., Malik, J.: Rich feature hierarchies for
  accurate object detection and semantic segmentation. In: IEEE Conf. Comput.
  Vis. Pattern Recog. pp. 580--587 (2014)

\bibitem{gong2019differentiable}
Gong, R., Liu, X., Jiang, S., Li, T., Hu, P., Lin, J., Yu, F., Yan, J.:
  Differentiable soft quantization: Bridging full-precision and low-bit neural
  networks. In: Int. Conf. Comput. Vis. pp. 4852--4861 (2019)

\bibitem{he2017mask}
He, K., Gkioxari, G., Doll{\'a}r, P., Girshick, R.: Mask r-cnn. In: Int. Conf.
  Comput. Vis. pp. 2961--2969 (2017)

\bibitem{he2015delving}
He, K., Zhang, X., Ren, S., Sun, J.: Delving deep into rectifiers: Surpassing
  human-level performance on imagenet classification. In: Int. Conf. Comput.
  Vis. pp. 1026--1034 (2015)

\bibitem{he2015spatial}
He, K., Zhang, X., Ren, S., Sun, J.: Spatial pyramid pooling in deep
  convolutional networks for visual recognition. IEEE Trans. Pattern Anal.
  Mach. Intell.  \textbf{37}(9),  1904--1916 (2015)

\bibitem{he2016deep}
He, K., Zhang, X., Ren, S., Sun, J.: Deep residual learning for image
  recognition. In: IEEE Conf. Comput. Vis. Pattern Recog. pp. 770--778 (2016)

\bibitem{he2020proxybnn}
He, X., Mo, Z., Cheng, K., Xu, W., Hu, Q., Wang, P., Liu, Q., Cheng, J.:
  Proxybnn: Learning binarized neural networks via proxy matrices. In: Eur.
  Conf. Comput. Vis. pp. 223--241. Springer (2020)

\bibitem{helwegen2019latent}
Helwegen, K., Widdicombe, J., Geiger, L., Liu, Z., Cheng, K.T., Nusselder, R.:
  Latent weights do not exist: Rethinking binarized neural network
  optimization. Adv. Neural Inform. Process. Syst.  \textbf{32} (2019)

\bibitem{hinton2015distilling}
Hinton, G., Vinyals, O., Dean, J., et~al.: Distilling the knowledge in a neural
  network. arXiv preprint arXiv:1503.02531  \textbf{2}(7) (2015)

\bibitem{horowitz20141}
Horowitz, M.: 1.1 computing's energy problem (and what we can do about it). In:
  2014 IEEE international solid-state circuits conference digest of technical
  papers (ISSCC). pp. 10--14. IEEE (2014)

\bibitem{jacob2018quantization}
Jacob, B., Kligys, S., Chen, B., Zhu, M., Tang, M., Howard, A., Adam, H.,
  Kalenichenko, D.: Quantization and training of neural networks for efficient
  integer-arithmetic-only inference. In: IEEE Conf. Comput. Vis. Pattern Recog.
  pp. 2704--2713 (2018)

\bibitem{kingma2014adam}
Kingma, D.P., Ba, J.: Adam: A method for stochastic optimization. arXiv
  preprint arXiv:1412.6980  (2014)

\bibitem{krizhevsky2009learning}
Krizhevsky, A., Hinton, G., et~al.: Learning multiple layers of features from
  tiny images  (2009)

\bibitem{krizhevsky2012imagenet}
Krizhevsky, A., Sutskever, I., Hinton, G.E.: Imagenet classification with deep
  convolutional neural networks. Adv. Neural Inform. Process. Syst.
  \textbf{25} (2012)

\bibitem{lee2023insta}
Lee, C., Kim, H., Park, E., Kim, J.J.: Insta-bnn: Binary neural network with
  instance-aware threshold. In: Int. Conf. Comput. Vis. pp. 17325--17334 (2023)

\bibitem{lee2021network}
Lee, J., Kim, D., Ham, B.: Network quantization with element-wise gradient
  scaling. In: IEEE Conf. Comput. Vis. Pattern Recog. pp. 6448--6457 (2021)

\bibitem{visualloss}
Li, H., Xu, Z., Taylor, G., Studer, C., Goldstein, T.: Visualizing the loss
  landscape of neural nets. In: Adv. Neural Inform. Process. Syst. (2018)

\bibitem{li2022learning}
Li, S., Lin, M., Wang, Y., Fei, C., Shao, L., Ji, R.: Learning efficient gans
  for image translation via differentiable masks and co-attention distillation.
  IEEE Trans. Multimedia  (2022)

\bibitem{lin2020hrank}
Lin, M., Ji, R., Wang, Y., Zhang, Y., Zhang, B., Tian, Y., Shao, L.: Hrank:
  Filter pruning using high-rank feature map. In: IEEE Conf. Comput. Vis.
  Pattern Recog. pp. 1529--1538 (2020)

\bibitem{lin2022siman}
Lin, M., Ji, R., Xu, Z., Zhang, B., Chao, F., Lin, C.W., Shao, L.: Siman:
  Sign-to-magnitude network binarization. IEEE Trans. Pattern Anal. Mach.
  Intell.  (2022)

\bibitem{lin2020rotated}
Lin, M., Ji, R., Xu, Z., Zhang, B., Wang, Y., Wu, Y., Huang, F., Lin, C.W.:
  Rotated binary neural network. Adv. Neural Inform. Process. Syst.
  \textbf{33},  7474--7485 (2020)

\bibitem{lin2017towards}
Lin, X., Zhao, C., Pan, W.: Towards accurate binary convolutional neural
  network. Adv. Neural Inform. Process. Syst.  \textbf{30} (2017)

\bibitem{liu2021adam}
Liu, Z., Shen, Z., Li, S., Helwegen, K., Huang, D., Cheng, K.T.: How do adam
  and training strategies help bnns optimization. In: Int. Conf. Mach. Learn.
  pp. 6936--6946. PMLR (2021)

\bibitem{liu2020reactnet}
Liu, Z., Shen, Z., Savvides, M., Cheng, K.T.: Reactnet: Towards precise binary
  neural network with generalized activation functions. In: Eur. Conf. Comput.
  Vis. pp. 143--159. Springer (2020)

\bibitem{liu2018bi}
Liu, Z., Wu, B., Luo, W., Yang, X., Liu, W., Cheng, K.T.: Bi-real net:
  Enhancing the performance of 1-bit cnns with improved representational
  capability and advanced training algorithm. In: Eur. Conf. Comput. Vis. pp.
  722--737 (2018)

\bibitem{long2015fully}
Long, J., Shelhamer, E., Darrell, T.: Fully convolutional networks for semantic
  segmentation. In: IEEE Conf. Comput. Vis. Pattern Recog. pp. 3431--3440
  (2015)

\bibitem{luo2020autopruner}
Luo, J.H., Wu, J.: Autopruner: An end-to-end trainable filter pruning method
  for efficient deep model inference. Pattern Recognition  \textbf{107},
  107461 (2020)

\bibitem{martinez2020training}
Martinez, B., Yang, J., Bulat, A., Tzimiropoulos, G.: Training binary neural
  networks with real-to-binary convolutions. arXiv preprint arXiv:2003.11535
  (2020)

\bibitem{nagel2022overcoming}
Nagel, M., Fournarakis, M., Bondarenko, Y., Blankevoort, T.: Overcoming
  oscillations in quantization-aware training. In: Int. Conf. Mach. Learn. pp.
  16318--16330. PMLR (2022)

\bibitem{Paszke_PyTorch_An_Imperative_2019}
Paszke, A., Gross, S., Massa, F., Lerer, A., Bradbury, J., Chanan, G., Killeen,
  T., Lin, Z., Gimelshein, N., Antiga, L., Desmaison, A., Kopf, A., Yang, E.,
  DeVito, Z., Raison, M., Tejani, A., Chilamkurthy, S., Steiner, B., Fang, L.,
  Bai, J., Chintala, S.: {PyTorch: An Imperative Style, High-Performance Deep
  Learning Library}. In: Wallach, H., Larochelle, H., Beygelzimer, A., d'Alché
  Buc, F., Fox, E., Garnett, R. (eds.) Adv. Neural Inform. Process. Syst. pp.
  8024--8035. Curran Associates, Inc. (2019),
  \url{http://papers.neurips.cc/paper/9015-pytorch-an-imperative-style-high-performance-deep-learning-library.pdf}

\bibitem{qin2020forward}
Qin, H., Gong, R., Liu, X., Shen, M., Wei, Z., Yu, F., Song, J.: Forward and
  backward information retention for accurate binary neural networks. In: IEEE
  Conf. Comput. Vis. Pattern Recog. pp. 2250--2259 (2020)

\bibitem{rajbhandari2020zero}
Rajbhandari, S., Rasley, J., Ruwase, O., He, Y.: Zero: Memory optimizations
  toward training trillion parameter models. In: SC20: International Conference
  for High Performance Computing, Networking, Storage and Analysis. pp. 1--16.
  IEEE (2020)

\bibitem{rastegari2016xnor}
Rastegari, M., Ordonez, V., Redmon, J., Farhadi, A.: Xnor-net: Imagenet
  classification using binary convolutional neural networks. In: Eur. Conf.
  Comput. Vis. pp. 525--542. Springer (2016)

\bibitem{ren2016faster}
Ren, S., He, K., Girshick, R., Sun, J.: Faster r-cnn: Towards real-time object
  detection with region proposal networks. IEEE Trans. Pattern Anal. Mach.
  Intell.  \textbf{39}(6),  1137--1149 (2016)

\bibitem{russakovsky2015imagenet}
Russakovsky, O., Deng, J., Su, H., Krause, J., Satheesh, S., Ma, S., Huang, Z.,
  Karpathy, A., Khosla, A., Bernstein, M., et~al.: Imagenet large scale visual
  recognition challenge. Int. J. Comput. Vis.  \textbf{115}(3),  211--252
  (2015)

\bibitem{shang2022network}
Shang, Y., Xu, D., Zong, Z., Yan, Y.: Network binarization via contrastive
  learning. arXiv preprint arXiv:2207.02970  (2022)

\bibitem{simonyan2014very}
Simonyan, K., Zisserman, A.: Very deep convolutional networks for large-scale
  image recognition. arXiv preprint arXiv:1409.1556  (2014)

\bibitem{su2021prioritized}
Su, X., Huang, T., Li, Y., You, S., Wang, F., Qian, C., Zhang, C., Xu, C.:
  Prioritized architecture sampling with monto-carlo tree search. In: IEEE
  Conf. Comput. Vis. Pattern Recog. pp. 10968--10977 (2021)

\bibitem{tu2022adabin}
Tu, Z., Chen, X., Ren, P., Wang, Y.: Adabin: Improving binary neural networks
  with adaptive binary sets. In: Eur. Conf. Comput. Vis. pp. 379--395. Springer
  (2022)

\bibitem{wu2023estimator}
Wu, X.M., Zheng, D., Liu, Z., Zheng, W.S.: Estimator meets equilibrium
  perspective: A rectified straight through estimator for binary neural
  networks training. In: Proceedings of the IEEE/CVF International Conference
  on Computer Vision. pp. 17055--17064 (2023)

\bibitem{xu2023resilient}
Xu, S., Li, Y., Ma, T., Lin, M., Dong, H., Zhang, B., Gao, P., Lu, J.:
  Resilient binary neural network. In: AAAI. vol.~37, pp. 10620--10628 (2023)

\bibitem{xu2022recurrent}
Xu, S., Li, Y., Wang, T., Ma, T., Zhang, B., Gao, P., Qiao, Y., L{\"u}, J.,
  Guo, G.: Recurrent bilinear optimization for binary neural networks. In: Eur.
  Conf. Comput. Vis. pp. 19--35. Springer (2022)

\bibitem{xu2021learning}
Xu, Y., Han, K., Xu, C., Tang, Y., Xu, C., Wang, Y.: Learning frequency domain
  approximation for binary neural networks. Adv. Neural Inform. Process. Syst.
  \textbf{34},  25553--25565 (2021)

\bibitem{xu2021recu}
Xu, Z., Lin, M., Liu, J., Chen, J., Shao, L., Gao, Y., Tian, Y., Ji, R.: Recu:
  Reviving the dead weights in binary neural networks. In: Int. Conf. Comput.
  Vis. pp. 5198--5208 (2021)

\bibitem{yang2019quantization}
Yang, J., Shen, X., Xing, J., Tian, X., Li, H., Deng, B., Huang, J., Hua, X.s.:
  Quantization networks. In: IEEE Conf. Comput. Vis. Pattern Recog. pp.
  7308--7316 (2019)

\bibitem{yang2020searching}
Yang, Z., Wang, Y., Han, K., Xu, C., Xu, C., Tao, D., Xu, C.: Searching for
  low-bit weights in quantized neural networks. Adv. Neural Inform. Process.
  Syst.  \textbf{33},  4091--4102 (2020)

\bibitem{you2017large}
You, Y., Gitman, I., Ginsburg, B.: Large batch training of convolutional
  networks. arXiv preprint arXiv:1708.03888  (2017)

\bibitem{zhang2018lq}
Zhang, D., Yang, J., Ye, D., Hua, G.: Lq-nets: Learned quantization for highly
  accurate and compact deep neural networks. In: Eur. Conf. Comput. Vis. pp.
  365--382 (2018)

\bibitem{zhang2018shufflenet}
Zhang, X., Zhou, X., Lin, M., Sun, J.: Shufflenet: An extremely efficient
  convolutional neural network for mobile devices. In: IEEE Conf. Comput. Vis.
  Pattern Recog. pp. 6848--6856 (2018)

\bibitem{zhao2021uncertainty}
Zhao, J., Yang, L., Zhang, B., Guo, G., Doermann, D.S.: Uncertainty-aware
  binary neural networks. In: IJCAI. pp. 3441--3447 (2021)

\bibitem{zhou2016dorefa}
Zhou, S., Wu, Y., Ni, Z., Zhou, X., Wen, H., Zou, Y.: Dorefa-net: Training low
  bitwidth convolutional neural networks with low bitwidth gradients. arXiv
  preprint arXiv:1606.06160  (2016)

\bibitem{zhu2020towards}
Zhu, F., Gong, R., Yu, F., Liu, X., Wang, Y., Li, Z., Yang, X., Yan, J.:
  Towards unified int8 training for convolutional neural network. In: IEEE
  Conf. Comput. Vis. Pattern Recog. pp. 1969--1979 (2020)

\end{thebibliography}

\newpage
\appendix
\section*{Appendix}

In this Appendix we provide the following material:

\begin{itemize}
\item \cref{appendix:sec:more_results} demonstrates more weight flipping information for vanilla BNNs and OvSW in additional to \cref{fig:main_1}.
\item \cref{appendix:sec:generalize} extrapolates the conclusions in \cref{sec:gradient_and_weight} to a more generalized scenario.
\item \cref{appdendix:sec:ablation_analysis_weight_init} conducts ablation analysis for different weight initialization methods,
including kaiming normal~\cite{he2015delving} and kaiming uniform~\cite{he2015delving} and the same weight initialization methods with different std or range.
\item \cref{appendix:sec:ags_vs_lars} compares AGS with LARS.
\item \cref{appendix:sec:latent_adam} disscusses the difference of OvSW with latent weights~(\cref{appendix:ssec:latent}) and adam optimizer~(\cref{appendix:ssec:optimizer}) for BNNs.
\item \cref{appendix:sec:impact} describes societal impact of OvSW.
\end{itemize}

\newpage
\section{More Experimental Results}
\label{appendix:sec:more_results}

\begin{figure*}[h]
{
    \centering
    \begin{subtable}[h]{0.32\textwidth}
        \includegraphics[width=1.\linewidth]{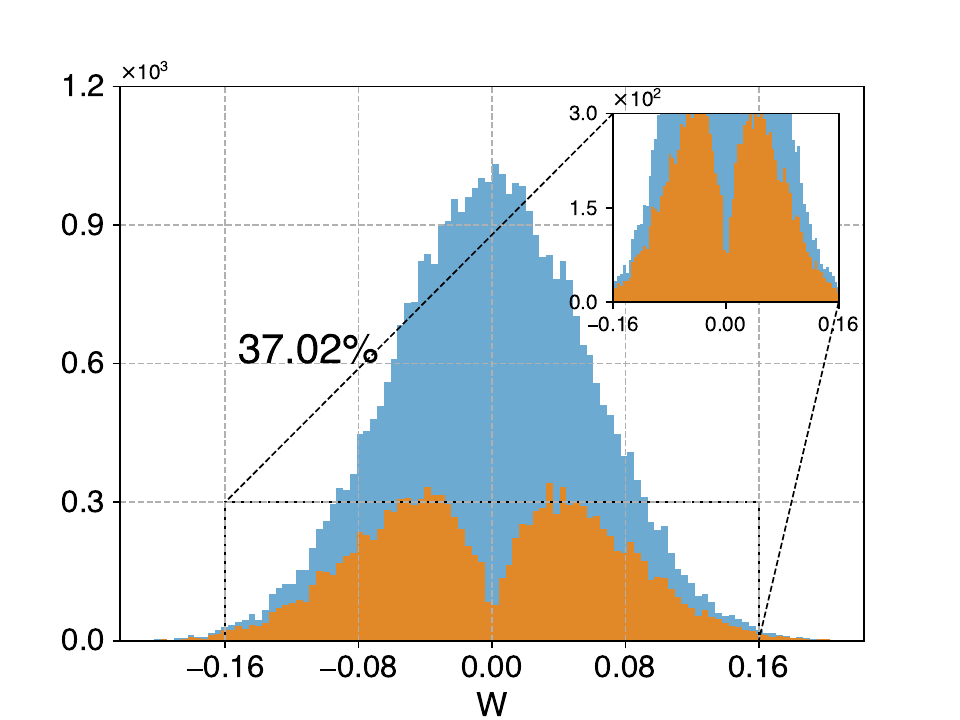}
        \caption{layer1.1.conv2.weight}
    \end{subtable}
    \begin{subtable}[h]{0.32\textwidth}
        \includegraphics[width=1\linewidth]{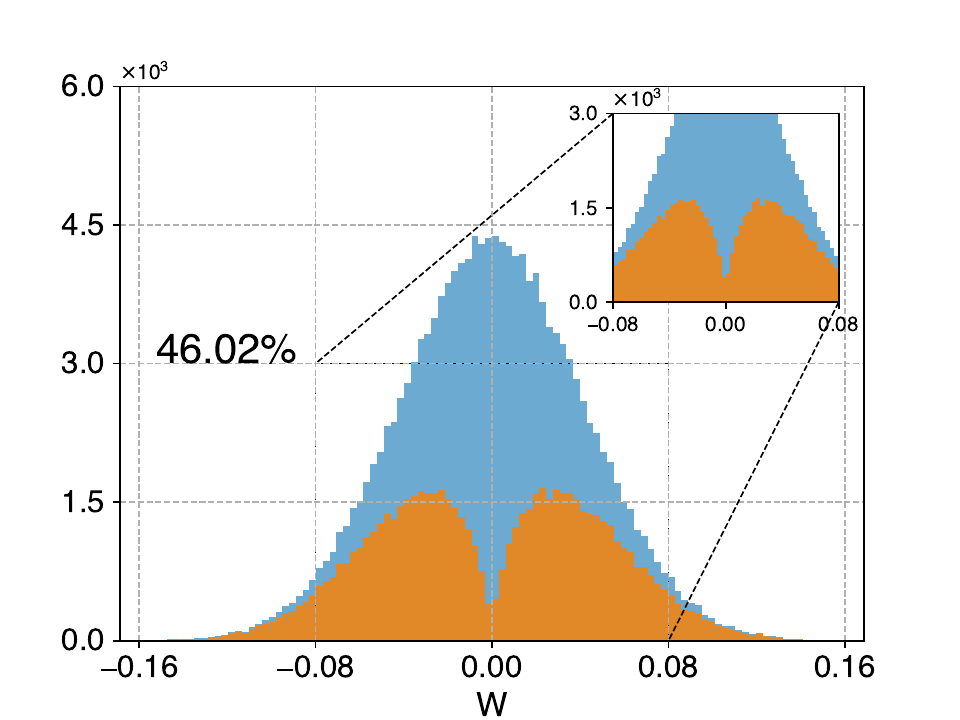}
        \caption{layer2.1.conv2.weight}
    \end{subtable}
    \begin{subtable}[h]{0.32\textwidth}
        \includegraphics[width=1\linewidth]{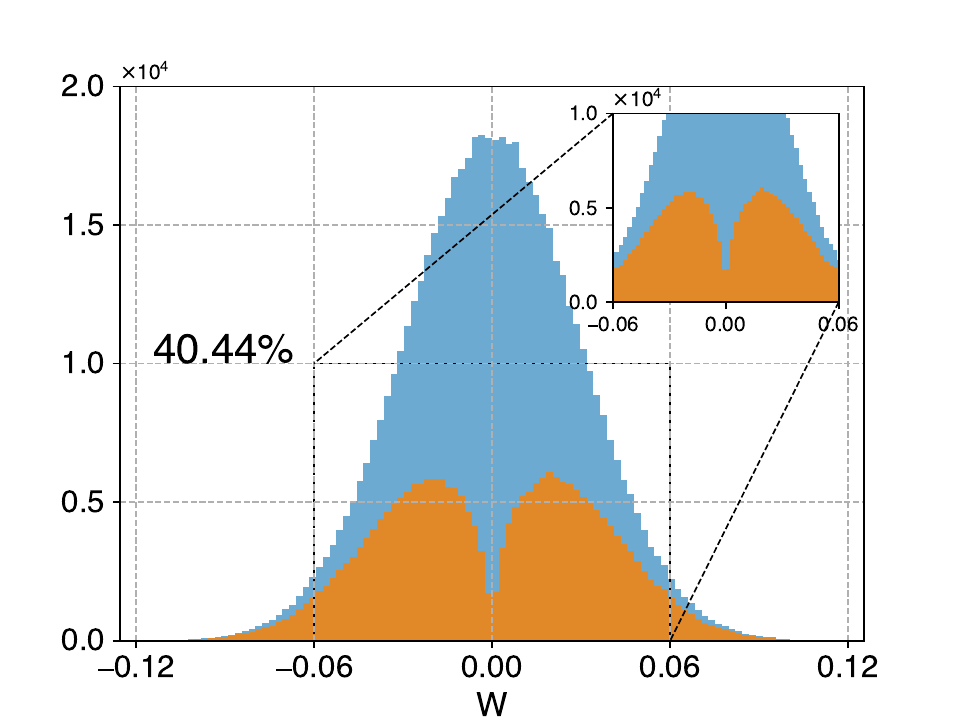}
        \caption{layer3.1.conv2.weight}
    \end{subtable}
    \caption{
    Histogram of the initialized weight distribution~(blue) and the weights
    that never update signs throughout training~(orange) for Vanilla BNNs.
    37.02\%, 46.02\% and 40.44\% represent the ratio of the corresponding orange area to the blue.
    }
    \label{fig:add_vanilla}
}

{
    \centering
    \begin{subtable}[h]{0.32\textwidth}
        \includegraphics[width=1.\linewidth]{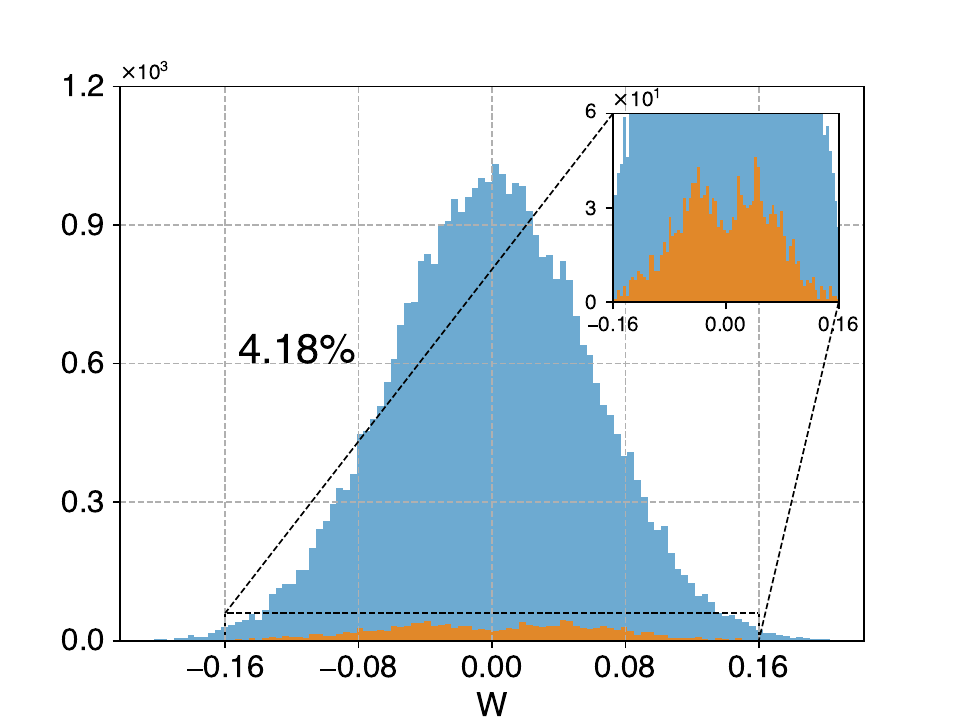}
        \caption{layer1.1.conv2.weight}
    \end{subtable}
    \begin{subtable}[h]{0.32\textwidth}
        \includegraphics[width=1\linewidth]{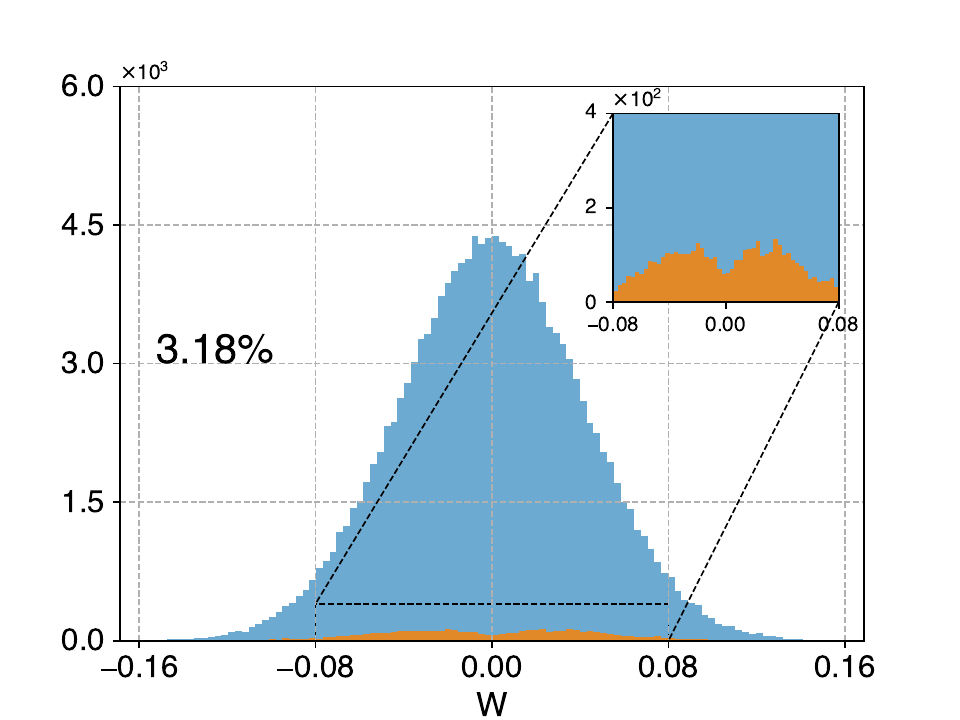}
        \caption{layer2.1.conv2.weight}
    \end{subtable}
    \begin{subtable}[h]{0.32\textwidth}
        \includegraphics[width=1\linewidth]{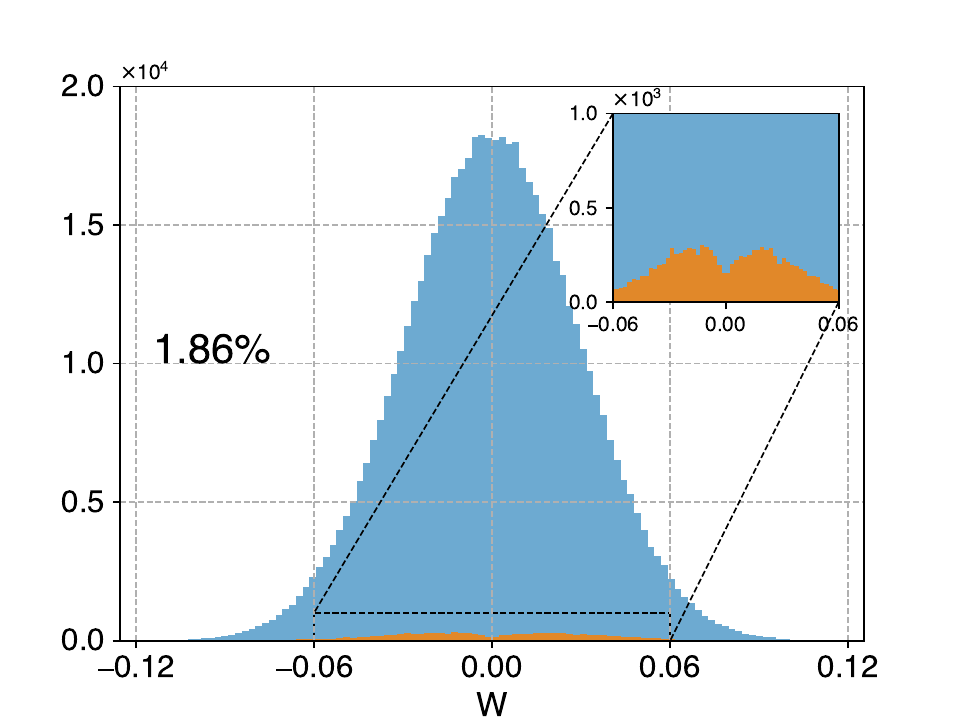}
        \caption{layer3.1.conv2.weight}
    \end{subtable}
    \caption{
    Histogram of the initialized weight distribution~(blue) and the weights
    that never update signs throughout training~(orange) for OvSW.
    4.18\%, 3.18\% and 1.86\% represent the ratio of the corresponding orange area to the blue.
    }
    \label{fig:add_ovsw}
}
\vspace{-0.2in}
\end{figure*}

We further demonstrate the weight flip information of the relevant layers based on \cref{fig:main_1}.
As seen in \cref{fig:add_vanilla} and \cref{fig:add_ovsw}, in addition to the weight flipping of layer4.conv2.weight, OvSW also promotes weight flip efficiency for other layers.

\newpage
\section{A More Generalized Scenario}
\label{appendix:sec:generalize}

In this section, we further generalize the conclusion in \cref{sec:gradient_and_weight} to the more general scenario,
where $\mathcal{N}$ and $\mathcal{N}^{'}$  have different $\alpha_j$ and $\alpha_j^{'}$.

\begin{figure}[t]
    \centering
    \includegraphics[width=\linewidth]{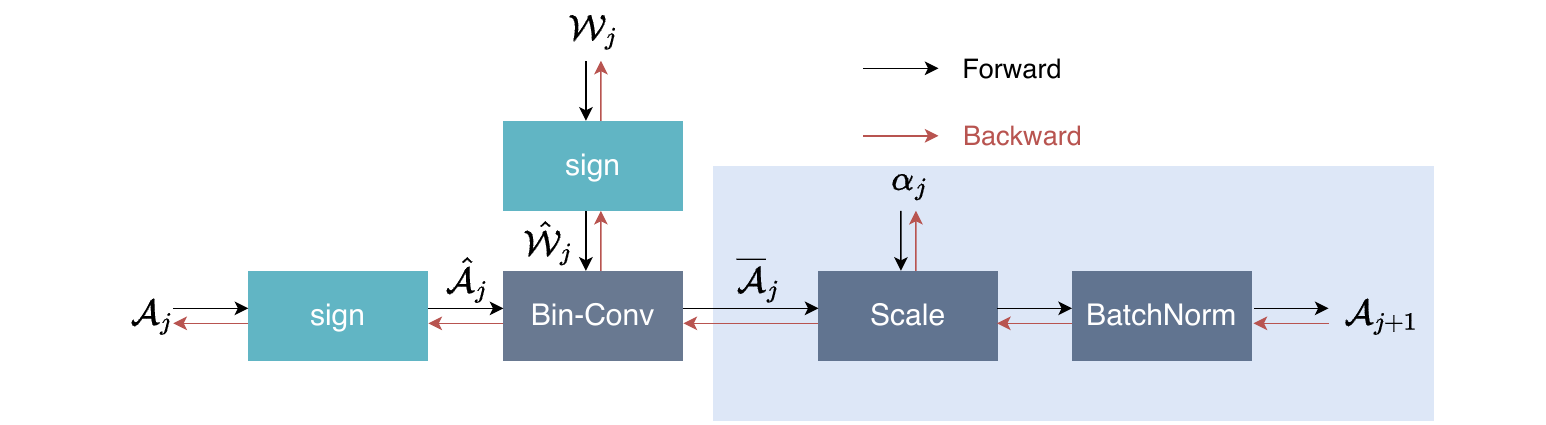}
    \caption{Forward and backward computation graph for binarized convolutional operation with quantization aware training.}
    \label{fig:illustration_appendix}
    \vspace{-0.2in}
\end{figure}

Firstly, we rewritten \cref{eq:binary_conv} as:
\begin{equation}
\begin{aligned}
\mathcal{A}_{j+1} & =\mathrm{BN}\left( \left(\hat{\mathcal{A}}_{j}\circledast \hat{\mathcal{W}}_{j} \right) \odot \alpha_j \right )
=\mathrm{BN}\left(\overline{\mathcal{A}}_j\odot \alpha_j \right) \\
& =\mathrm{BN}\left(\mathrm{diag}(\alpha_j)\overline{\mathcal{A}}_j\right)
=\mathrm{BN}\left(\Lambda_j\overline{\mathcal{A}}_j\right).
\end{aligned}
\label{eq:binary_conv_written}
\end{equation}

%
Assuming that for two networks $\mathcal{N}$ and $\mathcal{N}^{'}$, where $\mathcal{A}_{j}=\mathcal{A}_{j}^{'}$ and $\mathrm{sign}(\mathcal{W}_{j})=\mathrm{sign}(\mathcal{W}_{j}^{'})$,
it is easy to obtain that $\overline{\mathcal{A}}_j$ is equivalent to $\overline{\mathcal{A}}_{j}^{'}$.
Therefore, we pay our attention to $\frac{\partial \mathcal{L}}{\partial \overline{\mathcal{A}}_{j}}$ and $\frac{\partial \mathcal{L}}{\partial \overline{\mathcal{A}}_{j}^{'}}$.
Assuming that BN in both $\mathcal{N}$ and $\mathcal{N}^{'}$ can estimate the mean and variance of $\Lambda_j\overline{\mathcal{A}}_j$ and $\Lambda_j^{'}\overline{\mathcal{A}}_j^{'}$ with relative precision,
and learnable affine parameters $\Gamma_j=\Gamma_j^{'}, \beta_j=\beta_j^{'}$
we can know for the forward propagation: 
\begin{equation}
\begin{aligned}
\mathcal{A}_{j+1} & = \mathrm{BN}\left(\Lambda_j\overline{\mathcal{A}}_j\right) \\
& \approx \Gamma_j \odot \frac{\Lambda_j\overline{\mathcal{A}}_j-\mu \left(\Lambda_j\overline{\mathcal{A}}_j\right)}{\sigma\left(\Lambda_j\overline{\mathcal{A}}_j\right)}+\beta_j \\
& \approx \Gamma_j^{'} \odot \frac{\Lambda_j^{'}\overline{\mathcal{A}}_j^{'}-\mu \left(\Lambda_j^{'}\overline{\mathcal{A}}_j^{'}\right)}{\sigma\left(\Lambda_j^{'}\overline{\mathcal{A}}_j^{'}\right)}+\beta_j^{'} \\
& = \mathrm{BN}^{'}\left(\Lambda_j^{'}\overline{\mathcal{A}}_j^{'}\right) = \mathcal{A}^{'}_{j+1},
\end{aligned}  
\label{eq:general_forward}
\end{equation}
Thus, $\mathcal{N}$ and $\mathcal{N}^{'}$ will have the same output and loss, \ie $\mathcal{L}=\mathcal{L}^{'}$. For the back propagation:
{\begin{equation}
    \begin{aligned}
        \frac{\partial  \mathcal{L}}{\partial \overline{\mathcal{A}}_{j}} 
        & = \frac{\partial  \mathcal{L}}{\partial \mathcal{A}_{j+1}}
        \frac{\partial \mathcal{A}_{j+1}}{\partial \left(\Lambda_j\overline{\mathcal{A}}_j\right)} 
        \frac{\partial \left(\Lambda_j\overline{\mathcal{A}}_j\right)}{\partial \overline{\mathcal{A}}_{j}}
         \approx \frac{\partial  \mathcal{L}}{\partial \mathcal{A}_{j+1}}
        \frac{\partial \mathcal{A}_{j+1}}{\partial \overline{\mathcal{A}}_{j}} \Lambda_{j}^{-1}\Lambda_j
        \\ & = \frac{\partial  \mathcal{L}}{\partial \mathcal{A}_{j+1}^{'}}
        \frac{\partial \mathcal{A}_{j+1}^{'}}{\partial \overline{\mathcal{A}}_{j}^{'}}  \Lambda_{j}^{{'}^{-1}}\Lambda_j^{'}
        \approx \frac{\partial  \mathcal{L}^{'}}{\partial \mathcal{A}_{j+1}^{'}}
        \frac{\partial \mathcal{A}_{j+1}^{'}}{\partial \left(\Lambda_j^{'}\overline{\mathcal{A}}_j^{'}\right)} 
        \frac{\partial \left(\Lambda_j^{'}\overline{\mathcal{A}}_j^{'}\right)}{\partial \overline{\mathcal{A}}_{j}^{'}}
        =\frac{\partial  \mathcal{L}^{'}}{\partial \overline{\mathcal{A}}_{j}^{'}}.
    \end{aligned}
\label{eq:general_backward}
\end{equation}} 
Then we can obtain:
\begin{equation}
\begin{aligned}
\frac{\partial  \mathcal{L}}{\partial{\mathcal{A}}_{j}} = \frac{\partial  \mathcal{L}}{\partial \overline{\mathcal{A}}_{j}} \frac{\partial  \overline{\mathcal{A}}_{j}}{\partial {\mathcal{A}}_{j}} 
=\frac{\partial  \mathcal{L}^{'}}{\partial \overline{\mathcal{A}}_{j}^{'}} \frac{\partial  \overline{\mathcal{A}}_{j}^{'}}{\partial {\mathcal{A}}_{j}^{'}}=\frac{\partial  \mathcal{L}^{'}}{\partial{\mathcal{A}}_{j}^{'}}.
\end{aligned}
\label{eq:generation_final}
\end{equation}

From \cref{eq:general_forward}, \cref{eq:general_backward} and \cref{eq:generation_final},
we observe that due to the existence of BN, the value of $\alpha_j$ and $\alpha_j^{'}$ hardly affects the results of both forward and backward propagation.
Thus the flipping efficiency of the weight signs is also independent of the value of $\alpha$, 
and it is the relationship between the gradient and the weight distribution that really makes a contribution.

\newpage

\begin{figure*}[p]
    \centering
    \begin{subtable}[h]{0.48\textwidth}
        \includegraphics[width=1\linewidth]{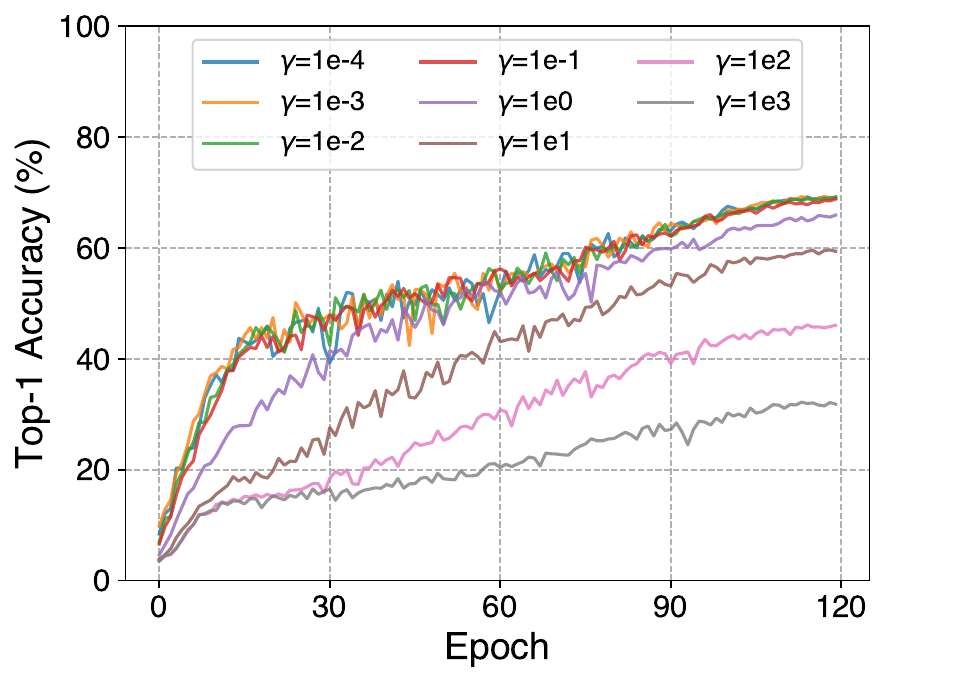}
        \caption{Kaiming Normal}
    \end{subtable}
    \begin{subtable}[h]{0.48\textwidth}
        \includegraphics[width=1\linewidth]{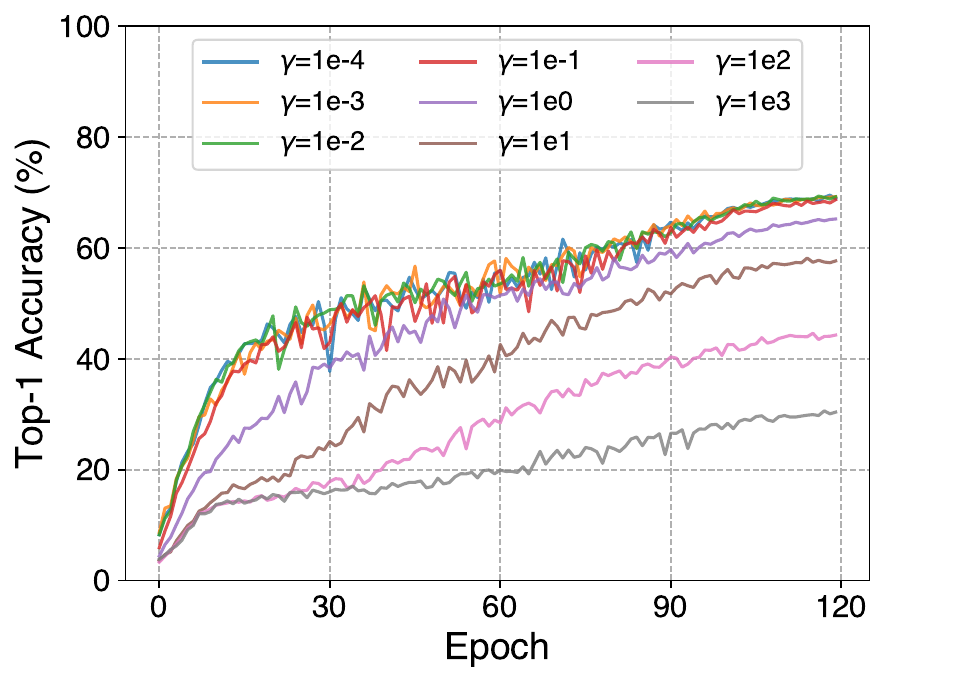}
        \caption{Kaiming Uniform}
    \end{subtable}
    \caption{Convergence for different weight initialization.    \label{fig:convergence_gamma}}
    \centering
    \includegraphics[width=0.5\linewidth]{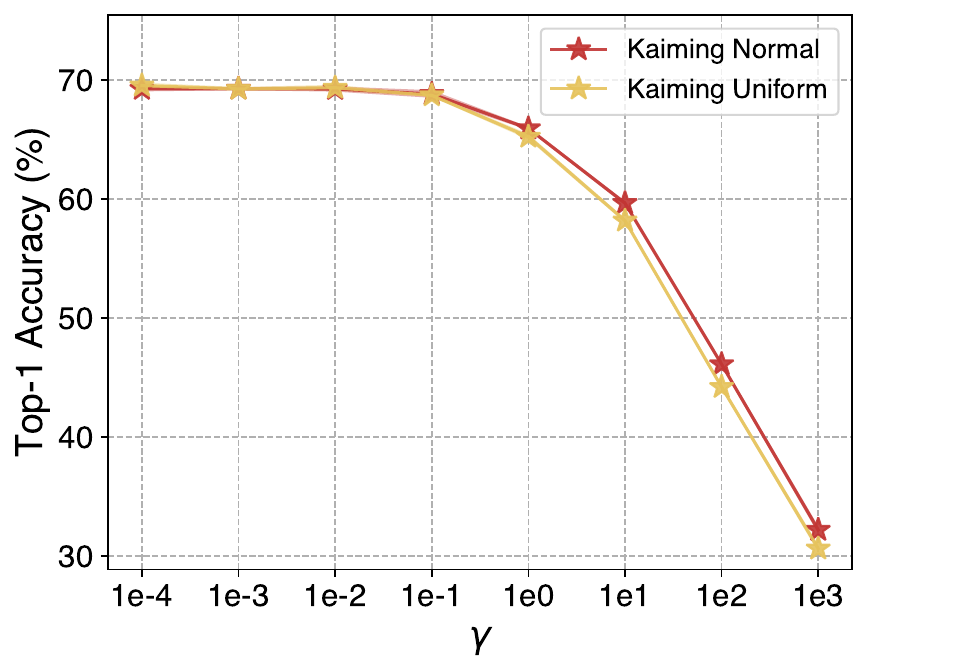}
    \vspace{-0.1in}
    \caption{Mean top-1 accuracy~(mean$\pm$std) of binarized ResNet18 \wrt different values for different $\gamma$ with different weight initialization methods on CIFAR100.    \label{fig:vary_gamma}}
    \vspace{-0.2in}
\end{figure*}

\begin{figure*}[p]
{
    \centering
    \begin{subtable}[h]{0.24\textwidth}
        \includegraphics[width=1\linewidth]{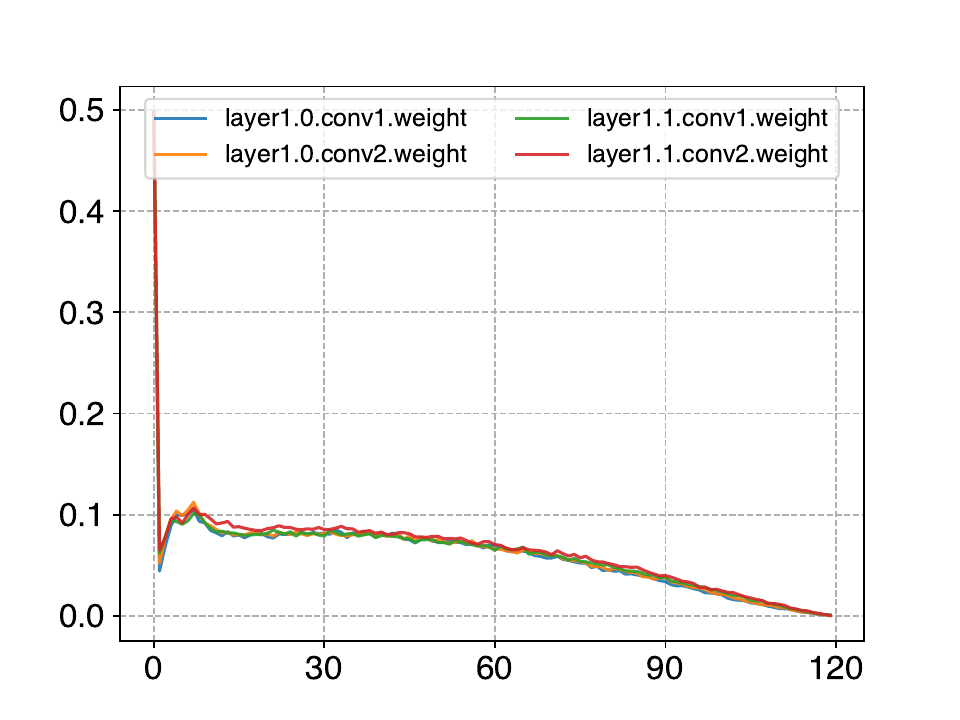}
        \caption{layer1}
    \end{subtable}
    \begin{subtable}[h]{0.24\textwidth}
        \includegraphics[width=1\linewidth]{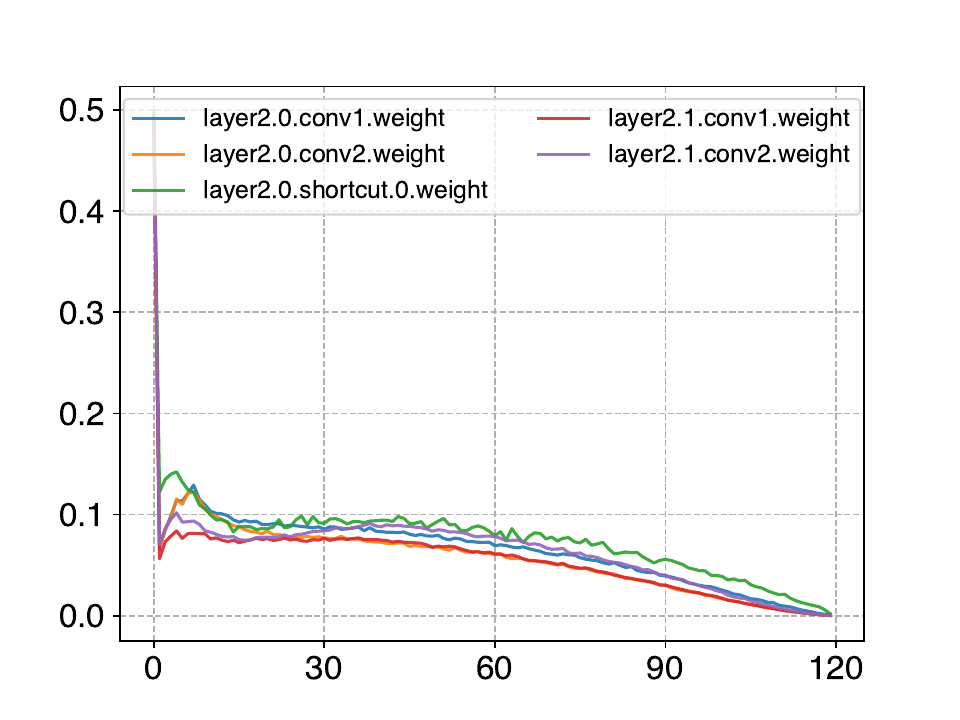}
        \caption{layer2}
    \end{subtable}
        \begin{subtable}[h]{0.24\textwidth}
        \includegraphics[width=1\linewidth]{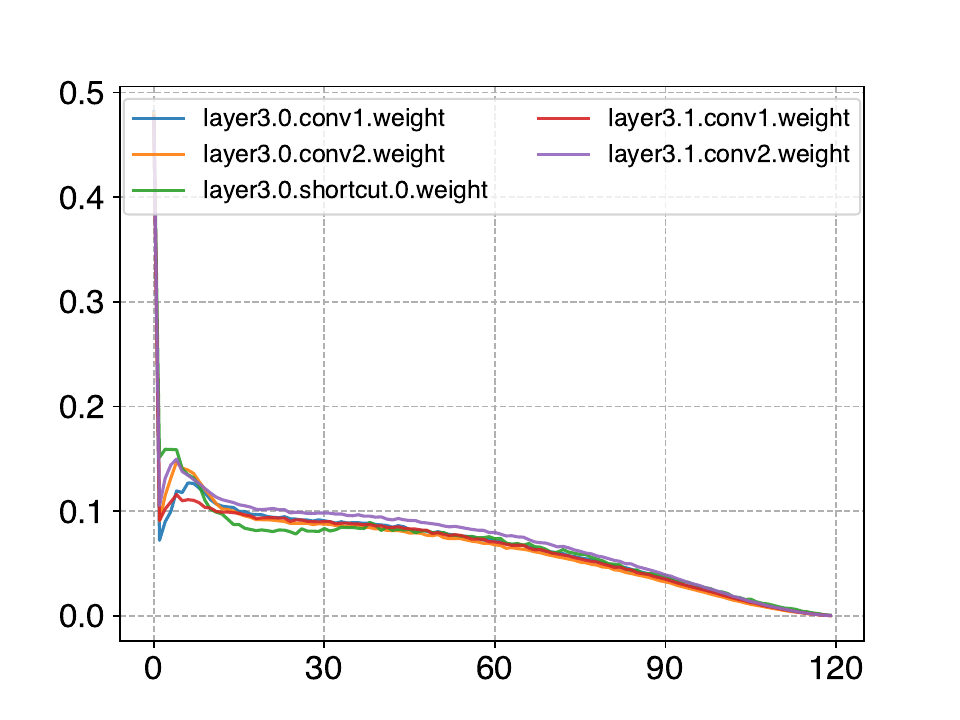}
        \caption{layer3}
    \end{subtable}
        \begin{subtable}[h]{0.24\textwidth}
        \includegraphics[width=1\linewidth]{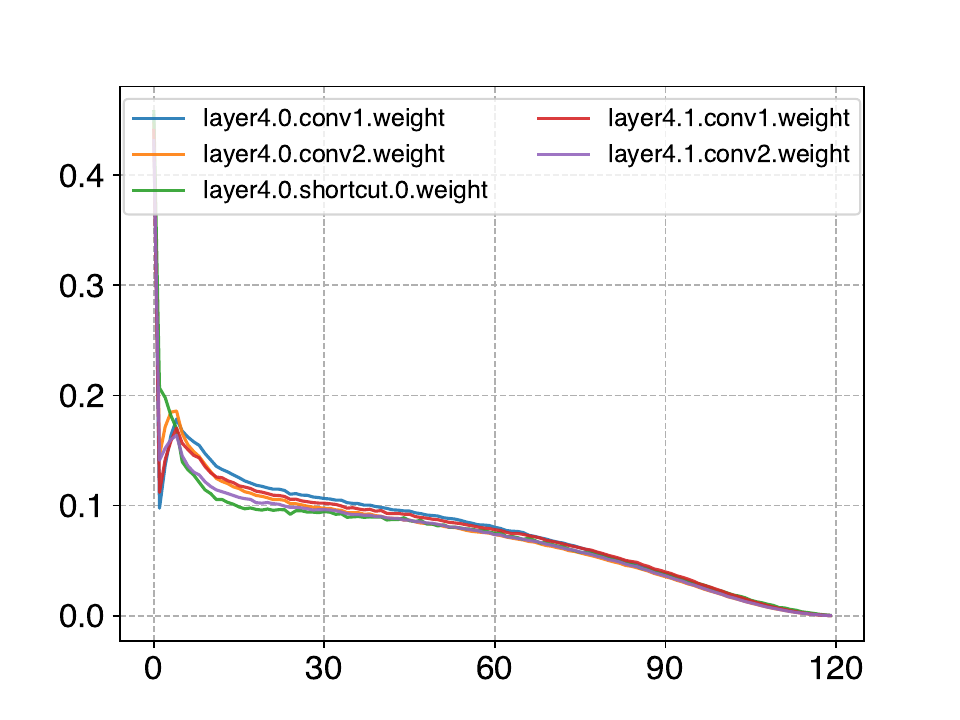}
        \caption{layer4}
    \end{subtable}
    \caption{Epoch-wise flip rate for $\gamma=0.0001$~(kaiming normal).}
    \label{fig:normal_0.0001}
}

    \centering
    \begin{subtable}[h]{0.24\textwidth}
        \includegraphics[width=1\linewidth]{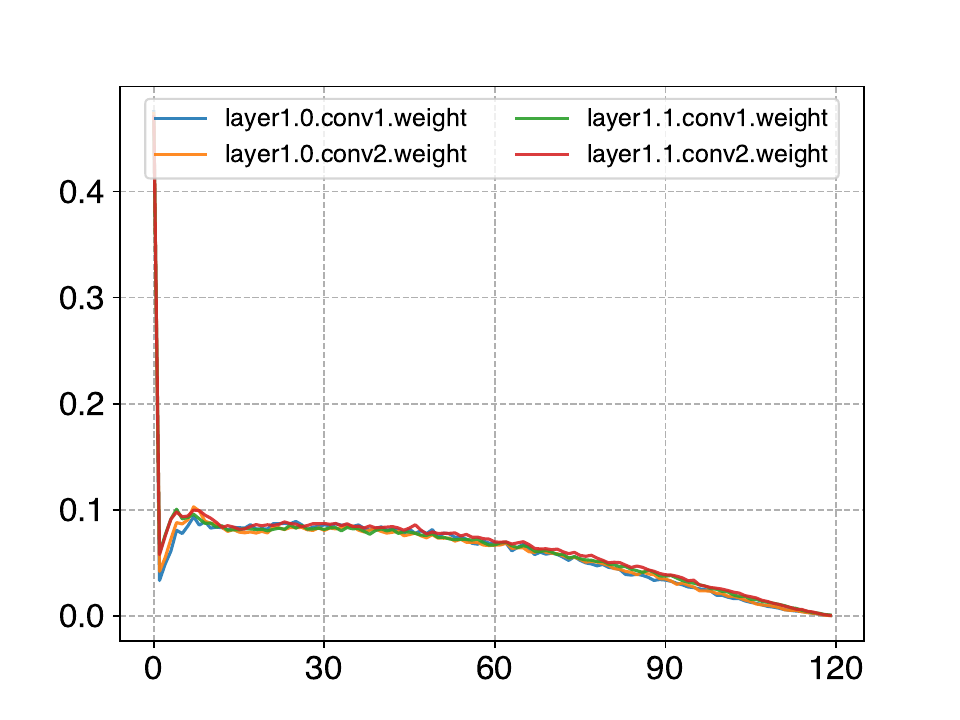}
        \caption{layer1}
    \end{subtable}
    \begin{subtable}[h]{0.24\textwidth}
        \includegraphics[width=1\linewidth]{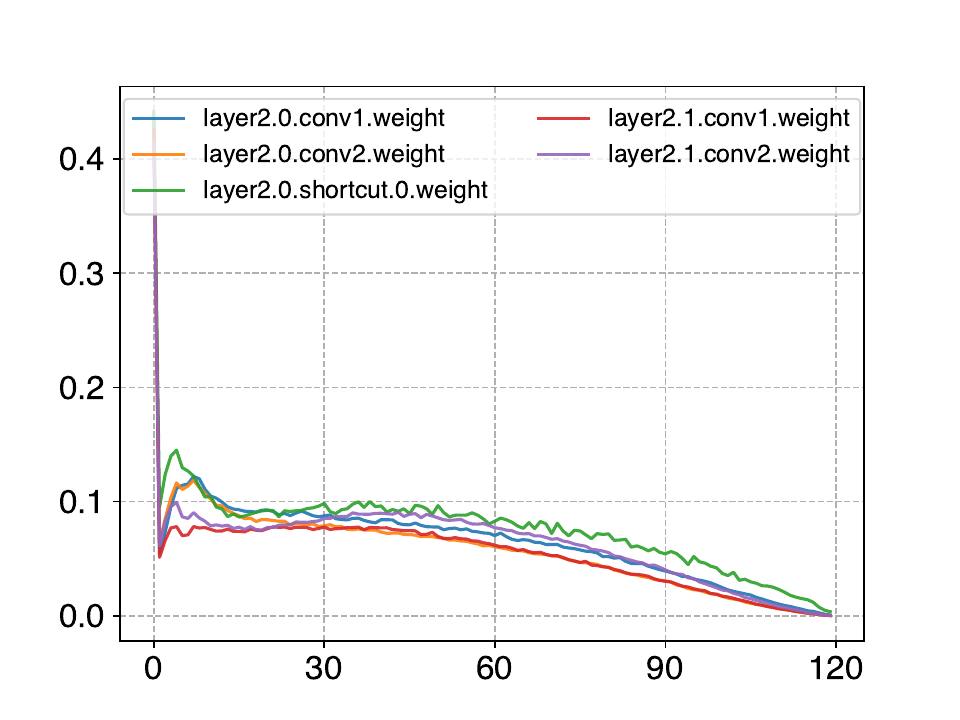}
        \caption{layer2}
    \end{subtable}
        \begin{subtable}[h]{0.24\textwidth}
        \includegraphics[width=1\linewidth]{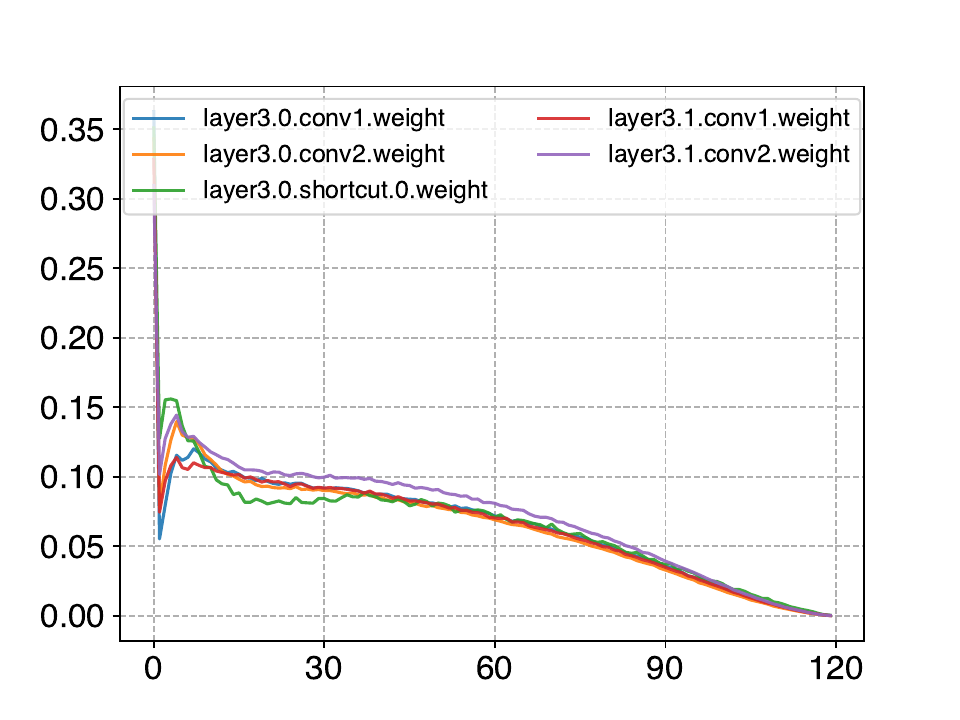}
        \caption{layer3}
    \end{subtable}
        \begin{subtable}[h]{0.24\textwidth}
        \includegraphics[width=1\linewidth]{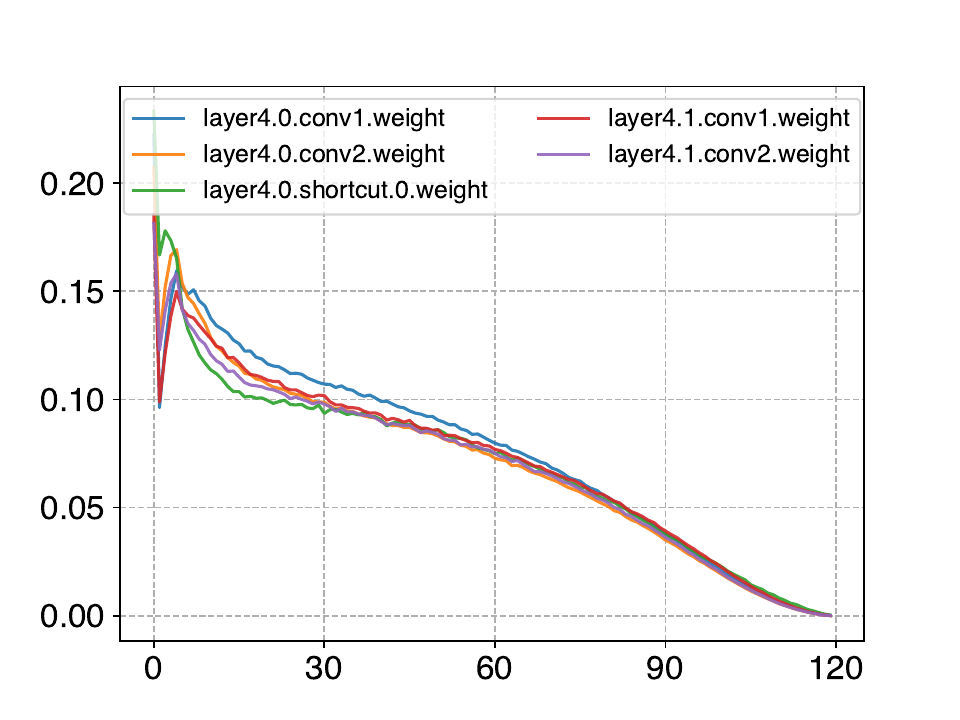}
        \caption{layer4}
    \end{subtable}
    \caption{Epoch-wise flip rate for $\gamma=0.001$~(kaiming normal).}

    \centering
    \begin{subtable}[h]{0.24\textwidth}
        \includegraphics[width=1\linewidth]{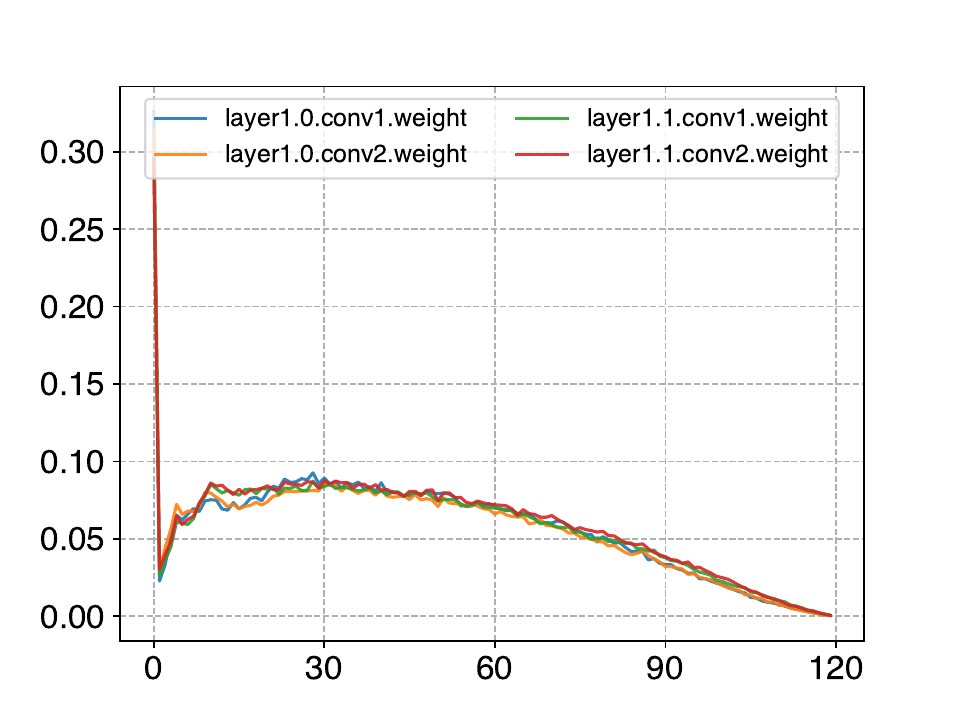}
        \caption{layer1}
    \end{subtable}
    \begin{subtable}[h]{0.24\textwidth}
        \includegraphics[width=1\linewidth]{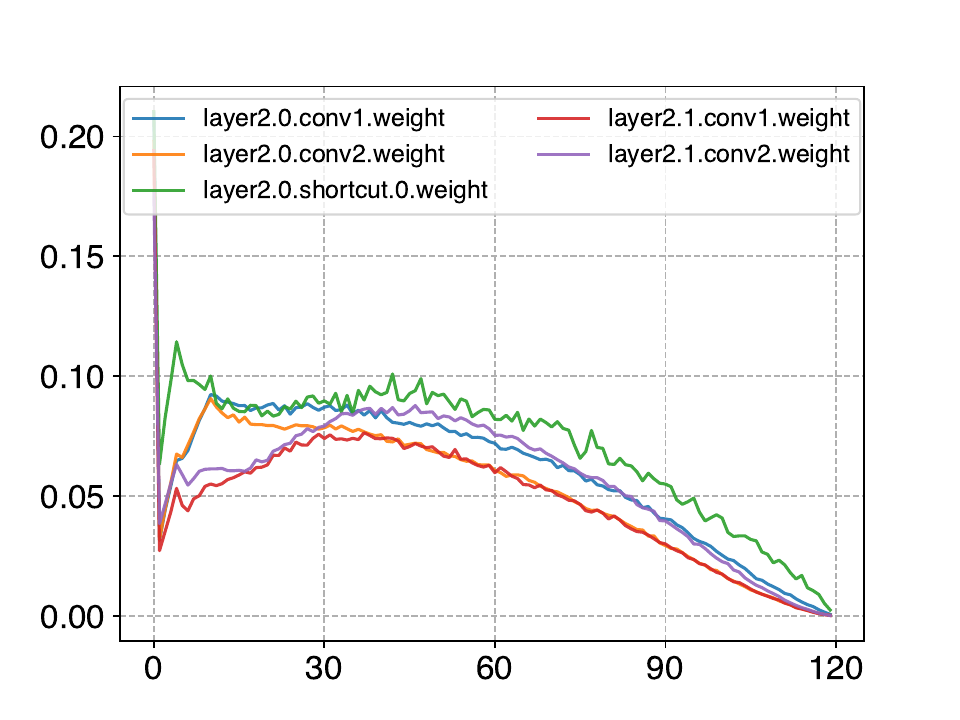}
        \caption{layer2}
    \end{subtable}
        \begin{subtable}[h]{0.24\textwidth}
        \includegraphics[width=1\linewidth]{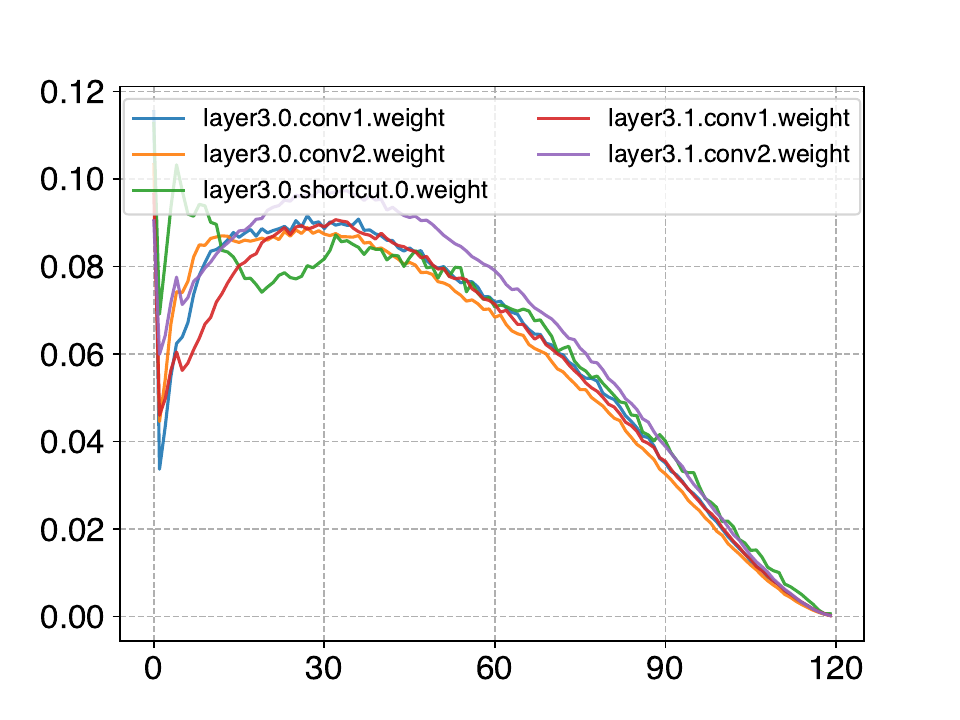}
        \caption{layer3}
    \end{subtable}
        \begin{subtable}[h]{0.24\textwidth}
        \includegraphics[width=1\linewidth]{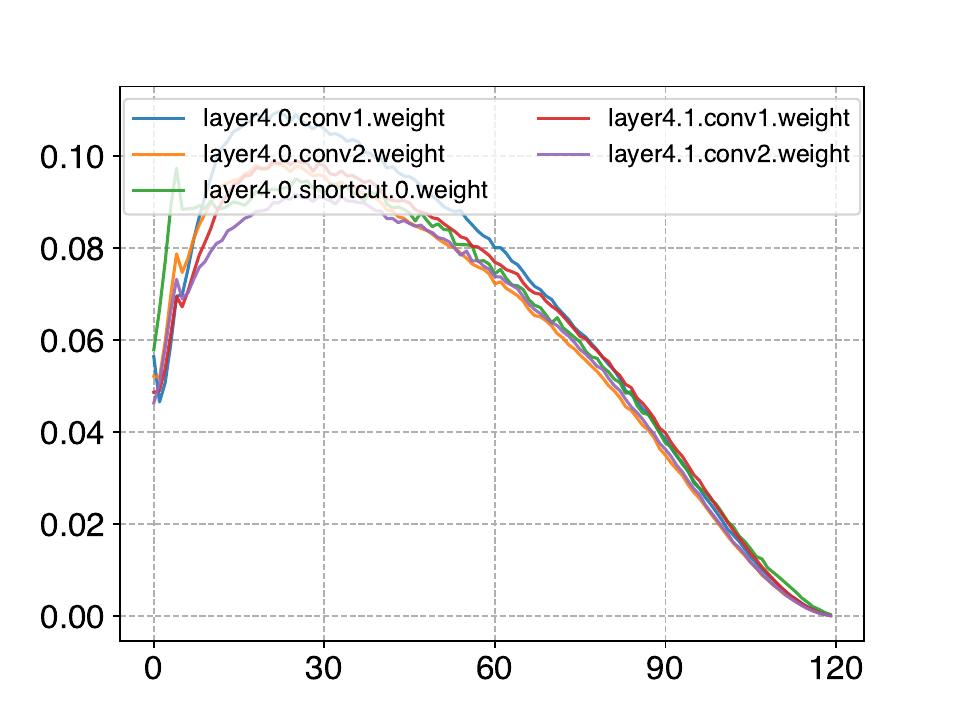}
        \caption{layer4}
    \end{subtable}
    \caption{Epoch-wise flip rate for $\gamma=0.01$~(kaiming normal).}

    \centering
    \begin{subtable}[h]{0.24\textwidth}
        \includegraphics[width=1\linewidth]{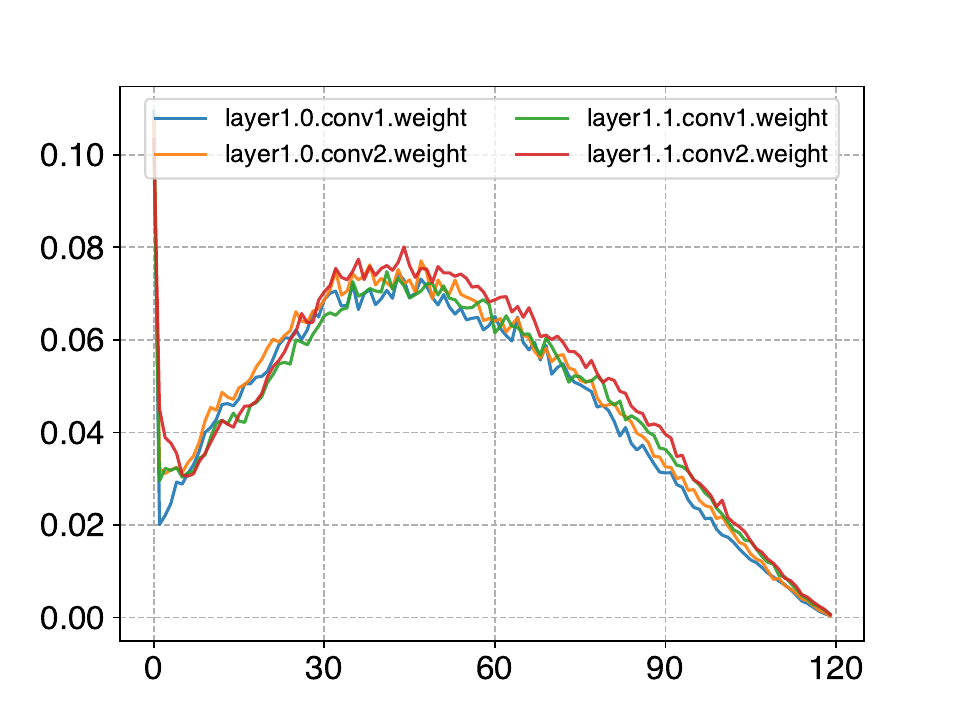}
        \caption{layer1}
    \end{subtable}
    \begin{subtable}[h]{0.24\textwidth}
        \includegraphics[width=1\linewidth]{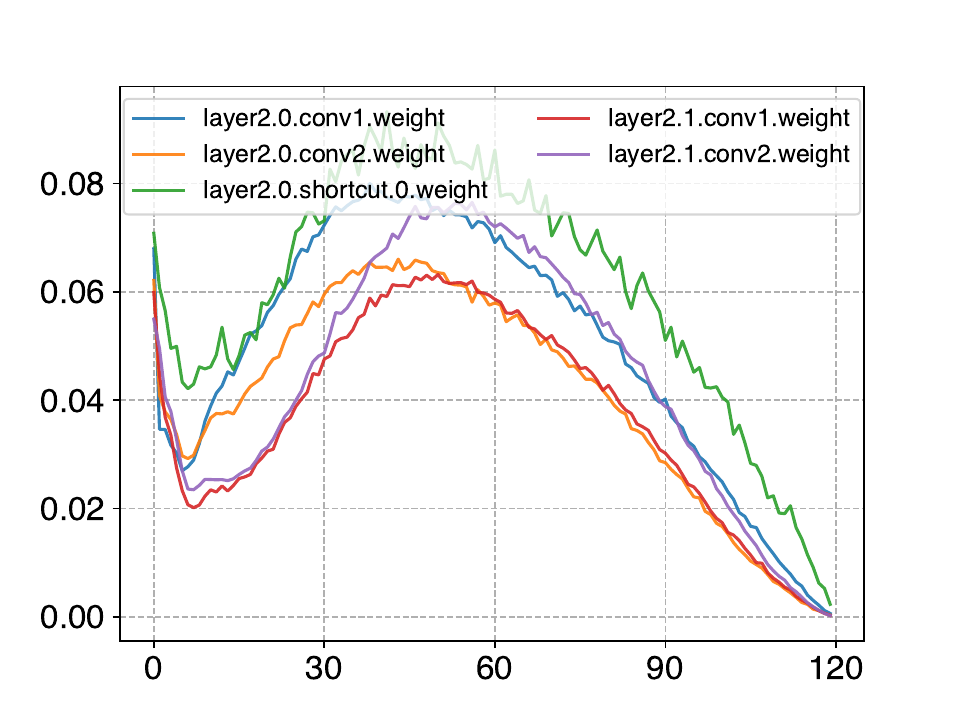}
        \caption{layer2}
    \end{subtable}
        \begin{subtable}[h]{0.24\textwidth}
        \includegraphics[width=1\linewidth]{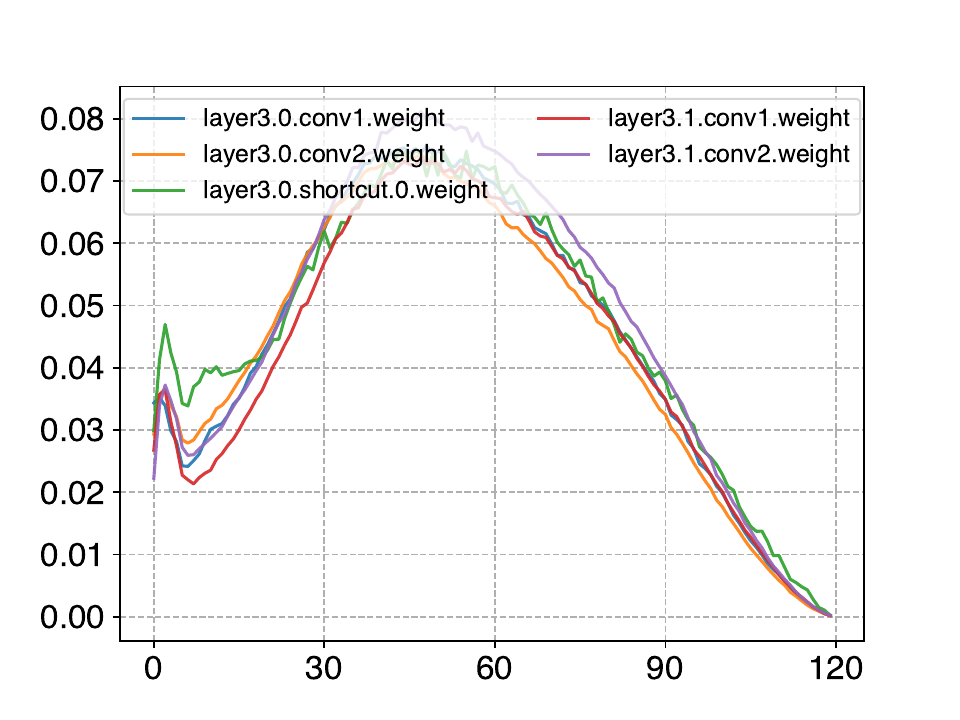}
        \caption{layer3}
    \end{subtable}
        \begin{subtable}[h]{0.24\textwidth}
        \includegraphics[width=1\linewidth]{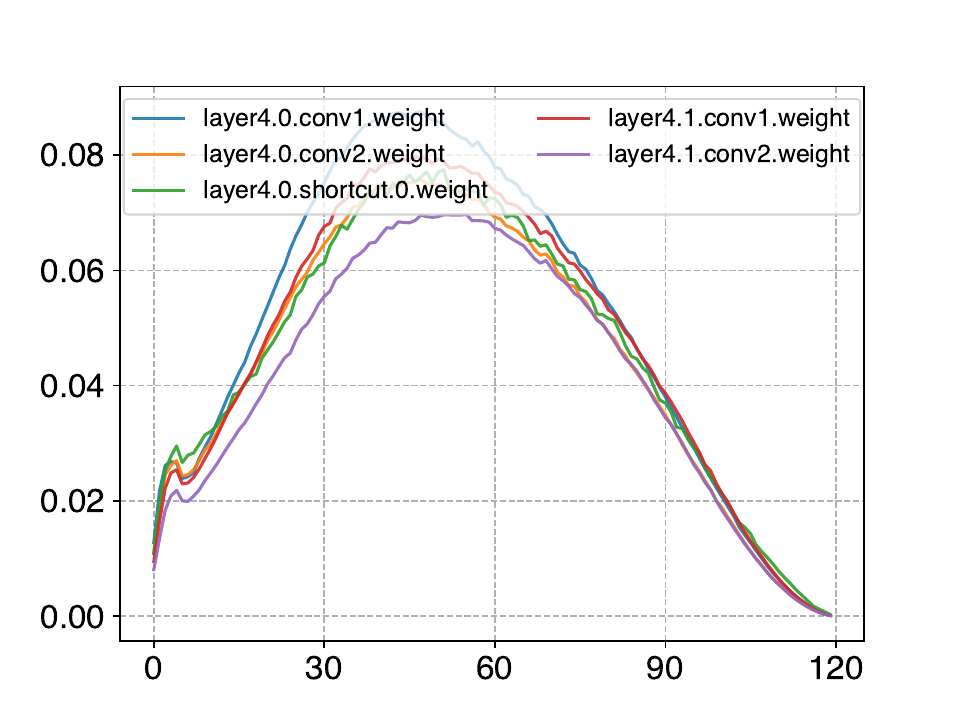}
        \caption{layer4}
    \end{subtable}
    \caption{Epoch-wise flip rate for $\gamma=0.1$~(kaiming normal).}
\end{figure*}

\begin{figure*}[p]
    \centering
    \begin{subtable}[h]{0.24\textwidth}
        \includegraphics[width=1\linewidth]{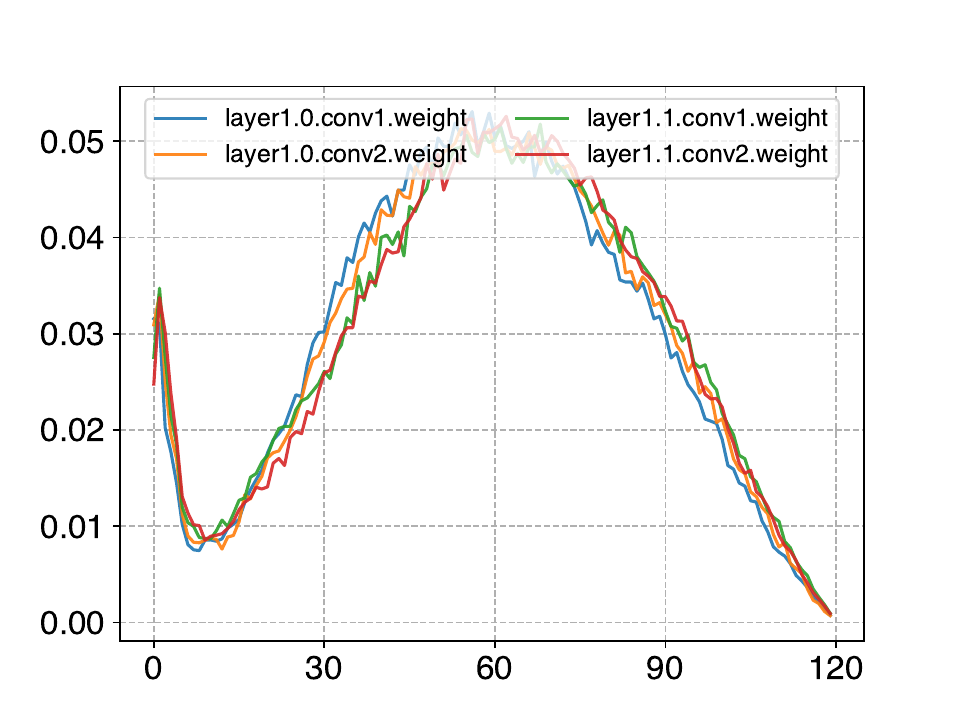}
        \caption{layer1}
    \end{subtable}
    \begin{subtable}[h]{0.24\textwidth}
        \includegraphics[width=1\linewidth]{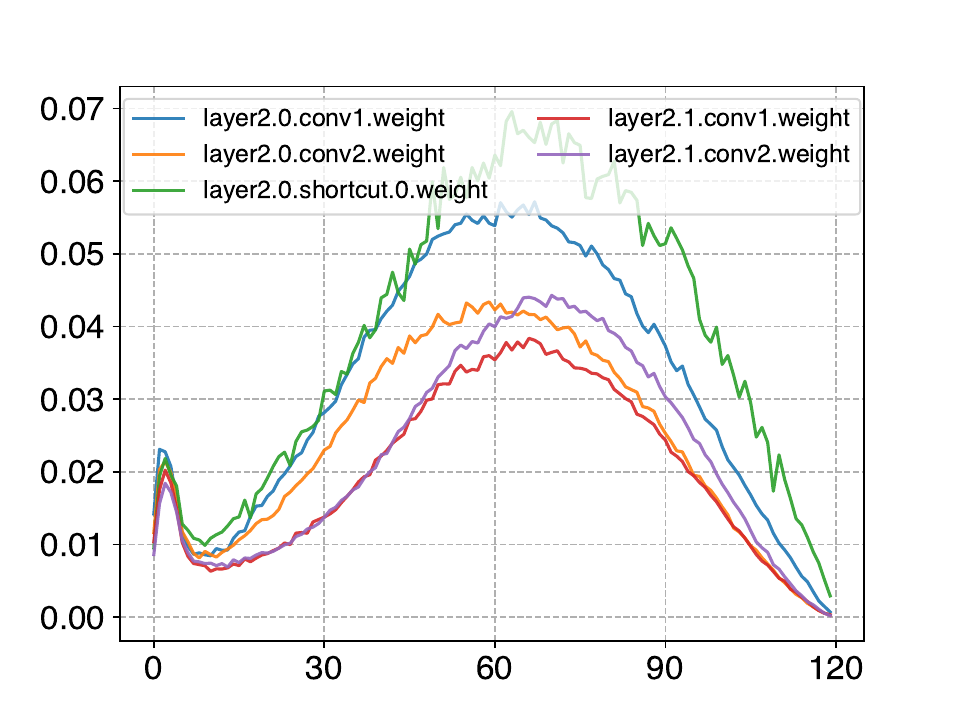}
        \caption{layer2}
    \end{subtable}
        \begin{subtable}[h]{0.24\textwidth}
        \includegraphics[width=1\linewidth]{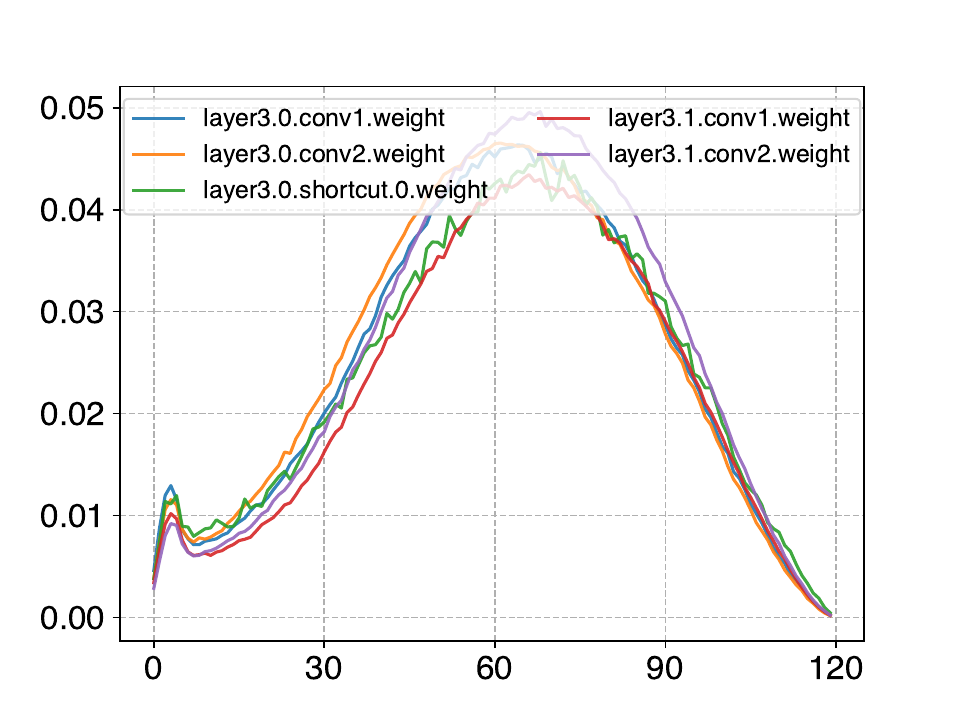}
        \caption{layer3}
    \end{subtable}
        \begin{subtable}[h]{0.24\textwidth}
        \includegraphics[width=1\linewidth]{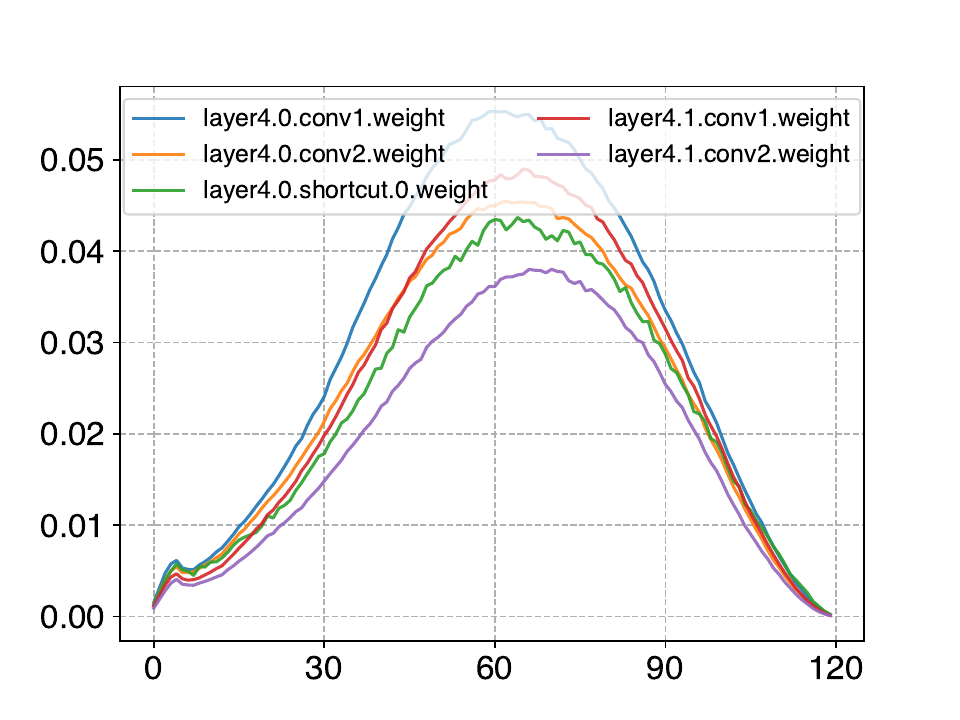}
        \caption{layer4}
    \end{subtable}
    \caption{Epoch-wise flip rate for $\gamma=1$~(kaiming normal).}

    \centering
    \begin{subtable}[h]{0.24\textwidth}
        \includegraphics[width=1\linewidth]{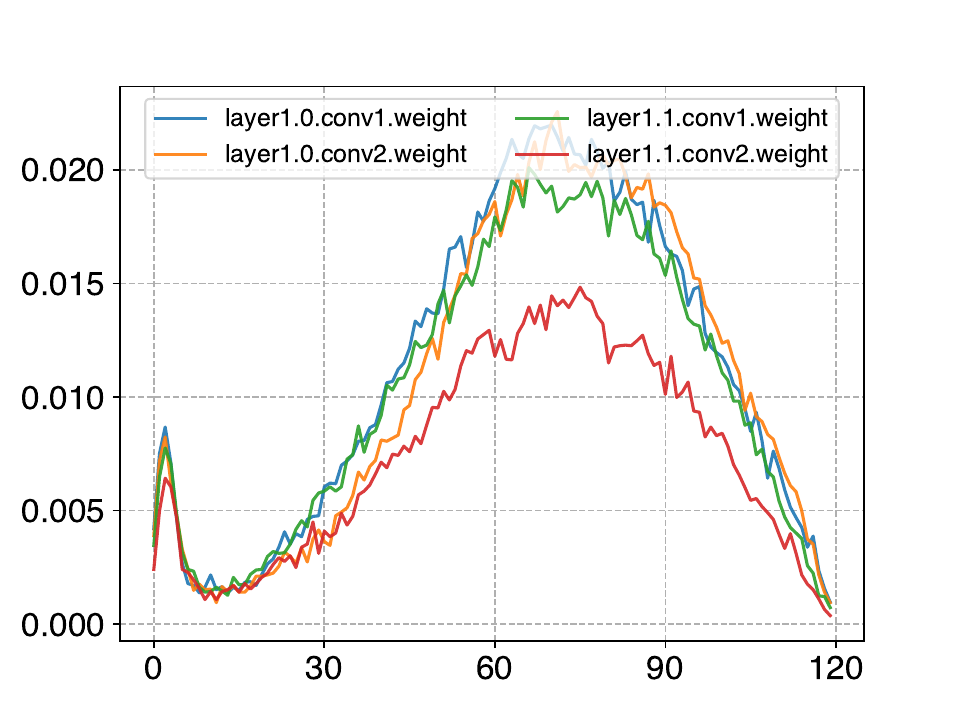}
        \caption{layer1}
    \end{subtable}
    \begin{subtable}[h]{0.24\textwidth}
        \includegraphics[width=1\linewidth]{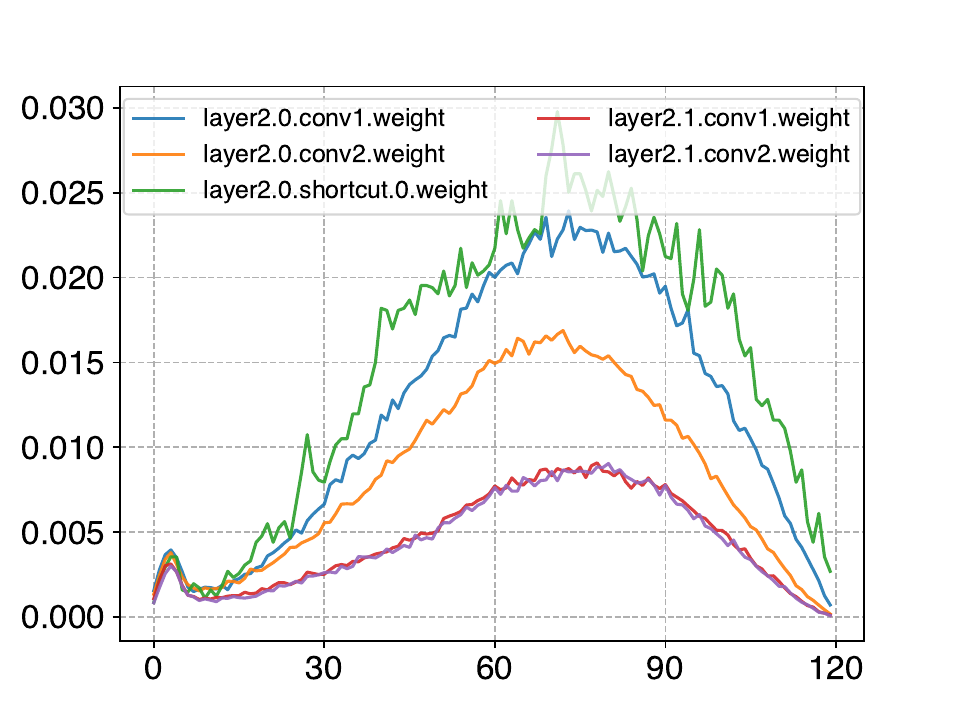}
        \caption{layer2}
    \end{subtable}
        \begin{subtable}[h]{0.24\textwidth}
        \includegraphics[width=1\linewidth]{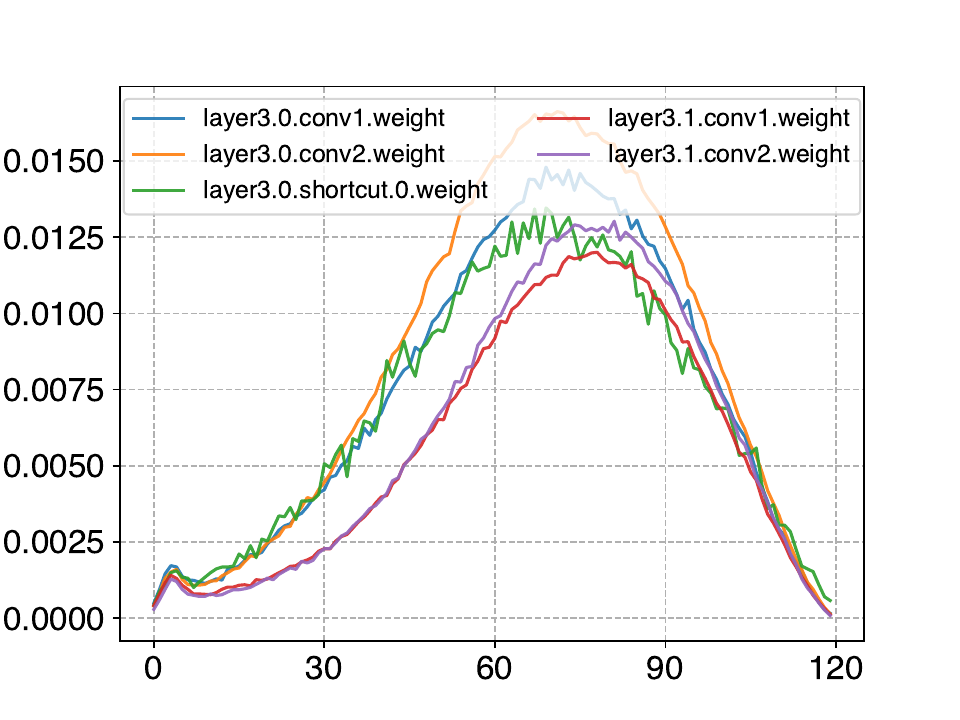}
        \caption{layer3}
    \end{subtable}
        \begin{subtable}[h]{0.24\textwidth}
        \includegraphics[width=1\linewidth]{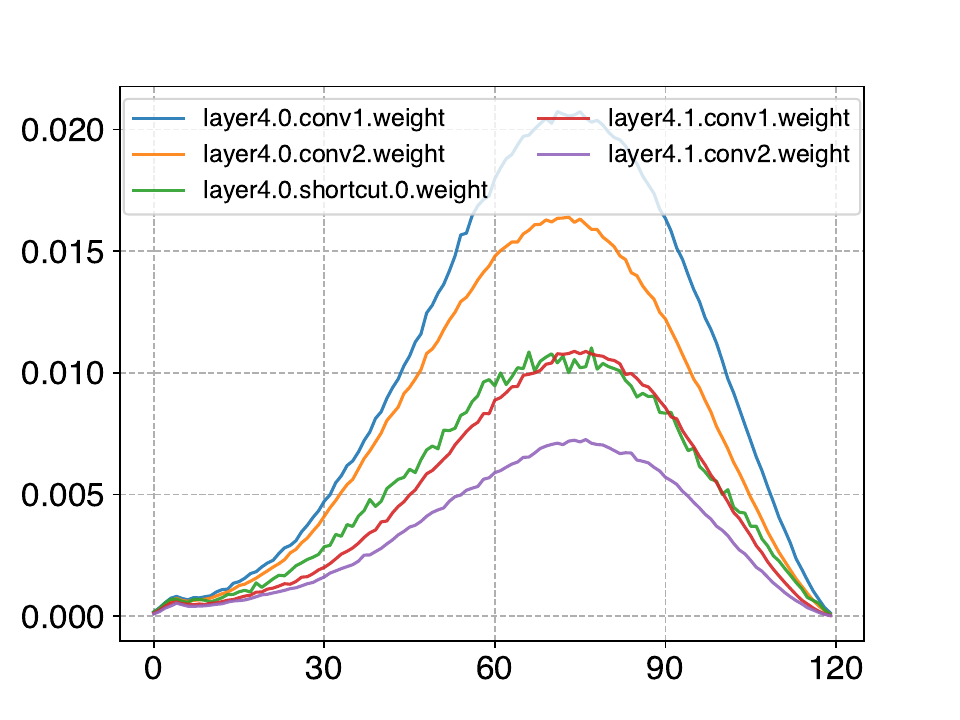}
        \caption{layer4}
    \end{subtable}
    \caption{Epoch-wise flip rate for $\gamma=10$~(kaiming normal).}

    \centering
    \begin{subtable}[h]{0.24\textwidth}
        \includegraphics[width=1\linewidth]{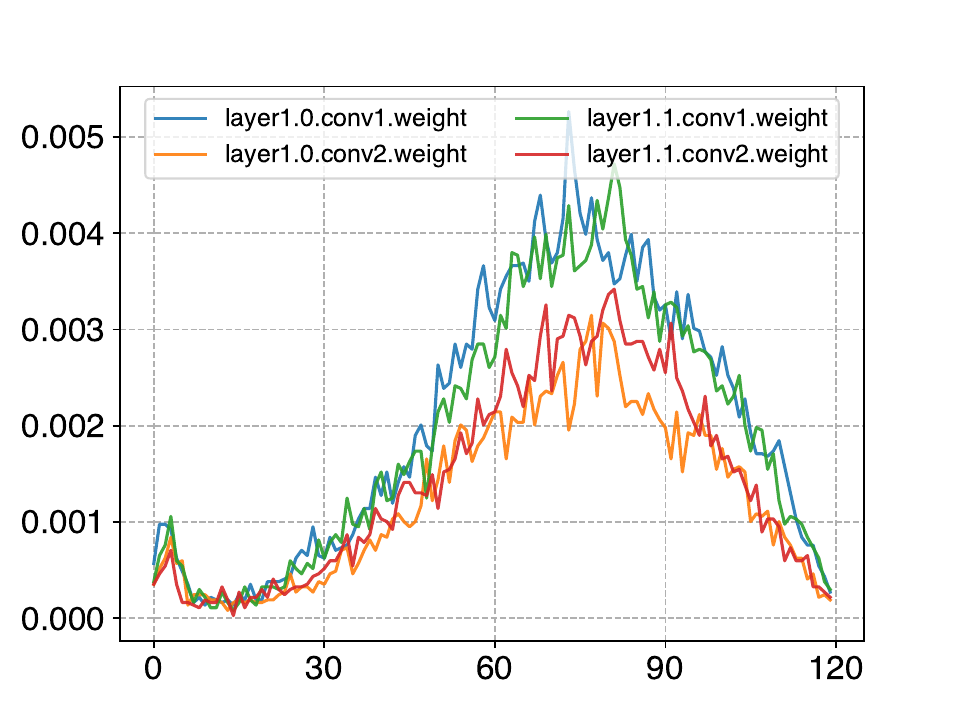}
        \caption{layer1}
    \end{subtable}
    \begin{subtable}[h]{0.24\textwidth}
        \includegraphics[width=1\linewidth]{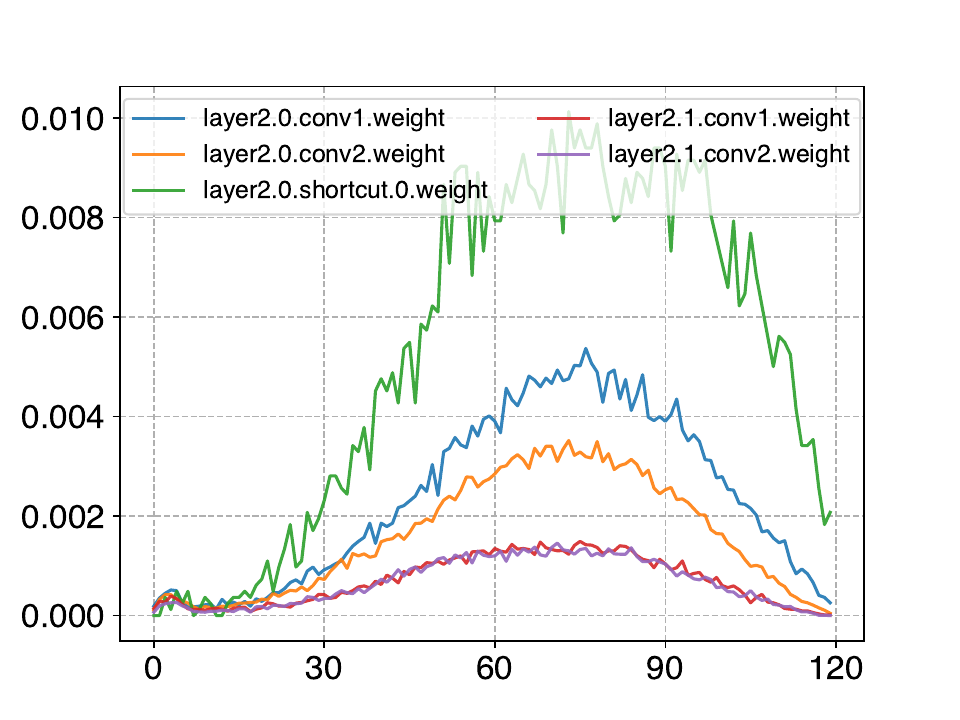}
        \caption{layer2}
    \end{subtable}
        \begin{subtable}[h]{0.24\textwidth}
        \includegraphics[width=1\linewidth]{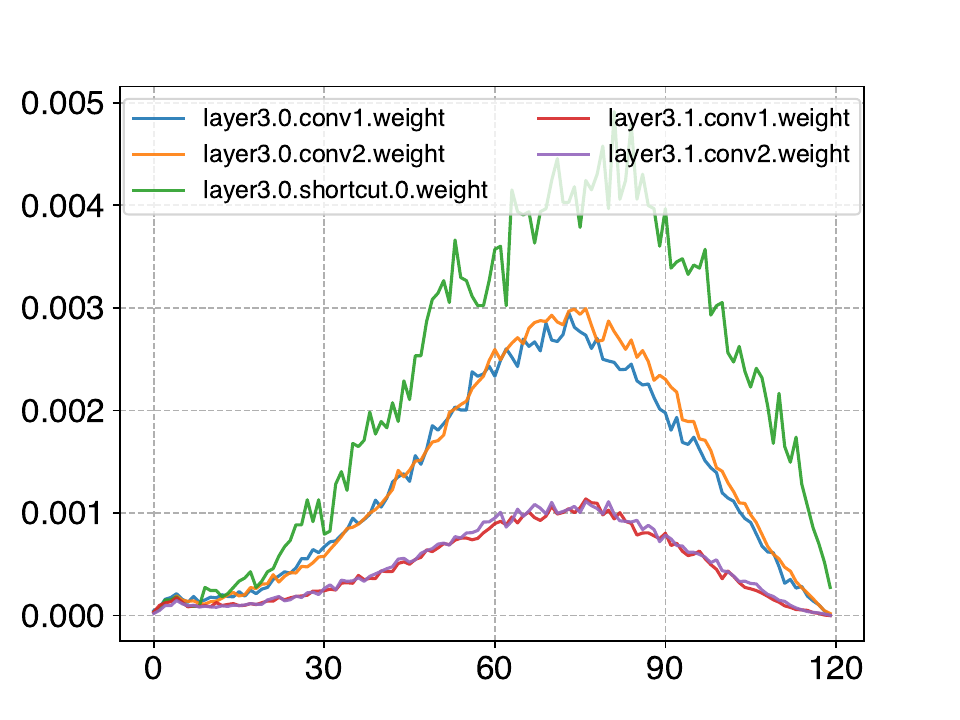}
        \caption{layer3}
    \end{subtable}
        \begin{subtable}[h]{0.24\textwidth}
        \includegraphics[width=1\linewidth]{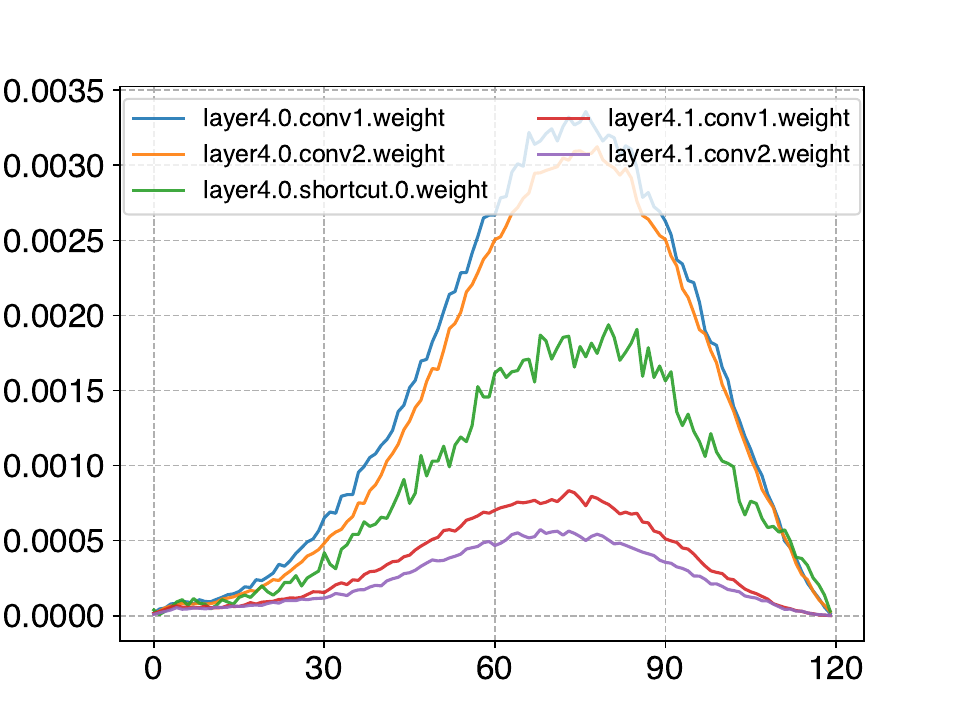}
        \caption{layer4}
    \end{subtable}
    \caption{Epoch-wise flip rate for $\gamma=100$~(kaiming normal).}

{
    \centering
    \begin{subtable}[h]{0.24\textwidth}
        \includegraphics[width=1\linewidth]{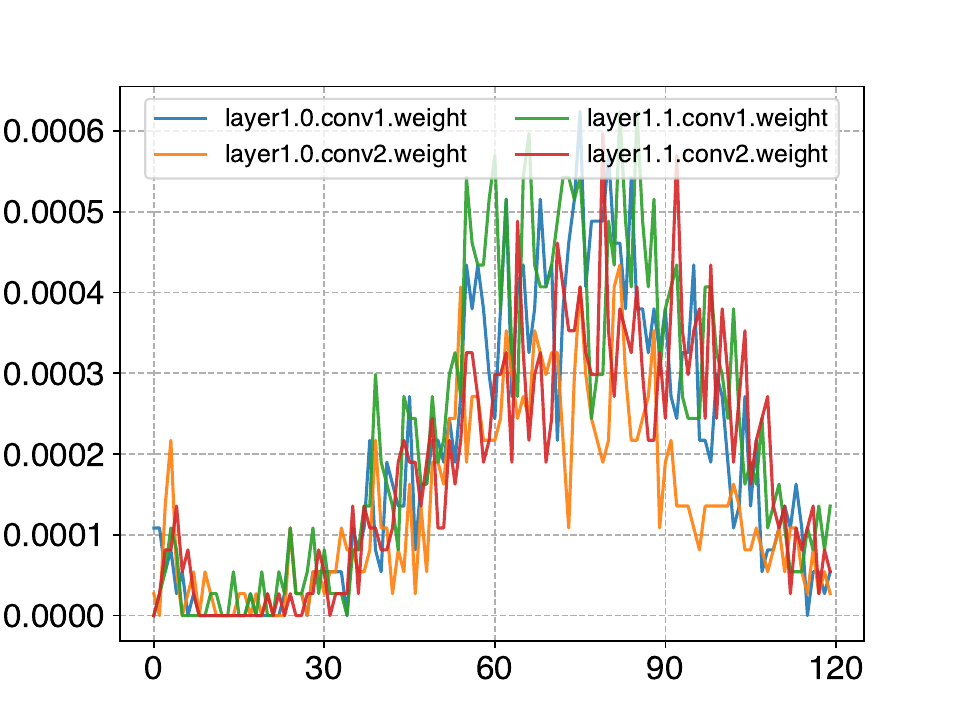}
        \caption{layer1}
    \end{subtable}
    \begin{subtable}[h]{0.24\textwidth}
        \includegraphics[width=1\linewidth]{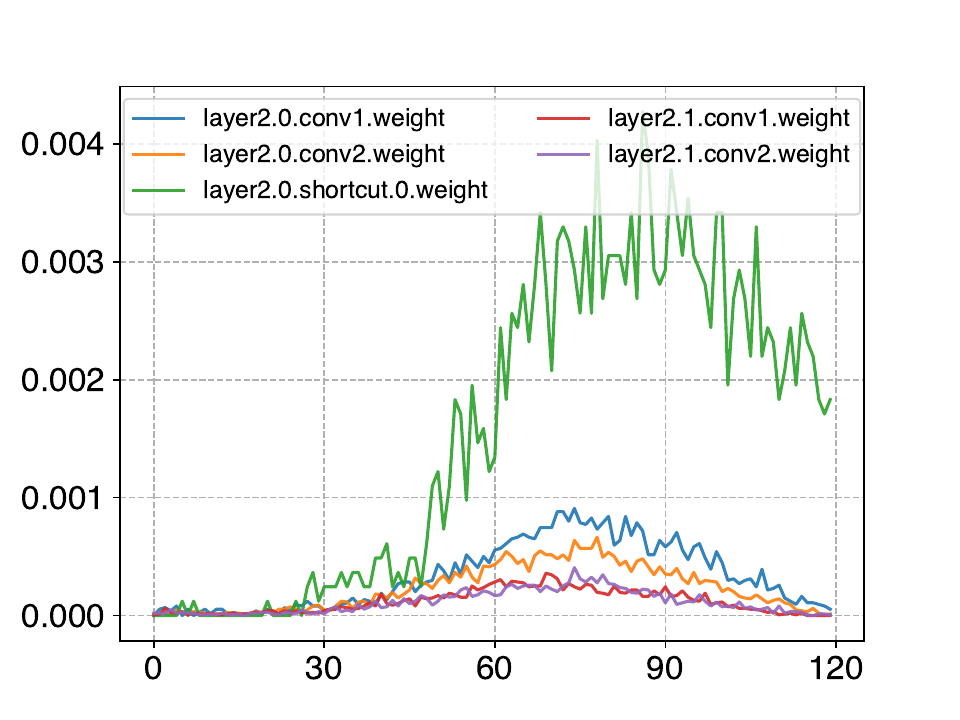}
        \caption{layer2}
    \end{subtable}
        \begin{subtable}[h]{0.24\textwidth}
        \includegraphics[width=1\linewidth]{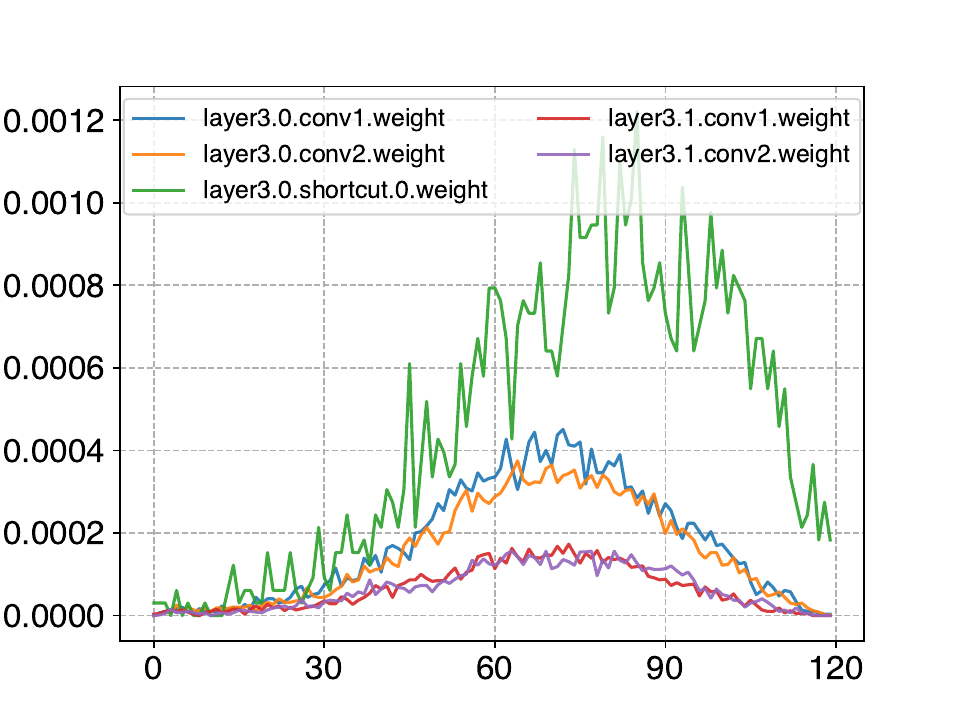}
        \caption{layer3}
    \end{subtable}
        \begin{subtable}[h]{0.24\textwidth}
        \includegraphics[width=1\linewidth]{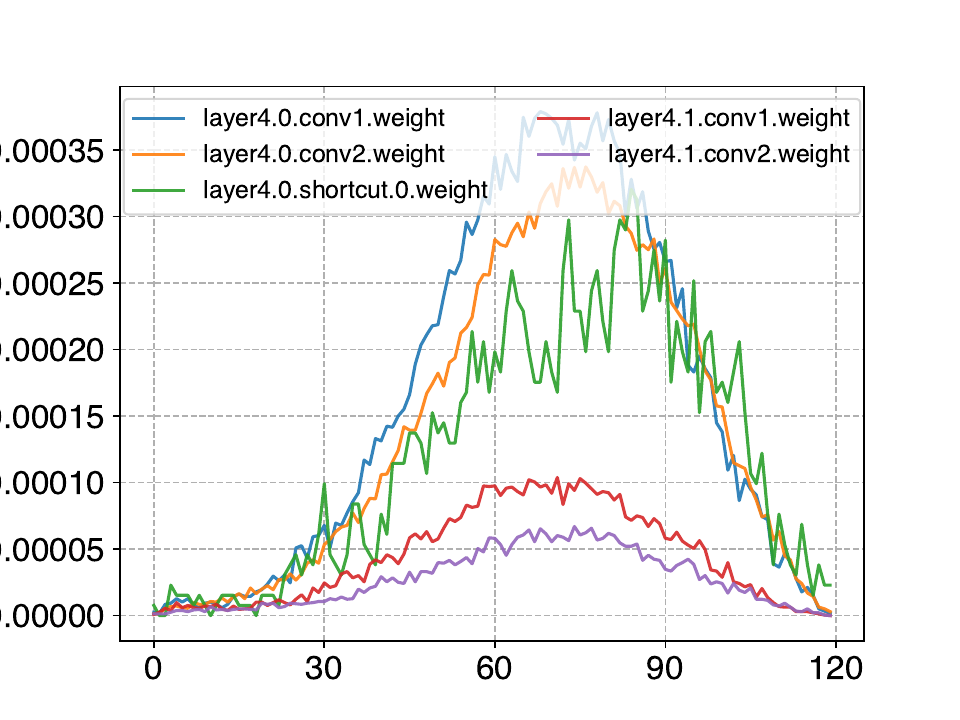}
        \caption{layer4}
    \end{subtable}
    \caption{Epoch-wise flip rate for $\gamma=1000$~(kaiming normal).}
    \label{fig:normal_1000}
}
\end{figure*}

\begin{figure*}[p]
{
    \centering
    \begin{subtable}[h]{0.24\textwidth}
        \includegraphics[width=1\linewidth]{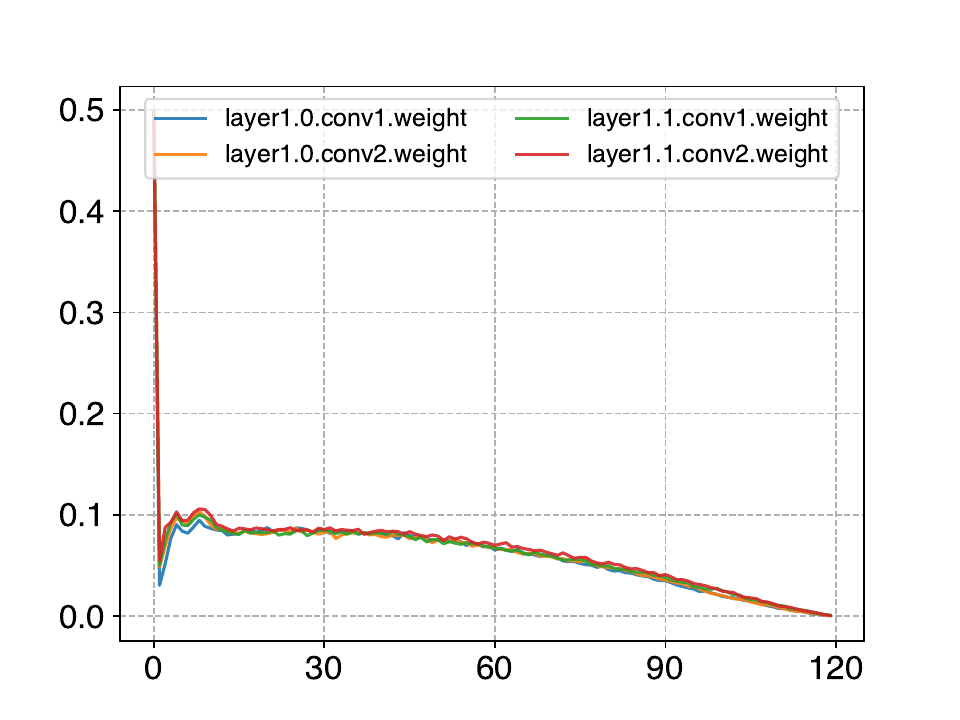}
        \caption{layer1}
    \end{subtable}
    \begin{subtable}[h]{0.24\textwidth}
        \includegraphics[width=1\linewidth]{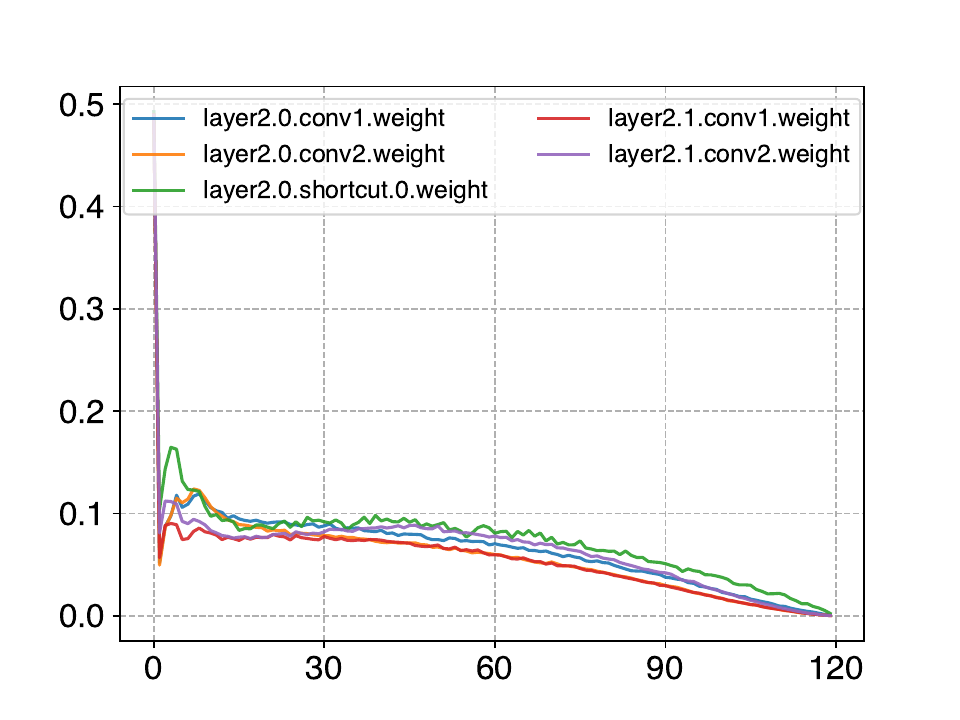}
        \caption{layer2}
    \end{subtable}
        \begin{subtable}[h]{0.24\textwidth}
        \includegraphics[width=1\linewidth]{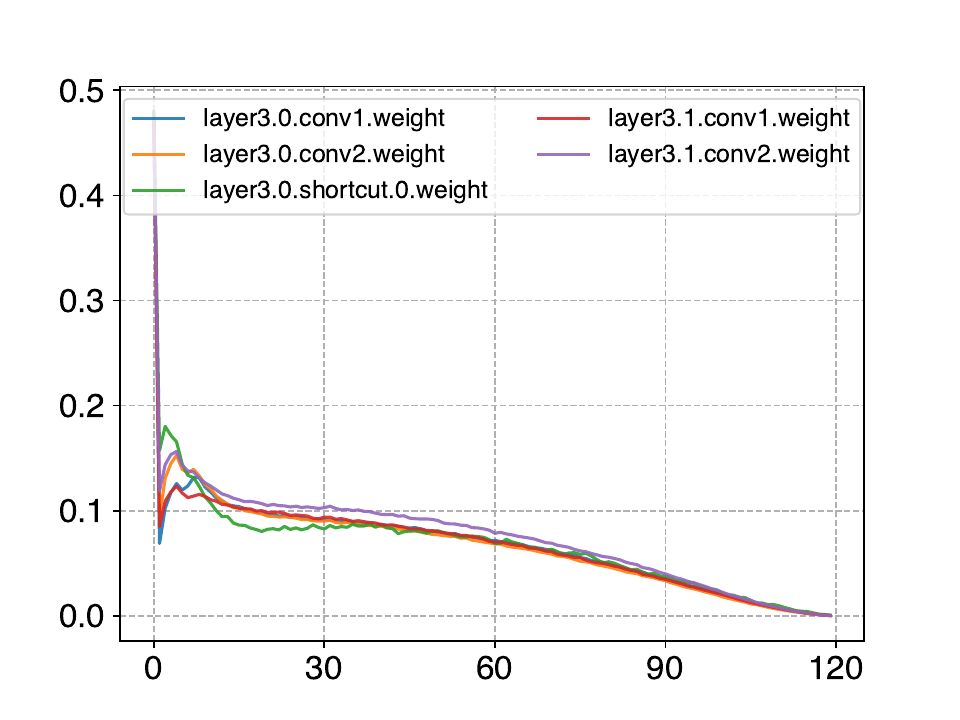}
        \caption{layer3}
    \end{subtable}
        \begin{subtable}[h]{0.24\textwidth}
        \includegraphics[width=1\linewidth]{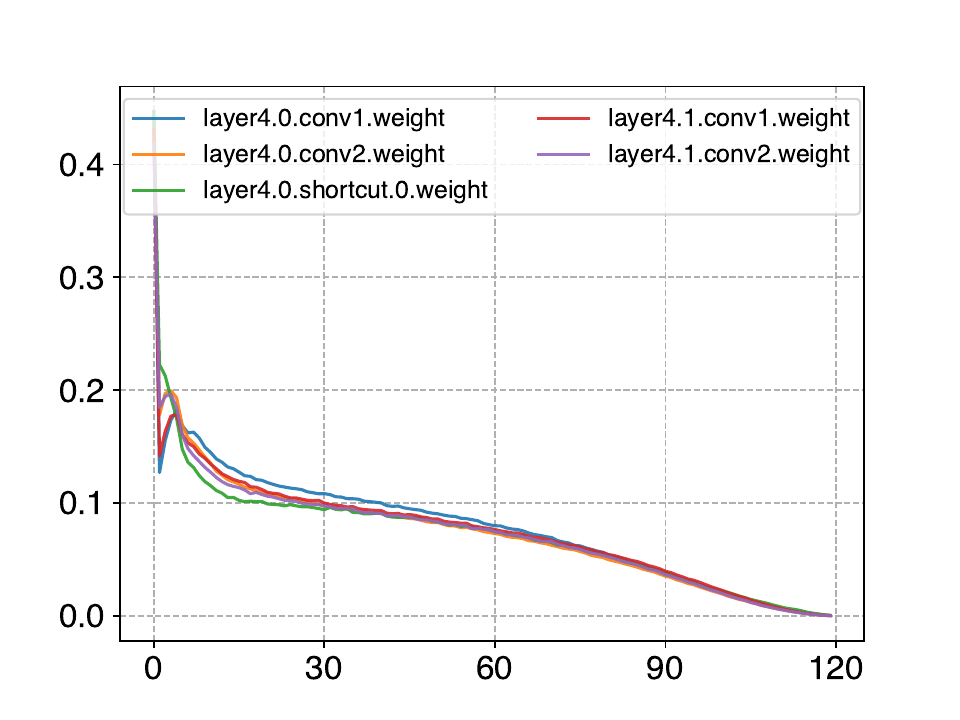}
        \caption{layer4}
    \end{subtable}
    \caption{Epoch-wise flip rate for $\gamma=0.0001$~(kaiming uniform).}
    \label{fig:uniform_0.0001}
    
}

    \centering
    \begin{subtable}[h]{0.24\textwidth}
        \includegraphics[width=1\linewidth]{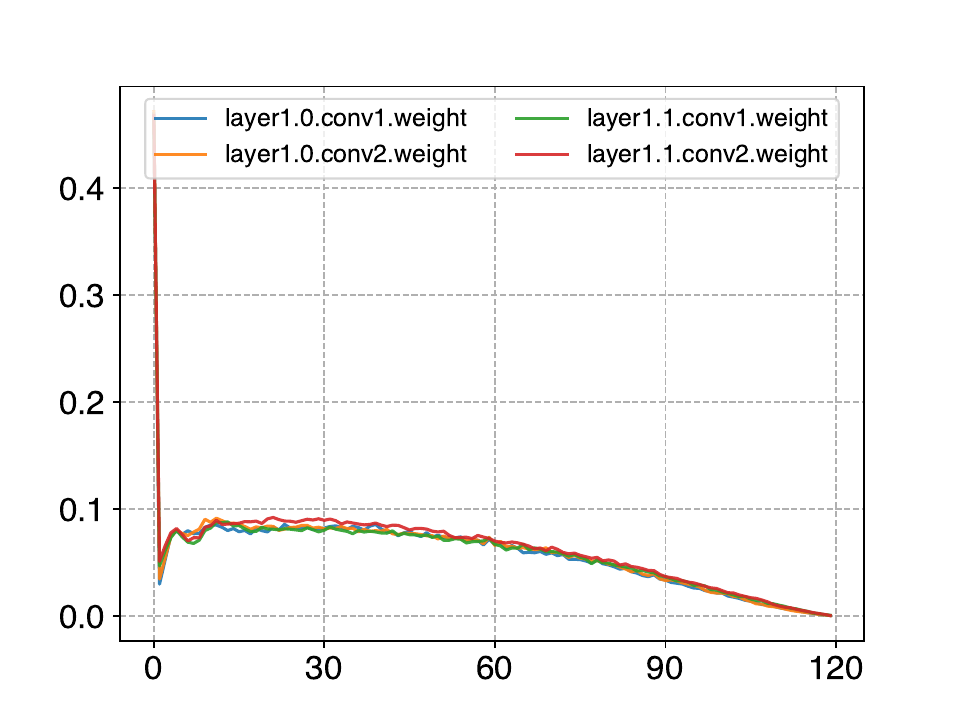}
        \caption{layer1}
    \end{subtable}
    \begin{subtable}[h]{0.24\textwidth}
        \includegraphics[width=1\linewidth]{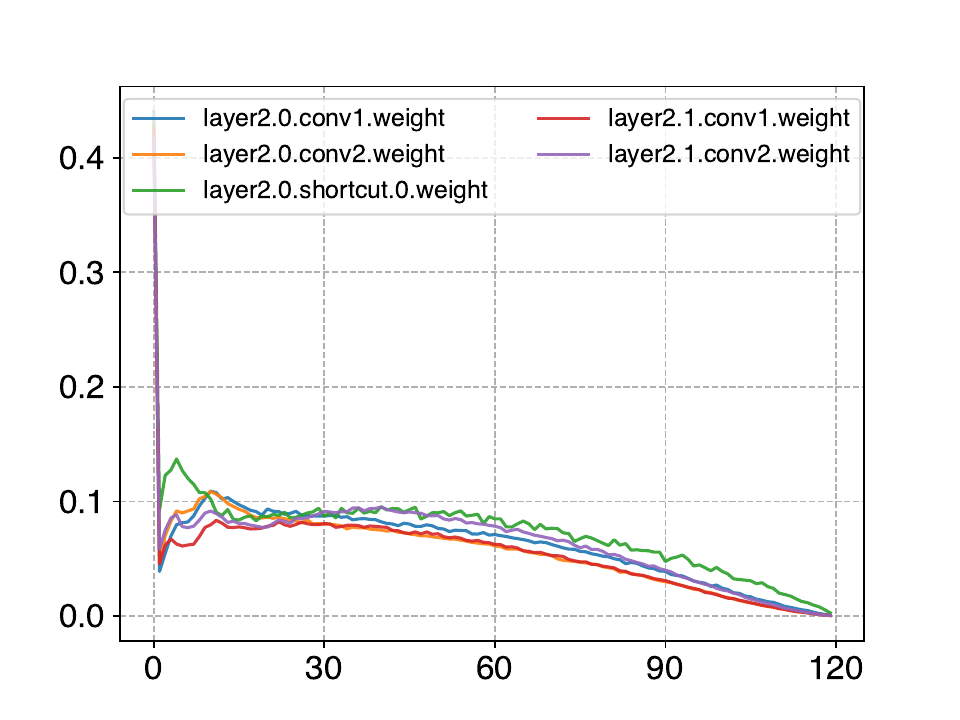}
        \caption{layer2}
    \end{subtable}
        \begin{subtable}[h]{0.24\textwidth}
        \includegraphics[width=1\linewidth]{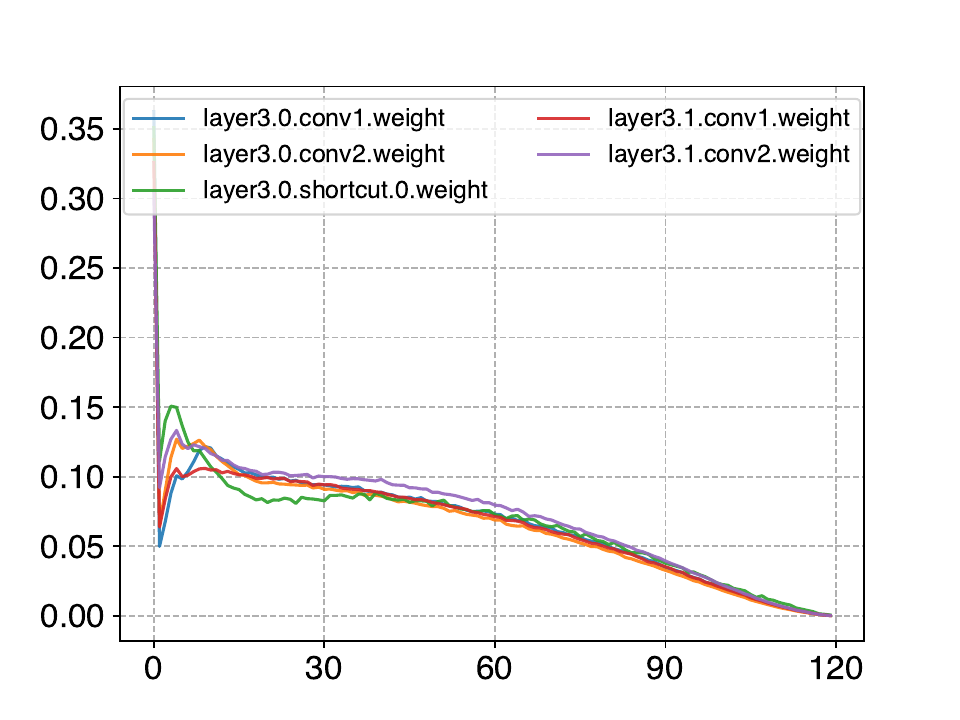}
        \caption{layer3}
    \end{subtable}
        \begin{subtable}[h]{0.24\textwidth}
        \includegraphics[width=1\linewidth]{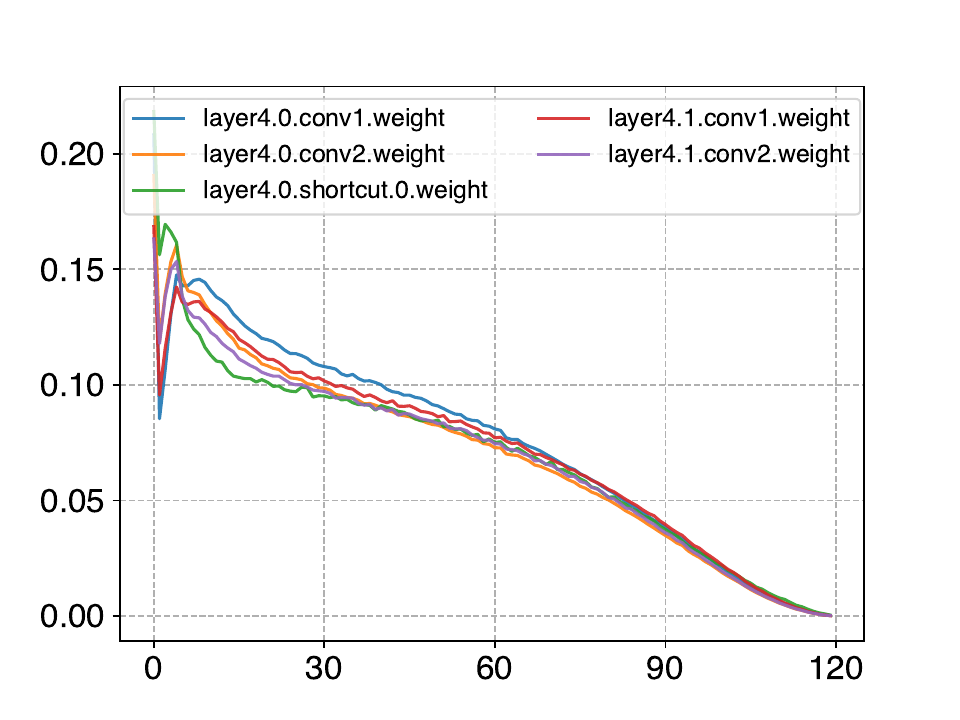}
        \caption{layer4}
    \end{subtable}
    \caption{Epoch-wise flip rate for $\gamma=0.001$~(kaiming uniform).}

    \centering
    \begin{subtable}[h]{0.24\textwidth}
        \includegraphics[width=1\linewidth]{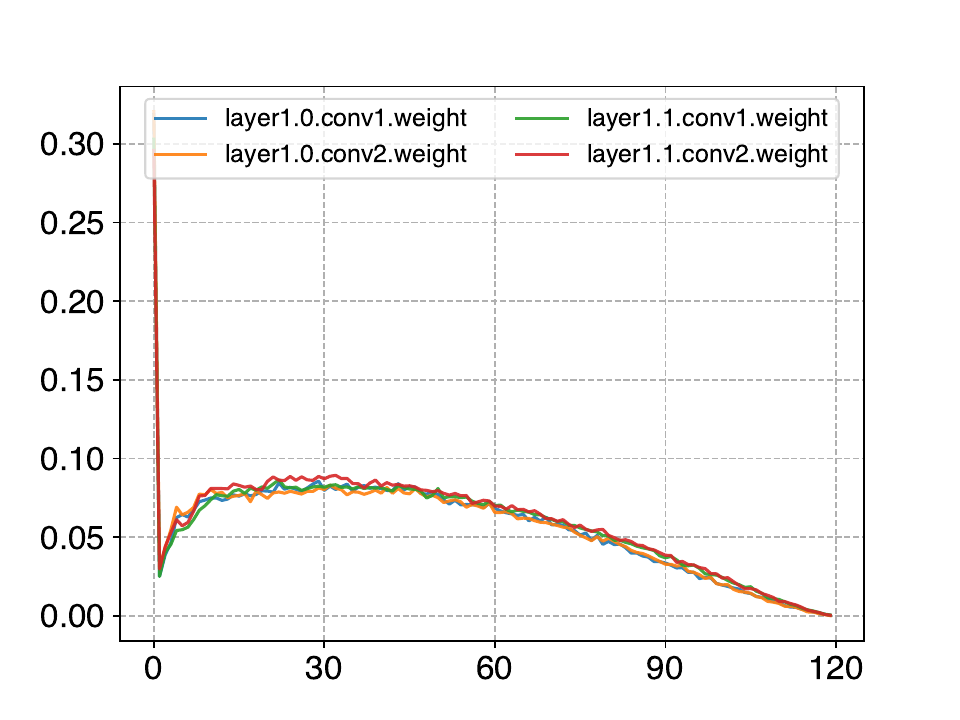}
        \caption{layer1}
    \end{subtable}
    \begin{subtable}[h]{0.24\textwidth}
        \includegraphics[width=1\linewidth]{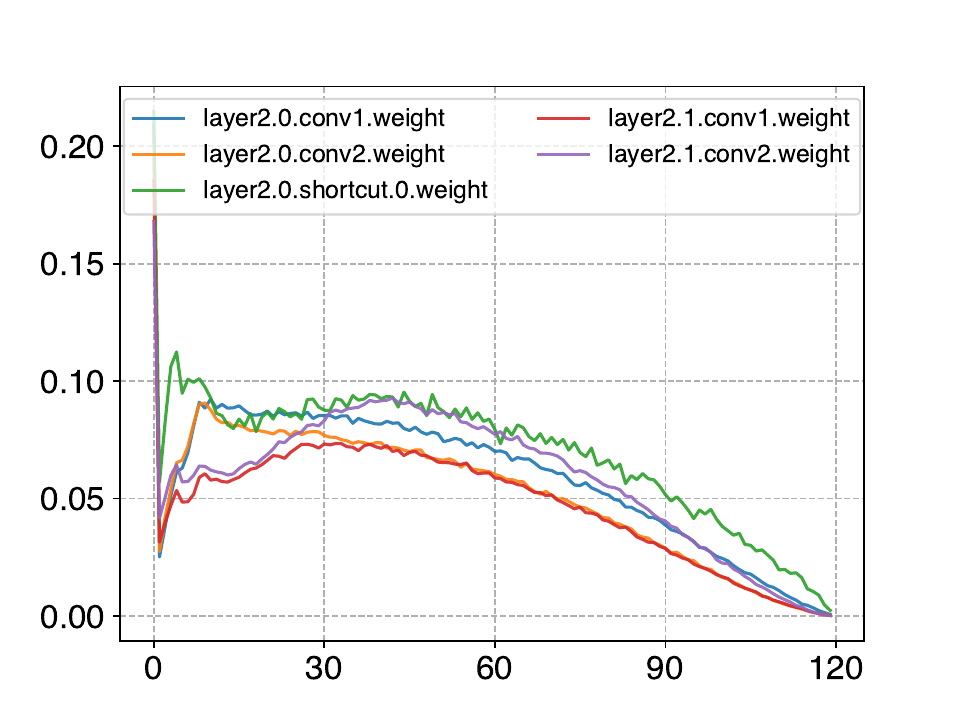}
        \caption{layer2}
    \end{subtable}
        \begin{subtable}[h]{0.24\textwidth}
        \includegraphics[width=1\linewidth]{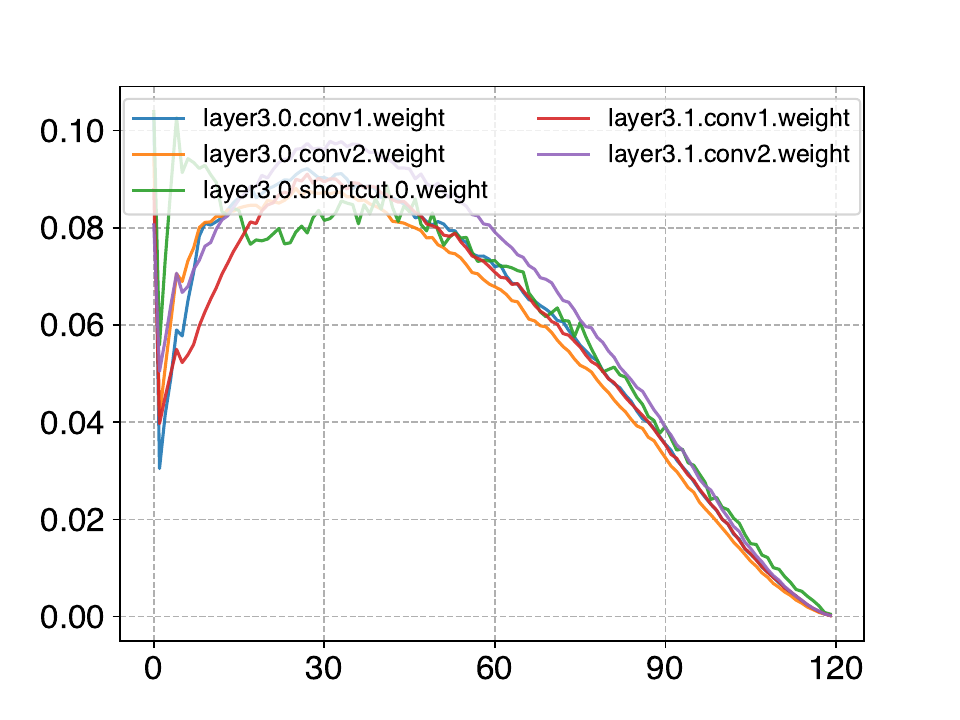}
        \caption{layer3}
    \end{subtable}
        \begin{subtable}[h]{0.24\textwidth}
        \includegraphics[width=1\linewidth]{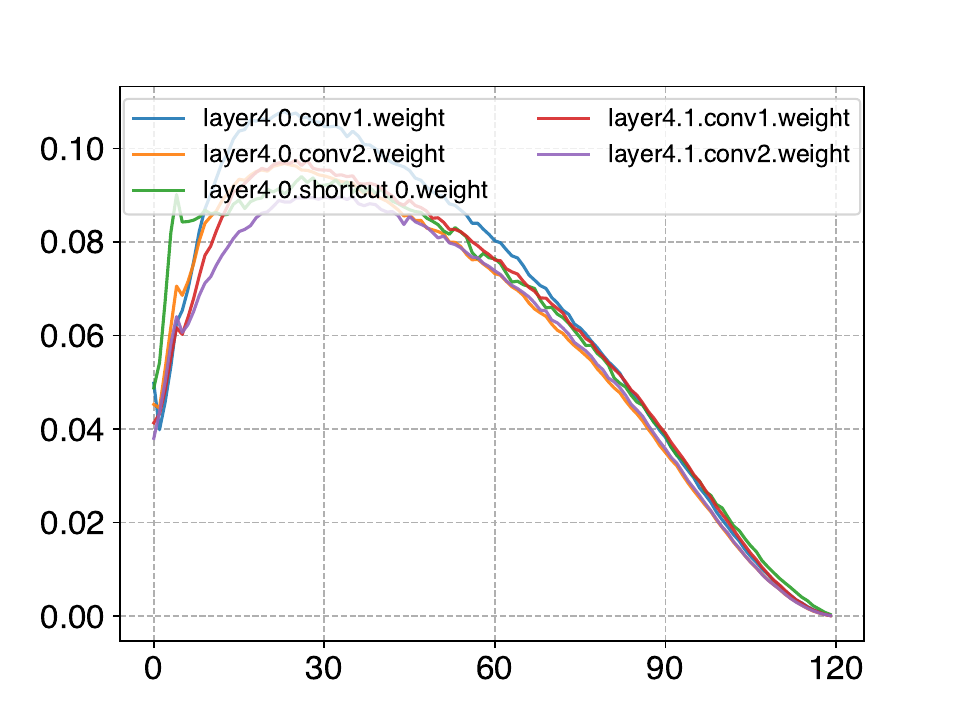}
        \caption{layer4}
    \end{subtable}
    \caption{Epoch-wise flip rate for $\gamma=0.01$~(kaiming uniform).}

    \centering
    \begin{subtable}[h]{0.24\textwidth}
        \includegraphics[width=1\linewidth]{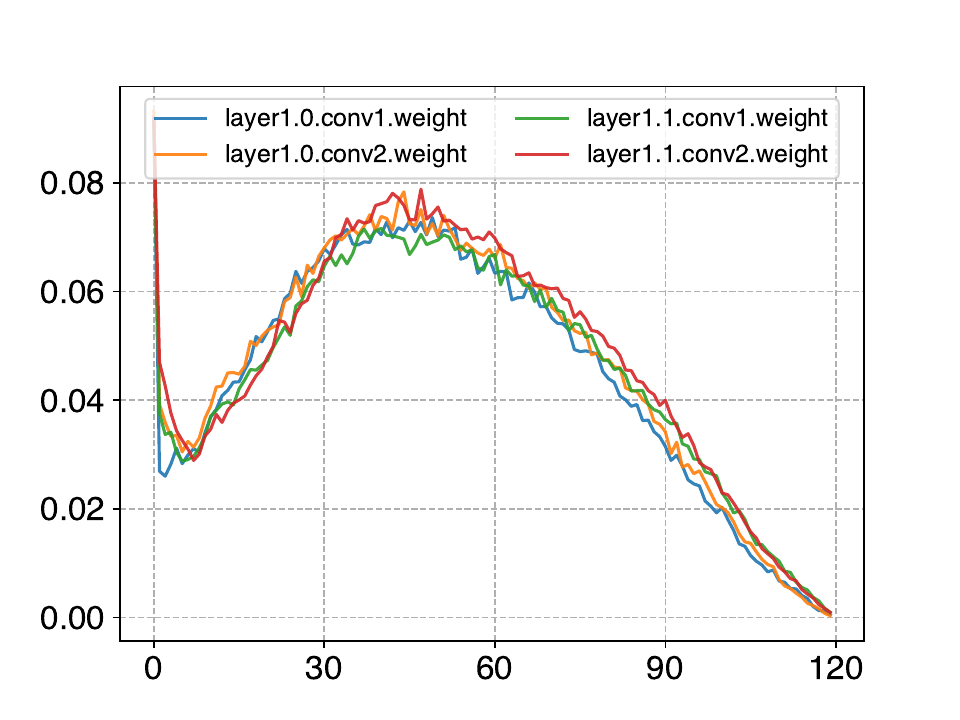}
        \caption{layer1}
    \end{subtable}
    \begin{subtable}[h]{0.24\textwidth}
        \includegraphics[width=1\linewidth]{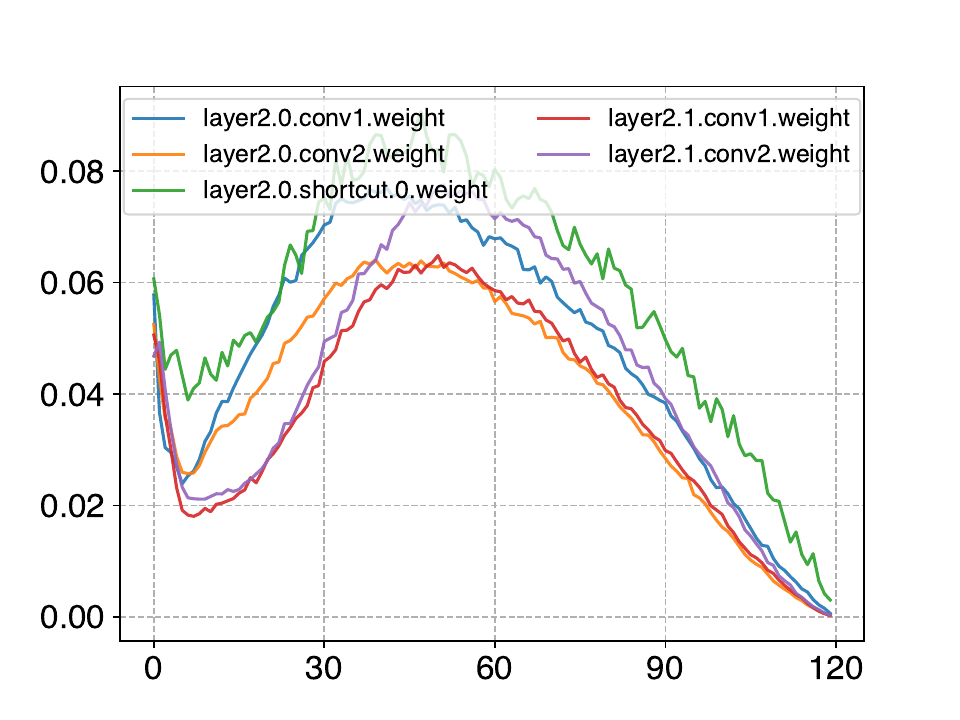}
        \caption{layer2}
    \end{subtable}
        \begin{subtable}[h]{0.24\textwidth}
        \includegraphics[width=1\linewidth]{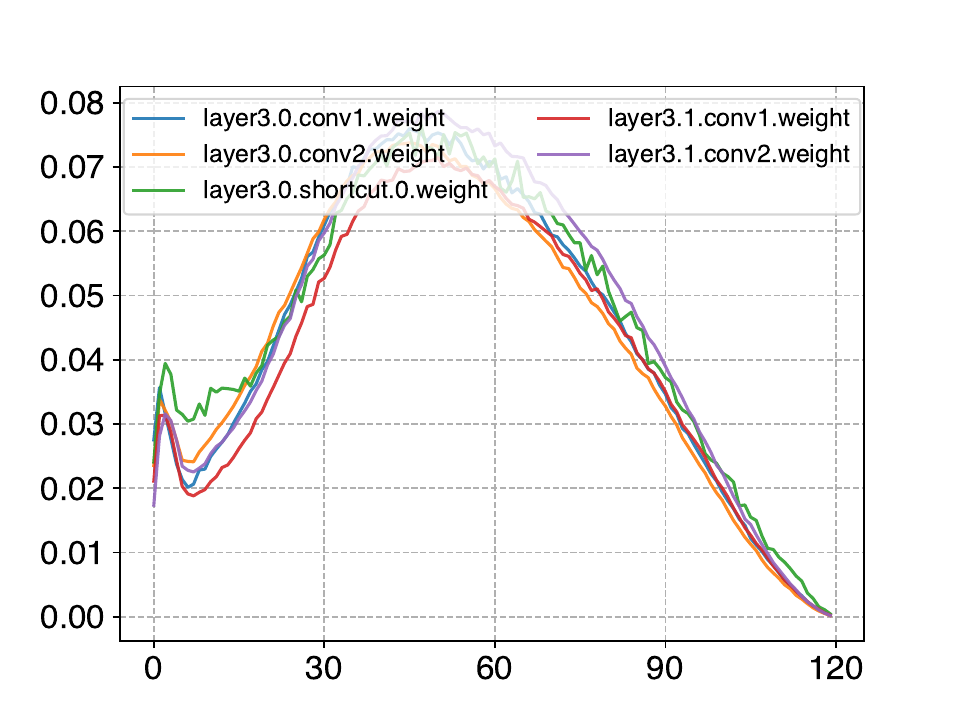}
        \caption{layer3}
    \end{subtable}
        \begin{subtable}[h]{0.24\textwidth}
        \includegraphics[width=1\linewidth]{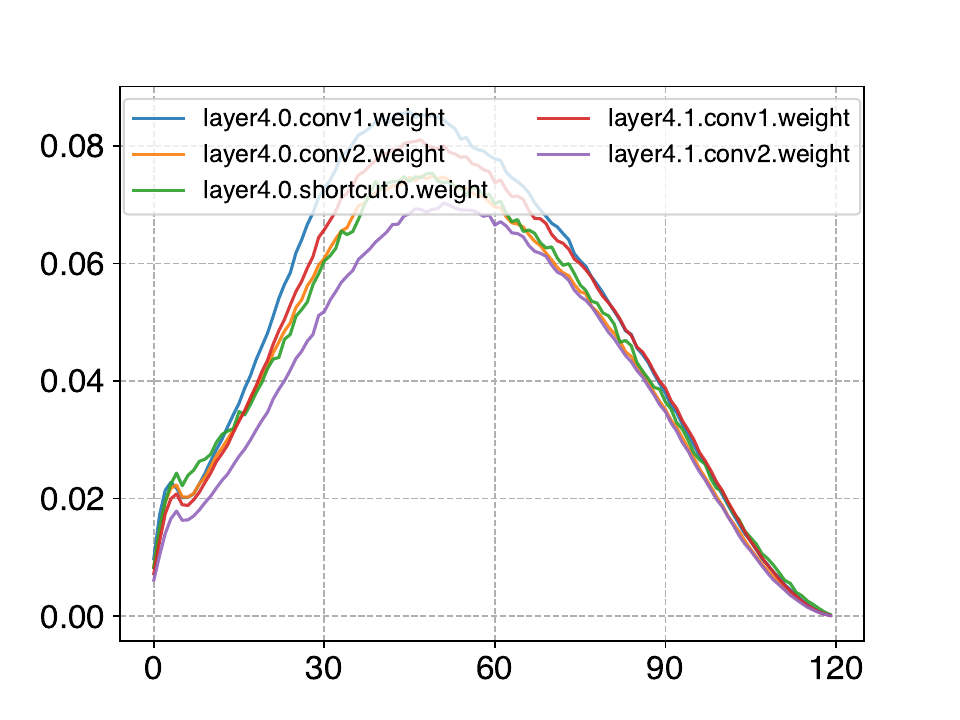}
        \caption{layer4}
    \end{subtable}
    \caption{Epoch-wise flip rate for $\gamma=0.1$~(kaiming uniform).}
\end{figure*}

\begin{figure*}[p]
    \centering
    \begin{subtable}[h]{0.24\textwidth}
        \includegraphics[width=1\linewidth]{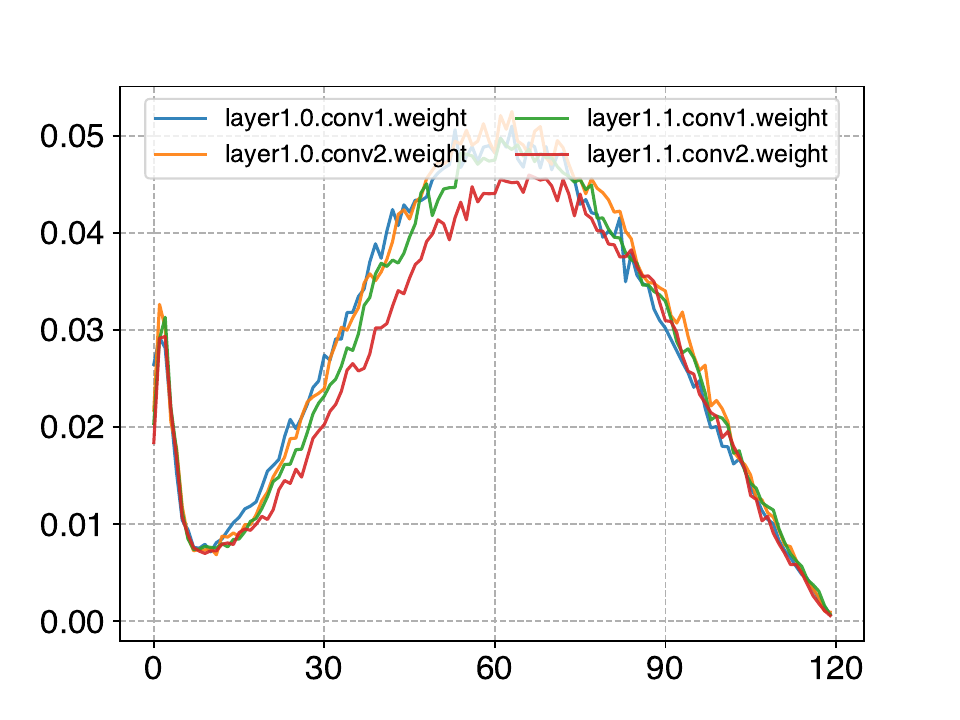}
        \caption{layer1}
    \end{subtable}
    \begin{subtable}[h]{0.24\textwidth}
        \includegraphics[width=1\linewidth]{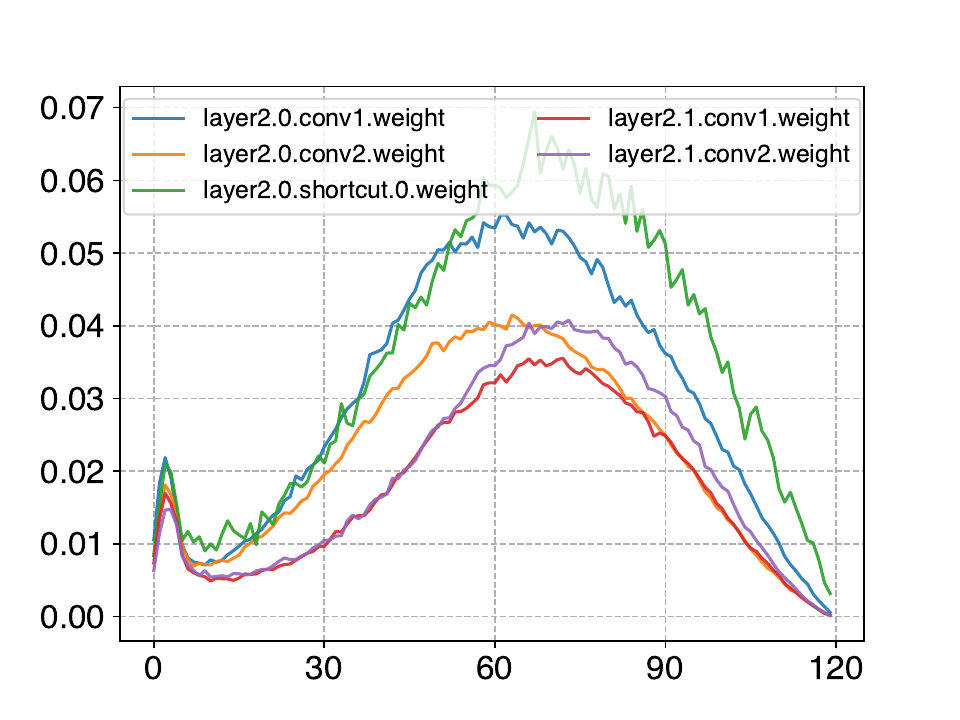}
        \caption{layer2}
    \end{subtable}
        \begin{subtable}[h]{0.24\textwidth}
        \includegraphics[width=1\linewidth]{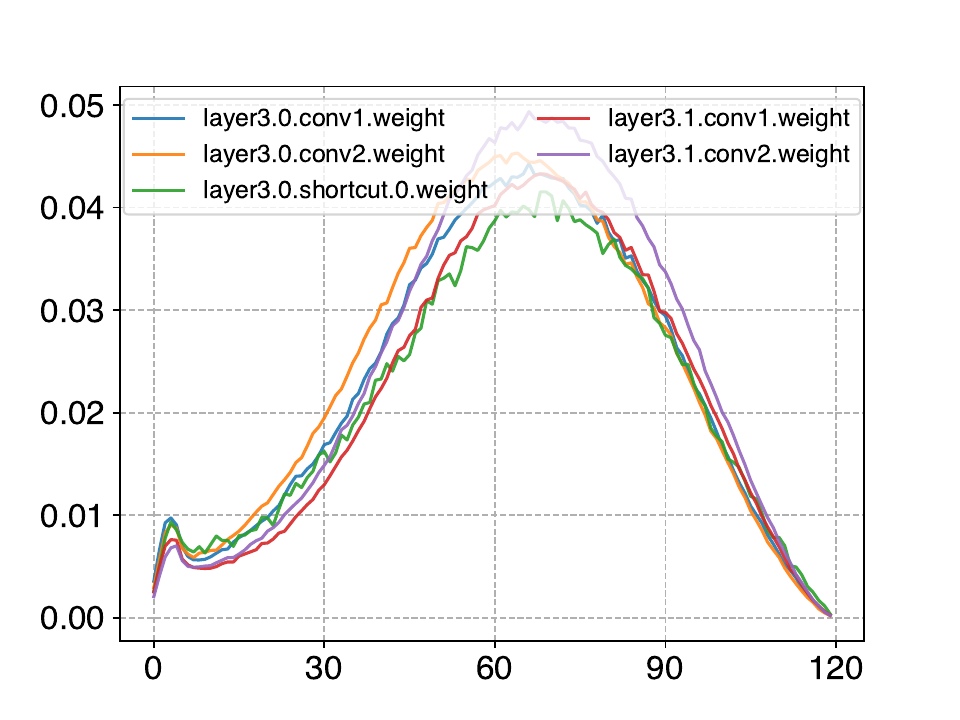}
        \caption{layer3}
    \end{subtable}
        \begin{subtable}[h]{0.24\textwidth}
        \includegraphics[width=1\linewidth]{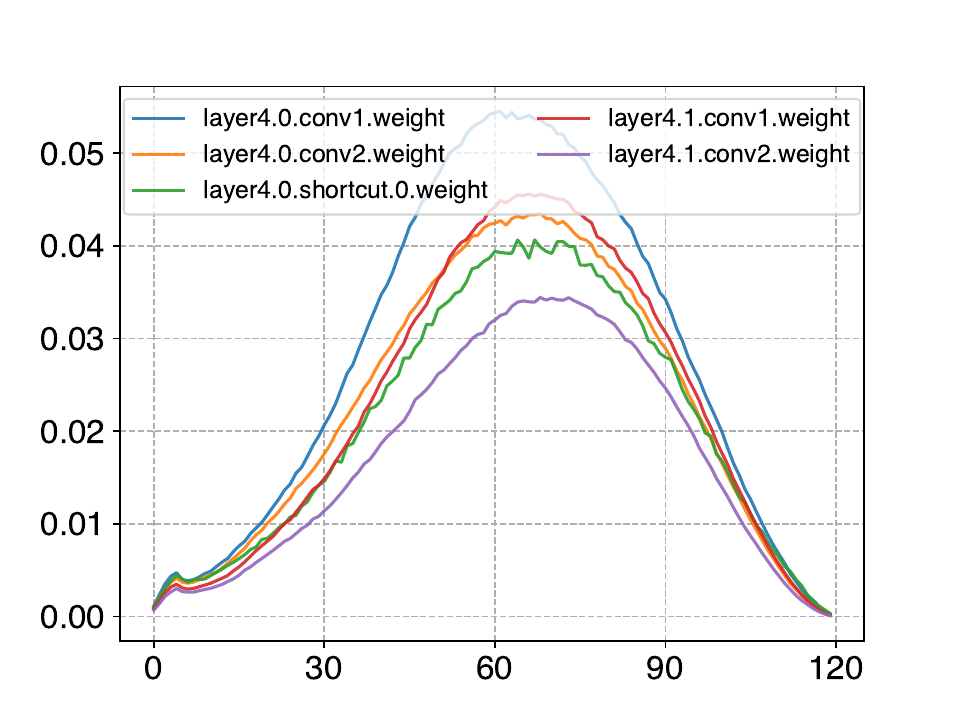}
        \caption{layer4}
    \end{subtable}
    \caption{Epoch-wise flip rate for $\gamma=1$~(kaiming uniform).}

    \centering
    \begin{subtable}[h]{0.24\textwidth}
        \includegraphics[width=1\linewidth]{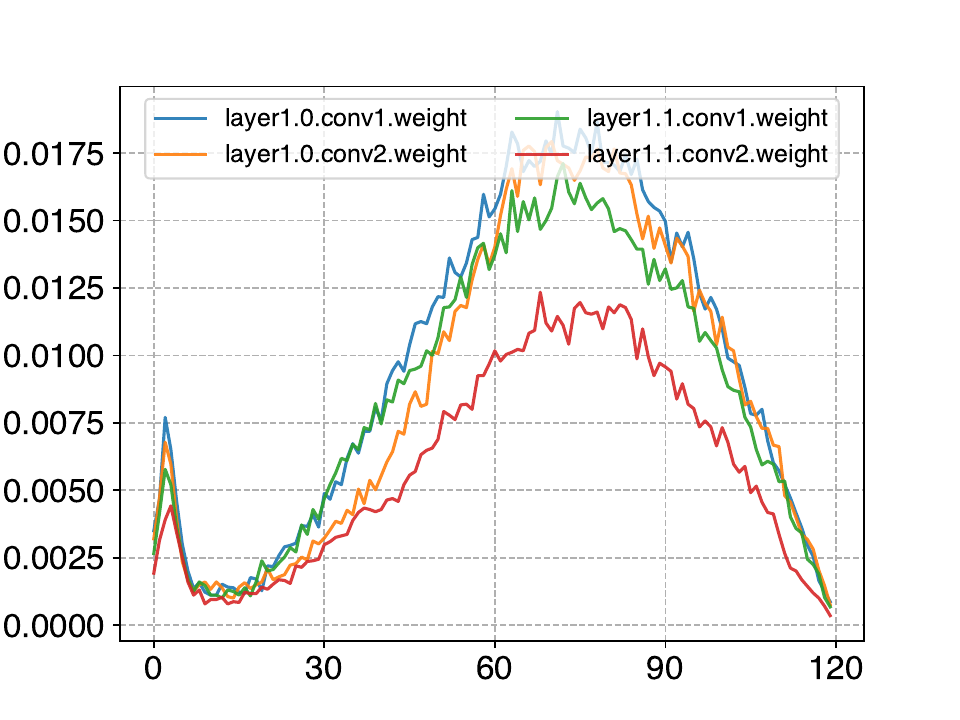}
        \caption{layer1}
    \end{subtable}
    \begin{subtable}[h]{0.24\textwidth}
        \includegraphics[width=1\linewidth]{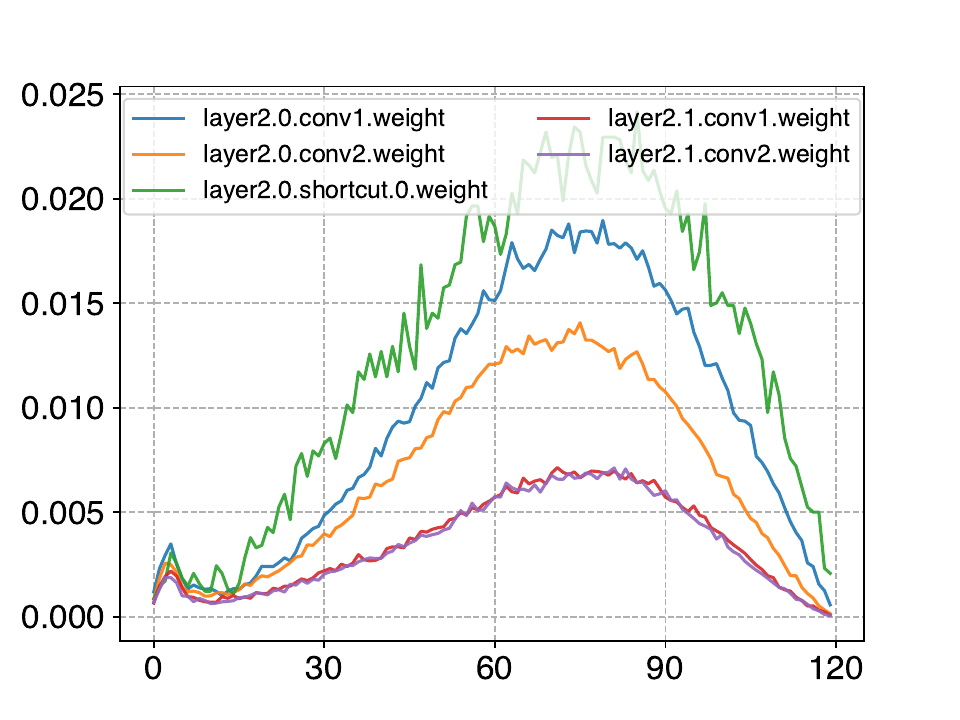}
        \caption{layer2}
    \end{subtable}
        \begin{subtable}[h]{0.24\textwidth}
        \includegraphics[width=1\linewidth]{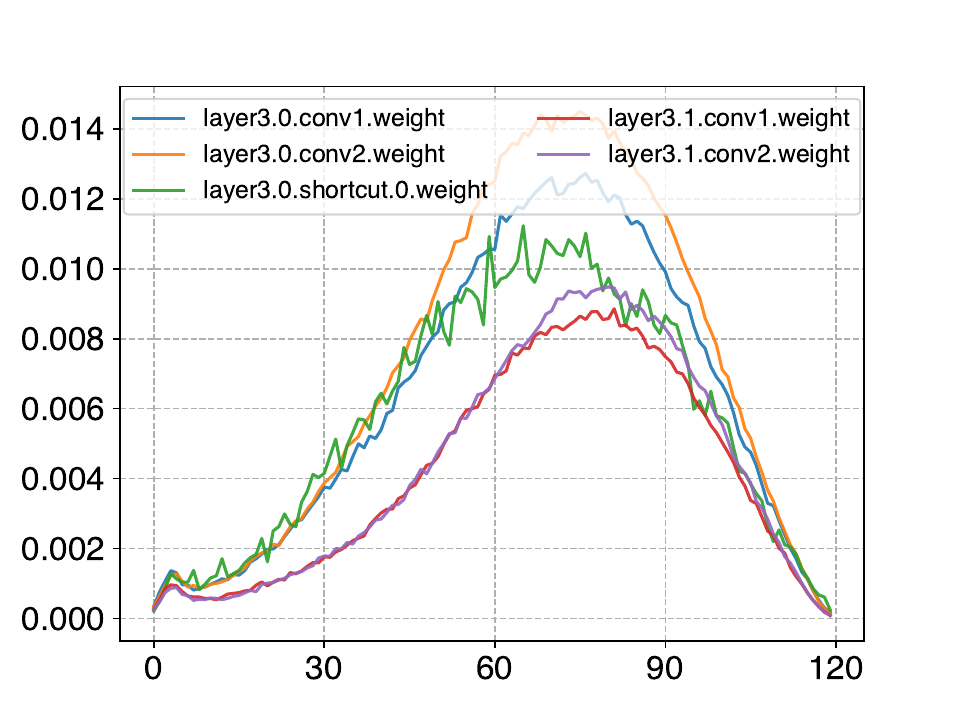}
        \caption{layer3}
    \end{subtable}
        \begin{subtable}[h]{0.24\textwidth}
        \includegraphics[width=1\linewidth]{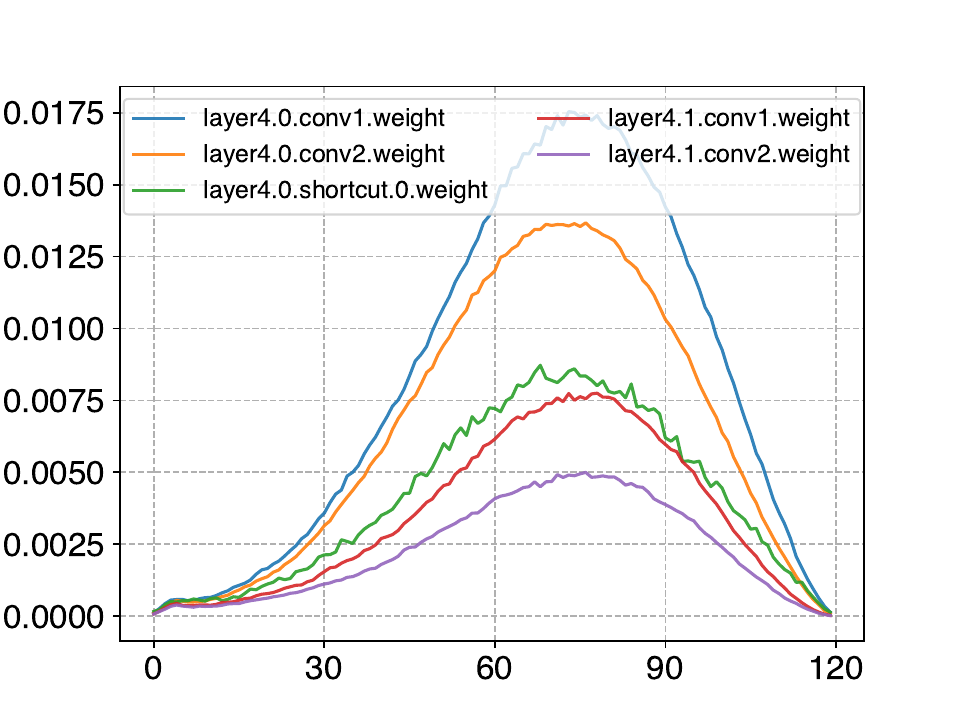}
        \caption{layer4}
    \end{subtable}
    \caption{Epoch-wise flip rate for $\gamma=10$~(kaiming uniform).}

    \centering
    \begin{subtable}[h]{0.24\textwidth}
        \includegraphics[width=1\linewidth]{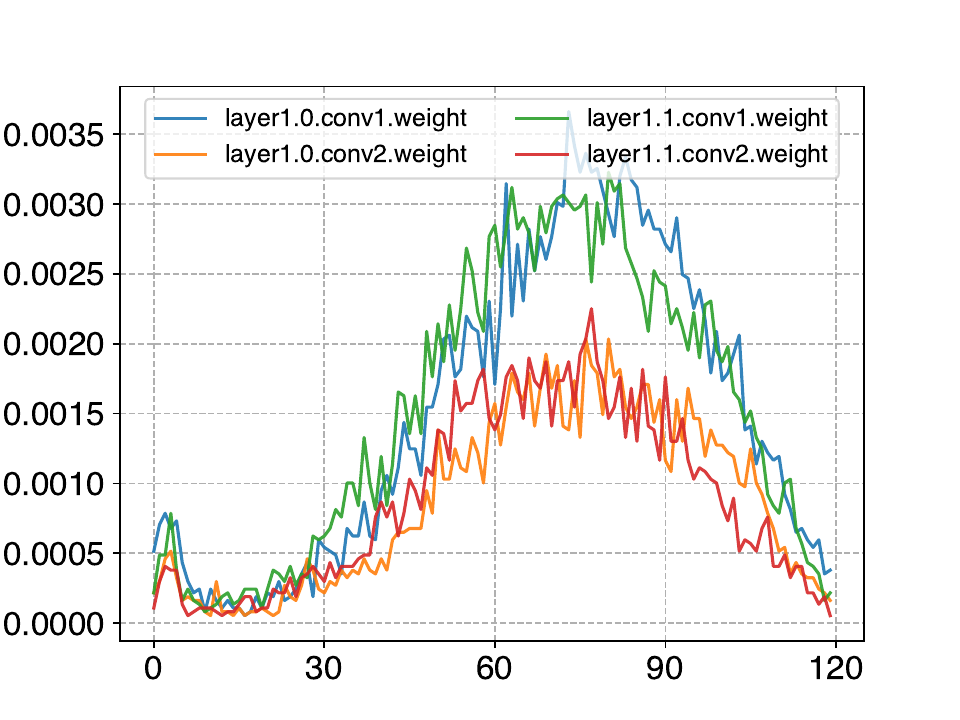}
        \caption{layer1}
    \end{subtable}
    \begin{subtable}[h]{0.24\textwidth}
        \includegraphics[width=1\linewidth]{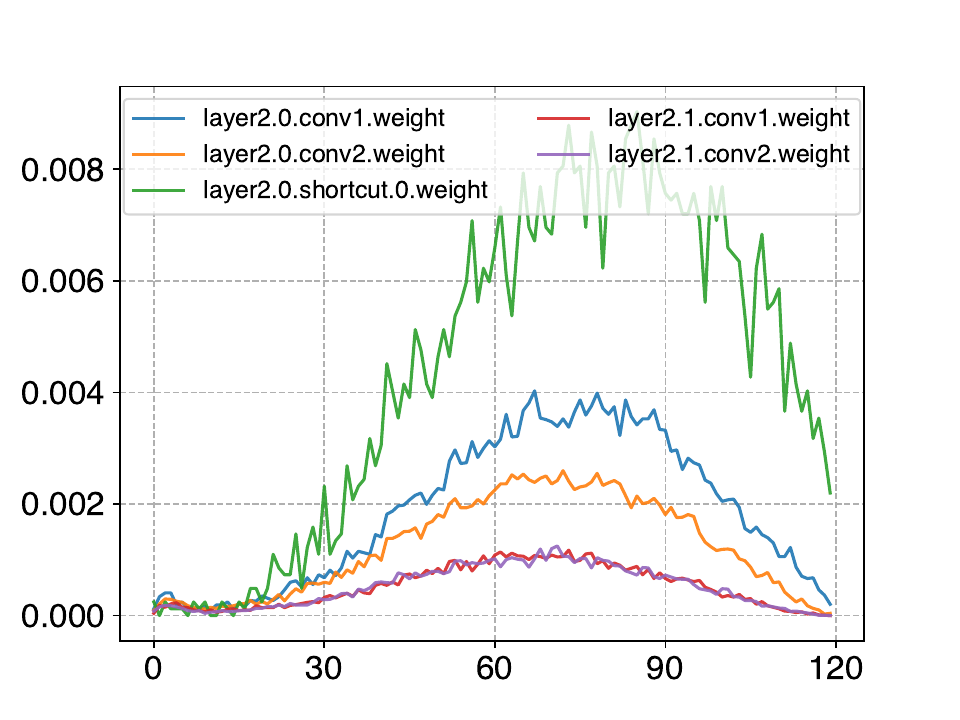}
        \caption{layer2}
    \end{subtable}
        \begin{subtable}[h]{0.24\textwidth}
        \includegraphics[width=1\linewidth]{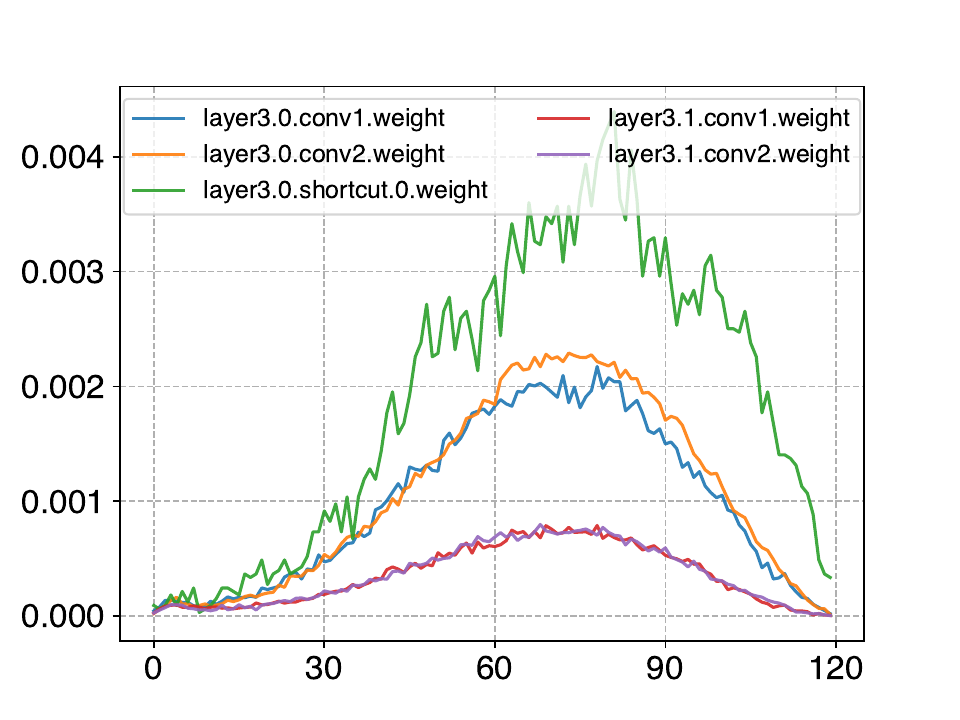}
        \caption{layer3}
    \end{subtable}
        \begin{subtable}[h]{0.24\textwidth}
        \includegraphics[width=1\linewidth]{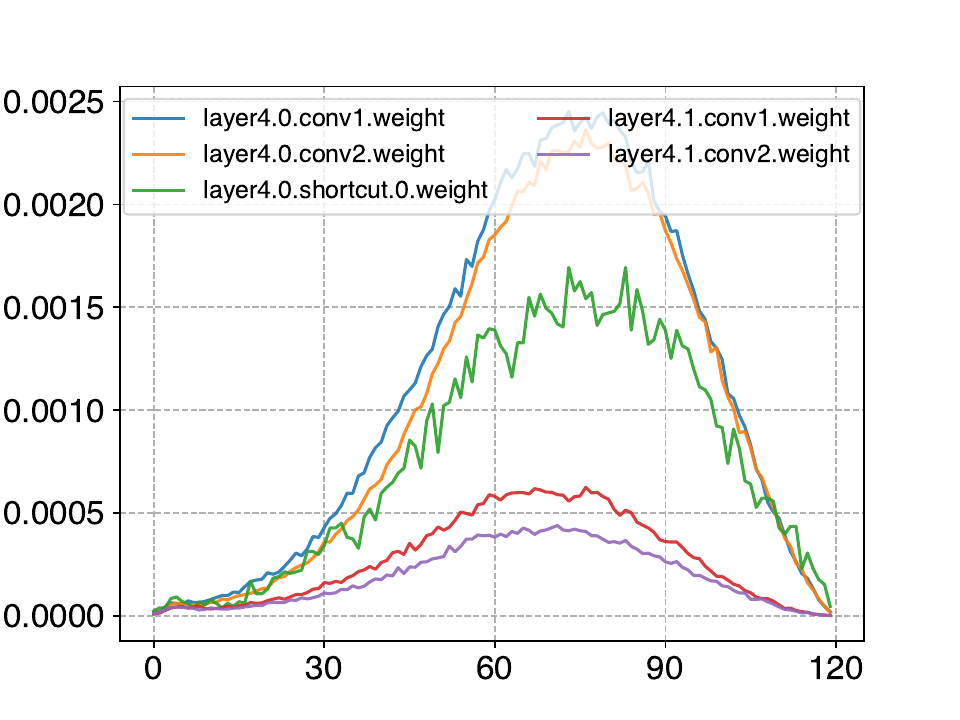}
        \caption{layer4}
    \end{subtable}
    \caption{Epoch-wise flip rate for $\gamma=100$~(kaiming uniform).}

{
    \centering
    \begin{subtable}[h]{0.24\textwidth}
        \includegraphics[width=1\linewidth]{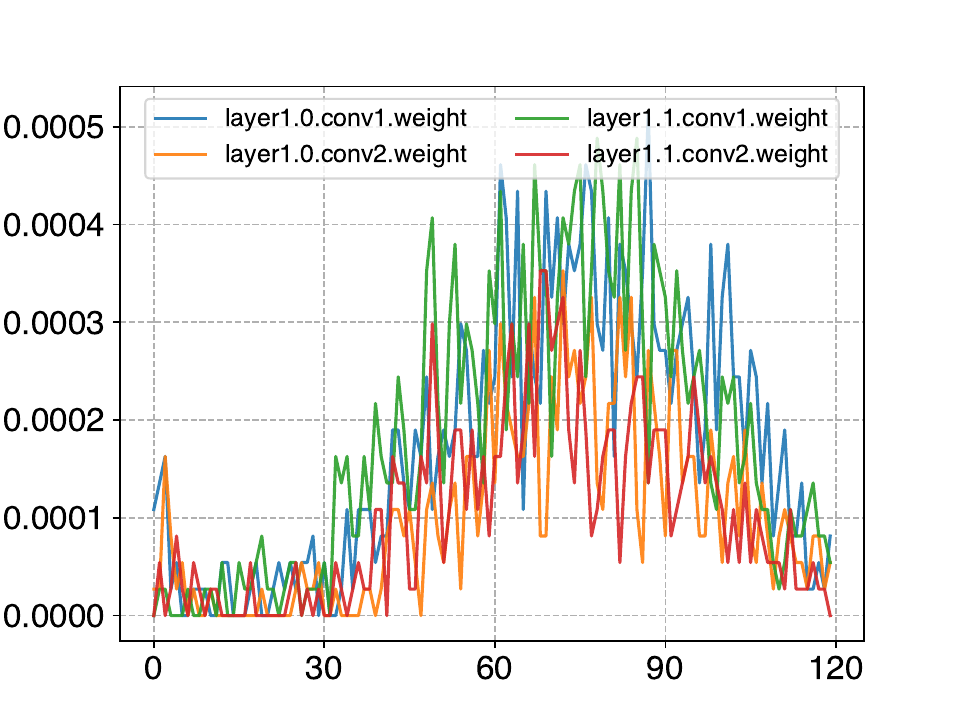}
        \caption{layer1}
    \end{subtable}
    \begin{subtable}[h]{0.24\textwidth}
        \includegraphics[width=1\linewidth]{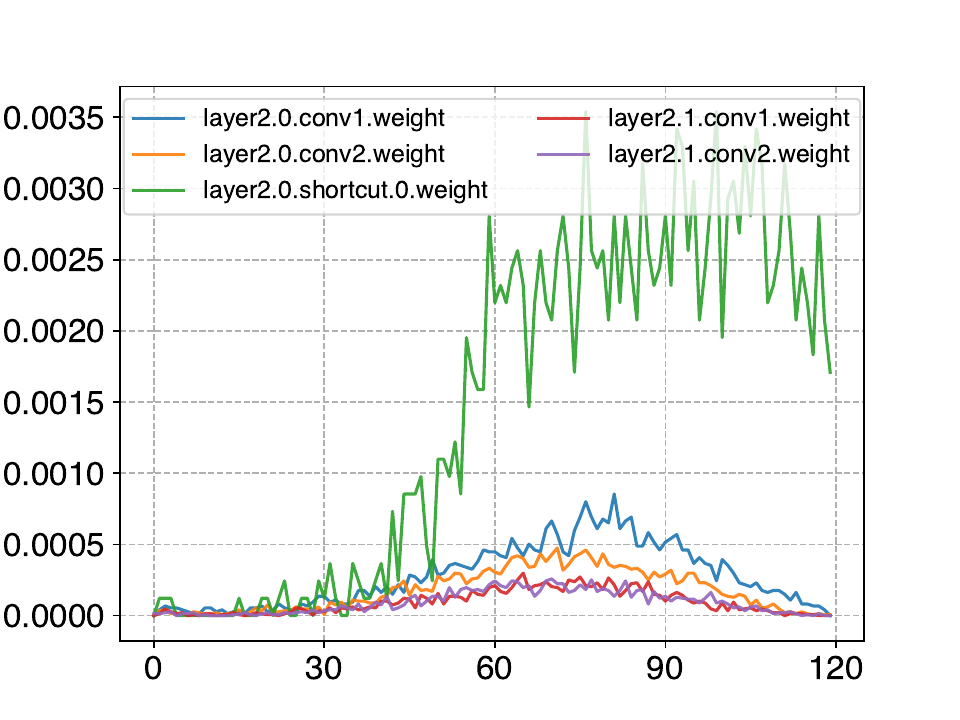}
        \caption{layer2}
    \end{subtable}
        \begin{subtable}[h]{0.24\textwidth}
        \includegraphics[width=1\linewidth]{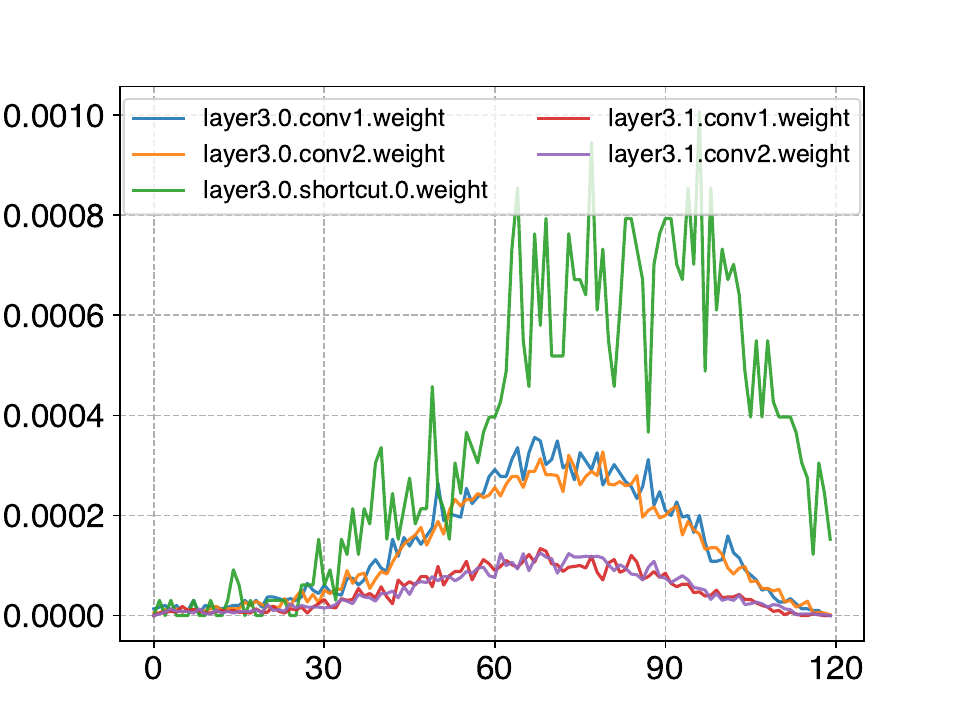}
        \caption{layer3}
    \end{subtable}
        \begin{subtable}[h]{0.24\textwidth}
        \includegraphics[width=1\linewidth]{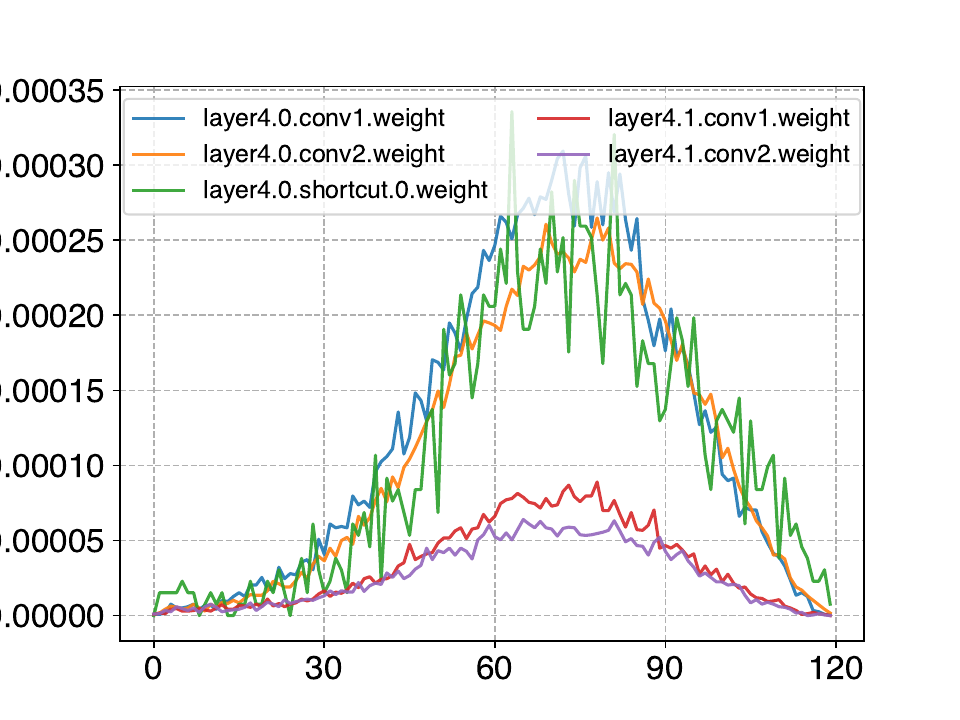}
        \caption{layer4}
    \end{subtable}
    \caption{Epoch-wise flip rate for $\gamma=1000$~(kaiming uniform).}
        \label{fig:uniform_1000}
}
\end{figure*}

\section{Ablation Analysis for Weight Initialization}
\label{appdendix:sec:ablation_analysis_weight_init}

We employ OvSW for two famous weight initialization methods, including kaiming normal~(default in this paper) and kaiming uniform, as shown in \cref{eq:kaiming_normal} and \cref{eq:kaiming_uniform} respectively:
\begin{equation}
\mathcal{W}_j\sim \mathrm{Normal}\big(0, \lambda^2\mathrm{std}^2\big)   
\label{eq:kaiming_normal}
\end{equation}
\begin{equation}
\mathcal{W}_j\sim \mathrm{Uniform}\big(-\lambda\mathrm{bound}, \lambda\mathrm{bound})
\label{eq:kaiming_uniform},
\end{equation}
where $\mathrm{std}=\frac{\mathrm{gain}}{\sqrt{\mathrm{fan\_in}}}$ and $\mathrm{bound}=\mathrm{gain}\times\sqrt{\frac{3}{\mathrm{fan\_in}}}$.
$\mathrm{fan\_in}$ is computed via ${C_{\text{in}}^{j}\times K^{j}_{\text{h}} \times K^{j}_{\text{w}}}$.
We train binarized ResNet18 for CIFAR100 with 120 epochs to demonstrate weight initialization analysis.
Without loss of generality, we set $\gamma$ to 1 and initialize $\mathcal{W}_{j}$ via the stand kaiming initialization and then employ $\mathcal{W}_{j}=\gamma\mathcal{W}_{j}^{{'}}$
to simulate \cref{eq:kaiming_normal} and \cref{eq:kaiming_uniform} in our implementations.

We present the epoch-wise flip rate for eight different settings of $\gamma$, which vary from 0.0001 to 1000.
The results in \cref{fig:normal_0.0001} to \cref{fig:normal_1000} and \cref{fig:uniform_0.0001} to \cref{fig:uniform_1000} are for kaiming normal and kaiming uniform respectively.
As seen, because of the gradients of the BNN being independent of their latent weight distribution, the epoch-wise flip rate and $\gamma$ have a significant negative correlation.
When $\gamma$ is set to 1000, the epoch-wise flip rate of both distributions undergoes a severe drop, greatly hurting the model's convergence as shown in \cref{fig:convergence_gamma} and performance as shown in \cref{fig:vary_gamma}.
Meanwhile, we can observe that by reducing the variance or range of the weight distribution to some extent,
a significant improvement can be achieved to the BNNs compared to the standard distribution, which demonstrates the reasonableness of AGS again.

\newpage
\section{AGS vs LARS\label{appendix:sec:ags_vs_lars}}

\begin{figure}[t]
    \centering
    \includegraphics[width=\linewidth]{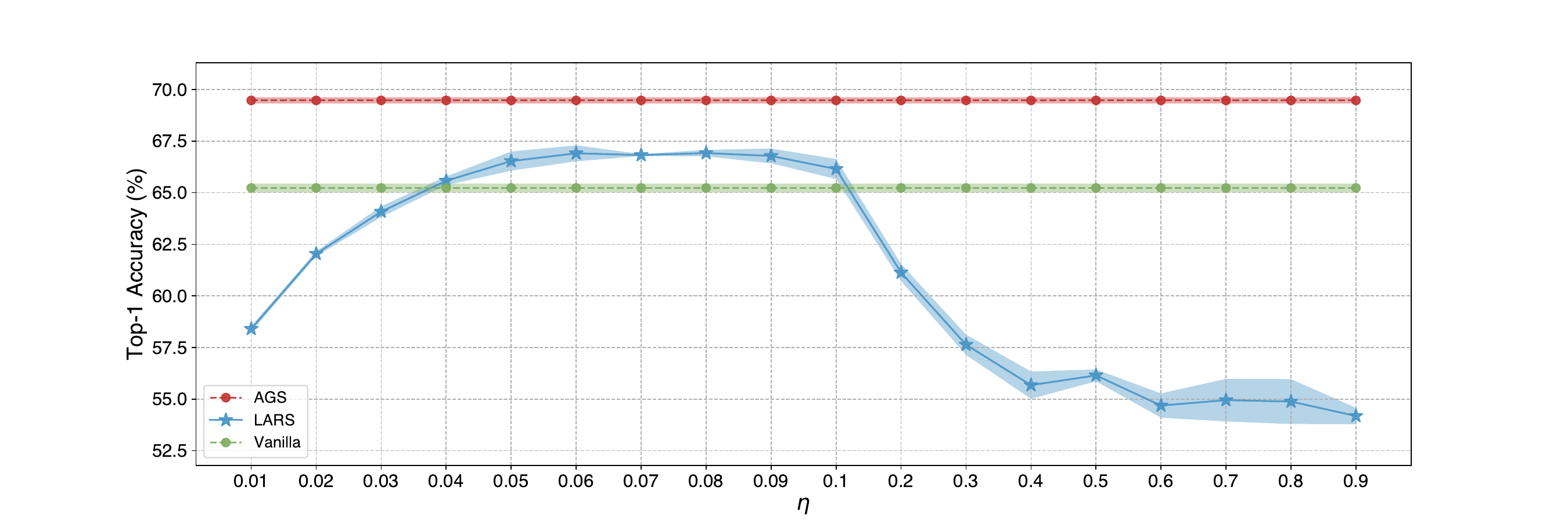}
    \vspace{-0.2in}
    \caption{Mean top-1 accuracy~(mean$\pm$std) of binarized ResNet18 \wrt different values for different $\eta$ with LARS on CIFAR100.}
    \label{fig:lars_ablation}
\end{figure}

%

%
In this section, we compare LARS and AGS~($\varphi=0$ in OvSW) and demonstrate their corresponding results in \cref{fig:lars_ablation}.
As seen, by selecting the appropriate parameter $\eta$, LARS is also able to achieve impressive gains by scaling the local learning rate.
However, the improvement of AGS is more significant over LARS.
By comparing the optimization processes of LARS and OvSW in \cref{algo:lars} and \cref{algo:ovsw},
We find that the most essential difference between them is that LARS only scales the learning rate at a single step,
and the scaling only acts on the current gradient descent and does not accumulate into the momentum;
in contrast, AGS directly modifies the gradient through adaptive gradient scaling,
and the gradient not only acts on the current, but also accumulates into the future optimization process through the momentum.
From the experimental results in \cref{fig:lars_ablation}, we can see that compared to LARS, AGS is more conducive to the efficient training of BNNs.

\newpage
\begin{algorithm}[htbp]
\begin{algorithmic}[1]
\STATE {\bf Parameters:} Learning rate $\beta(t)$, momentum $m$, weight decay $\varphi $, LARS coefficient $\eta$, number of steps $T$
\STATE {\bf Init:} $t = 0, v = 0$. Init weight $\mathcal{W}_j$ for each layer
\WHILE {$t < T$ for each layer} 
        \STATE $\mathcal{G}_j(t) \gets \frac{\partial  \mathcal{L}(t)}{\partial \mathcal{W}_{j}(t)}$   (obtain a stochastic gradient for the current mini-batch)
        \STATE $\lambda_{j}(t) \gets \frac{\eta \|\mathcal{W}_j(t)\|_{F}}{\|\mathcal{G}_j(t)\|_F + \varphi  \|\mathcal{W}_j(t)\|_{F}}$       (compute the local LR  $\lambda_{j}(t)$)
        \STATE {\color{blue}{$\mathcal{V}_{j}(t+1) \gets m\mathcal{V}_j(t) + \left(\mathcal{G}_j(t) + \varphi  \mathcal{W}_j(t) \right)$     (update the momentum)}}
        \STATE $\mathcal{W}_j(t+1) \gets \mathcal{W}_j(t) - \beta(t) \lambda_j(t) \mathcal{V}_{j}(t) $ (update the weights)
\ENDWHILE
\end{algorithmic}
\caption{SGD with LARS. Example with weight decay and momentum.}
\label{algo:lars}
\end{algorithm}

\begin{algorithm}[htbp]
\begin{algorithmic}[1]
\STATE {\bf Parameters:} Learning rate $\beta(t)$, momentum $m$, weight decay $\varphi $, $\lambda$ for AGS, $\tau$ for SAD, ($\mathcal{S}(t)$, $m$, $\sigma$) for flipping state detection, number of steps $T$
\STATE {\bf Init:} $t = 0, v = 0$. Init weight $\mathcal{W}_j$ for each layer
\WHILE {$t < T$ for each layer} 
        \STATE $\mathcal{G}_j(t) \gets \frac{\partial  \mathcal{L}(t)}{\partial \mathcal{W}_{j}(t)}$   (obtain a stochastic gradient for the current mini-batch)
        \STATE {\color{blue}{$\overline{\mathcal{G}}_{j}(t) \gets$  \cref{eq:ags}     (scale the gradient adaptively)}}
        \STATE $\overline{\mathcal{G}}_{j}(t) \gets$  \cref{eq:sad}     (silence awareness decaying)
        \STATE {\color{blue}{$\mathcal{V}_{j}(t+1) \gets m\mathcal{V}_j(t) + \left(\overline{\mathcal{G}}_j(t) + \varphi  \mathcal{W}_j(t) \right)$     (update the momentum)}}
        \STATE $\mathcal{W}_j(t+1) \gets \mathcal{W}_j(t) - \beta(t) \mathcal{V}_{j}(t) $ (update the weights)
        \STATE ${\mathcal{S}}_{j}(t) \gets$ \cref{eq:ema} (update state for the weights)
\ENDWHILE
\end{algorithmic}
\caption{SGD with OvSW. Example with weight decay and momentum.}
\label{algo:ovsw}
\end{algorithm}

\newpage
\section{Discussion for Latent Weights and Optimizer}
\label{appendix:sec:latent_adam}

\subsection{Latent Weights}
\label{appendix:ssec:latent}
We first discuss the similarities and differences between ours and Helwegen~\etal~\cite{helwegen2019latent}.
Similarly, both of us find that the gradient of the weights in the BNNs is independent of the magnitude of the weights.
Subsequently, Helwegen~\etal design a binary optimizer~(BOP), which determines the flipping of the weight signs by comparing the exponential moving average of gradients with a pre-defined threshold,
which is independent to the magnitude of weights and gradients.
While this approach avoids the problem that inappropriate weight distributions can lead to inefficient weight signs flipping, 
they ignore the fact that the weight magnitudes also play a role in sign changes during the optimization process.
OvSW constructed a correlation between the gradient distribution and the weight distribution via AGS,
which is essentially the same as BOP in facilitating weight signs flipping.
Meanwhile, the optimization takes the role of weight magnitude into account and thus achieves better results than BOP.
Apart from these, another advantage of OvSW is that SAD detects ``silent weights'' to further enhance the efficiency of weight signs flipping.

\subsection{Optimizer}
\label{appendix:ssec:optimizer}
OvSW emploies SGD to train BNNs, which is different from the previous state-of-the-art BNNs that uses Adam~\cite{kingma2014adam} as the optimizer,
including ReActNet~\cite{liu2020reactnet}, AdamBNN~\cite{liu2021adam}, RoBNN~\cite{xu2022recurrent}, ReBNN~\cite{xu2023resilient}.
%
%
From the perspective of weight signs flipping, we believe this is due to $\widehat{m_t}/\big(\sqrt{\widehat{v_t}} + \epsilon \big)$ in Adam
adaptively scales the gradients and facilitates the flipping efficiency.
%
%
However, Adam needs to preserve the first momentum and second momentum of the gradient during training, leading to additional storage.
In the mixed precision training scenario, a model with parameter number $\mathrm{\Psi}$ and Adam optimizer will consume $16\mathrm{\Psi}$~\cite{rajbhandari2020zero} storage for model and optimizer.
Even though OvSW introduce an auxiliary variable $\mathcal{S}$, it cause $14\mathrm{\Psi}$~\cite{rajbhandari2020zero} storage, which is 12.5\% less than Adam.
At the same time, OvSW can be simply and effectively implemented on GPUs, improving the performance of BNN with little or no degradation of training speed.

\newpage
\section{Societal Impact}
\label{appendix:sec:impact}

Increasing model size can result in tremendous resource consumption and carbon emissions during both training end inference.
OvSW can improve performance and accelerate the convergence efficiency of BNNs while introducing negligible memory and computational overhead.
It can facilitate the deployment of BNNs.
On the other hand, it also enables efficient training for BNNs under memory and computational resources constraints.
Both have far-reaching potential for the promotion of green AI.

\end{document}